\documentclass[letterpaper]{article} 
\usepackage{aaai2026}  
\usepackage{times}  
\usepackage{helvet}  
\usepackage{courier}  
\usepackage[hyphens]{url}  
\usepackage{graphicx} 
\usepackage{natbib}  
\usepackage{caption} 
\usepackage{algorithm}
\usepackage{algorithmic}

\usepackage{booktabs}
\usepackage[most]{tcolorbox}
\usepackage{enumitem}
\usepackage{multirow}
\usepackage{arydshln}
\usepackage{subfigure}
\usepackage{subcaption}
\usepackage{multirow}
\usepackage{pifont} 
\usepackage{longtable}

\usepackage{newfloat}
\usepackage{listings}
\DeclareCaptionStyle{ruled}{labelfont=normalfont,labelsep=colon,strut=off} 
\floatstyle{ruled}
\newfloat{listing}{tb}{lst}{}
\floatname{listing}{Listing}
\title{BhashaKritika: Building Synthetic Pretraining Data at Scale for Indic Languages}
\author{
    Guduru Manoj\equalcontrib, Neel Prabhanjan Rachamalla\equalcontrib, Ashish Kulkarni, Gautam Rajeev, \\
    Jay Piplodiya, Arul Menezes, Shaharukh Khan, Souvik Rana, \\
    Manya Sah, Chandra Khatri, and Shubham Agarwal \\[5pt]
}
\affiliations{
    \textsuperscript{\rm 1}Krutrim, India \\[5pt]
    \{neel.rachamalla1, ashish.kulkarni, shubham.agarwal1\}@olakrutrim.com 
}

\usepackage{bibentry}

\begin{document}

\maketitle

\begin{abstract}
In the context of pretraining of Large Language Models (LLMs), synthetic data has emerged as an alternative for generating high-quality pretraining data at scale. This is particularly beneficial in low-resource language settings where the benefits of recent LLMs have been unevenly distributed across languages. In this work, we present a systematic study on the generation and evaluation of synthetic multilingual pretraining data for Indic languages, where we construct a large-scale synthetic dataset \textit{BhashaKritika}, comprising 540B tokens using 5 different techniques for 10 languages. We explore the impact of grounding generation in documents, personas, and topics. We analyze how language choice, both in the prompt instructions and document grounding, affects data quality, and we compare translations of English content with native generation in Indic languages. To support scalable and language-sensitive evaluation, we introduce a modular quality evaluation pipeline that integrates script and language detection, metadata consistency checks, n-gram repetition analysis, and perplexity-based filtering using KenLM models. Our framework enables robust quality control across diverse scripts and linguistic contexts. Empirical results through model runs reveal key trade-offs in generation strategies and highlight best practices for constructing effective multilingual corpora.
\end{abstract}

\section{Introduction}
\label{sec:intro}

Most state-of-the-art LLMs \citep{ chowdhery2022palm, touvron2023llama,grattafiori2024llama,abdin2025phi} are trained predominantly on English corpora, available in abundance, leaving many of the world’s other languages underrepresented in both training data and model performance \citep{joshi2020state, mueller2022crosslingual}. 
\citet{villalobos2022will}
emphasize the finite nature of available pretraining data, often sourced from CommonCrawl~\citep{commoncrawl2007} and the need for alternative approaches to progress the state of LLMs. Even when multilingual datasets exist, they often suffer from issues related to quantity, quality, domain bias, diversity and inconsistent formatting \citep{conneau2020unsupervised}. Thus, open-access pretrained models with strong multilingual capabilities remain limited, especially for low-resource and morphologically rich Indian languages. 
Hindi, for instance, does not appear in the top 20 languages of Common Crawl despite being the third most spoken globally~\citep{Penedo2023TheRD} and Indian languages collectively constitute less than $1\%$~\citep{kallappa2025krutrim}. This scarcity of both data and models presents a major barrier to the development of culturally inclusive LLMs especially with recent data constrained scaling laws  \citep{muennighoff2023scaling} arguing that model performance show degradation after 4 epochs on repeated data. 

Synthetic data generation has thus emerged as a viable approach where, training data is artificially generated while mirroring the features, structures, and statistical attributes of real-world data \citep{nadas2025synthetic,liu2024best,yu2023large}. This offers a compelling alternative to conventional web-scraping and manual curation while providing control and diversity compared to web data. By leveraging existing LLMs as generators, it is possible to create large-scale, language-diverse corpora that is customizable and replicable
\citep{selfinstruct2022, taori2023stanford, longpre2023flan, chen2023synthia}.
The Phi series of models \citep{gunasekar2023textbooks, li2023textbooks,abdin2024phi} focused on proprietary synthetic data as part of their pre-training corpus and showed its efficacy in their training pipeline.  
\citet{benallal2024cosmopedia}, created the open-source \textit{Cosmopedia} consisting of $25$B English synthetic tokens, grounded in web documents.  \citet{ge2024scaling} introduced \textit{PersonaHub}, a collection of English personas, that are then used for persona-grounded synthetic generation. 
Here, a `persona' is defined as `a person with specific professional experiences and cultural backgrounds having unique interests in reading and writing'. 

\begin{figure*}[htbp]
    \centering
\includegraphics[width=0.9\linewidth]{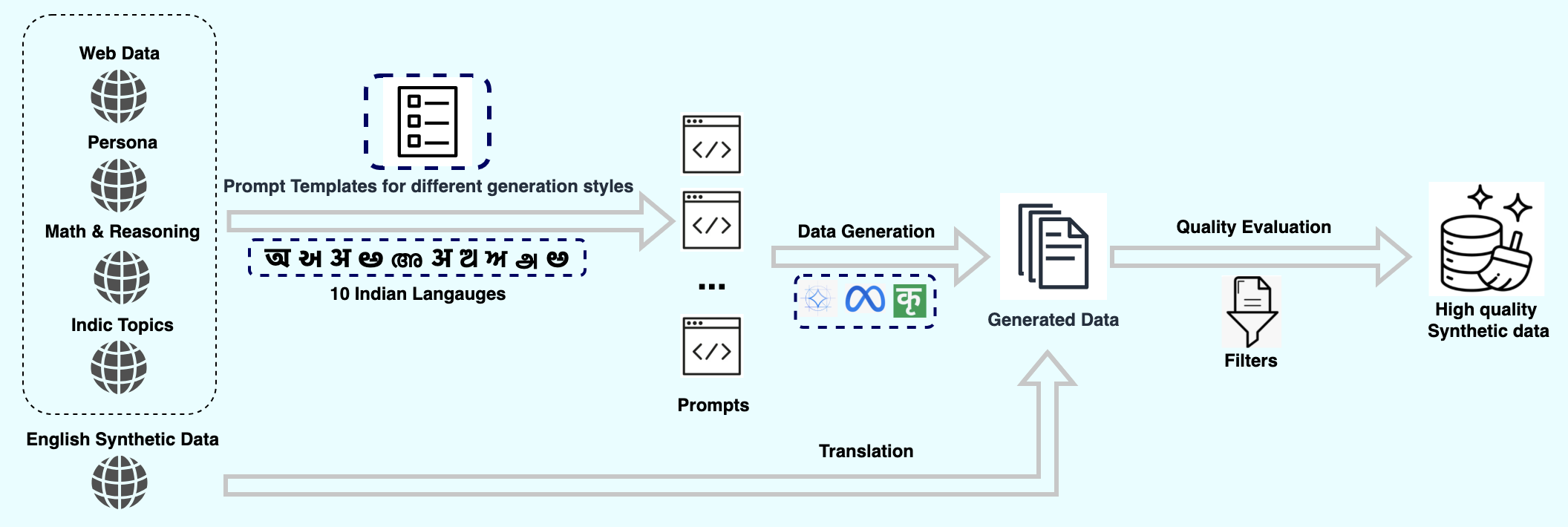} \caption{Overview of Synthetic data generation techniques  (Section \ref{sec:syn-gen}) followed by Quality Evaluation (Section \ref{sec:quality-eval}). We follow 5 approaches across 10 Indian languages using a pool of Multilingual LLMs to generate a large scale \textit{BhashaKritika} corpora.}
    \label{fig:mindmap}
\end{figure*}

In this work, we propose a pipeline for generating high-quality synthetic
pretraining text data focusing on both Indian languages and local context. Our approach builds on previous work and involves language-aware prompt engineering, style and domain variation, and automated quality filtering to ensure broad linguistic coverage and coherence.

Specifically, we make the following contributions:

\begin{itemize}
    \item We develop a modular pipeline for generating large-scale high-quality synthetic Indic multilingual corpora. We design prompt templates, data curation pipelines, generation strategies, and conduct ablations across languages and models that collectively ensure that the generated data is factually grounded, knowledge-dense, rich in Indic cultural context, and topically diverse.
    \item We propose a novel method for constructing math-focused pretraining data by transforming instruction-tuned datasets into pretraining-style corpora through controlled synthetic generation.
    \item We implement an automated quality filtering pipeline, covering language consistency, fluency, heuristic filters, statistical filters, quality classifiers, bias detection and mitigation strategy in the generated data. 
    \item We use our synthetic generation pipeline to generate \textit{BhashaKritika}, a $540$B tokens high-quality Indic multilingual synthetic corpus. We also share a part of this data for public use\footnote{\url{https://huggingface.co/datasets/krutrim-ai-labs/BhashaKritika}}.
    \item We perform extensive analysis with additional two controlled experiments including annealing as well as pretraining a 1B param from scratch and show models trained on synthetic data continue to improve and closely match the one trained with the real data. 
\end{itemize}

\section{Related Work}
\label{sec:related-work}

\subsection{Web Crawled Datasets}

For text-based pretraining, large-scale datasets such as The Pile~\citep{Gao2020ThePA}, C4~\cite{raffel2020exploring}, RedPajama~\citep{together2023redpajama}, RefinedWeb~\citep{Penedo2023TheRD}, Dolma~\citep{Soldaini2024DolmaAO}, DataComp-LM~\citep{Li2024DataCompLMIS}, and FineWeb~\citep{penedo2024fineweb} have been instrumental in training LLMs. Nemotron-CC \citep{su2024nemotron} further refines this effort with a high-quality subset of 6.4T tokens while MegaMath \citep{zhou2025megamathpushinglimitsopen} filters for Math datasets. Mostly sourced from CommonCrawl \citep{commoncrawl2007}, these datasets however, predominantly feature English and other high-resource languages, offering limited coverage of Indian languages or culturally grounded content.
More recently, FineWeb2 \citep{penedo2024fineweb2}
introduces broader language coverage, however only a small portion, around $40$B words, pertains to Indian languages.

\subsection{Indic LLM Research}
Most prior work has focused on adapting existing English-dominant models through fine-tuning or continued pretraining on Indic language corpora
\citep{balachandran2023tamilllama, kohli2023building, gala2024airavata, sarvam2023openhathi, nanda2024choudhury}. 
In contrast, only a few models
\citep{team2024krutrim, bendale2024sutra, sarvam2024llm} have been trained from scratch, aiming to create more culturally inclusive LLMs for the Indian context.
Alongside model development, there have been ongoing efforts to curate multilingual datasets focused on Indian languages. One of the earlier efforts in this direction, the IndicNLP Corpora \citep{kunchukuttan2020indicnlpcorpus}, compiled around $2.7$B tokens for 10 Indic languages from web-based content which was later expanded to IndicCorp \citep{kakwani-etal-2020-indicnlpsuite}, comprising 8.8B tokens across 11 Indian languages and English.  \citet{gala2024airavata} released Indic Instruct Data v0.1, a Hindi instruction-tuning dataset derived through translation of pre-existing instruction sets. Additionally, the Sangraha corpus \citep{khan2024indicllmsuite} offers a collection of $251$B tokens covering 22 Indian languages, nonetheless, its scale remains modest compared to the much larger corpora generally available for English (in $5-15$T tokens) and the other Western languages.

\subsection{Synthetic Data Generation}

Synthetic data techniques have become a valuable resource for enriching both fine-tuning and pretraining corpora. Early instruction-tuning methods include Self-Instruct \citep{wang2022self}, Evol-Instruct \citep{xu2023wizardlm} and Magpie \citep{xu2024magpie} to name a few. Beyond fine-tuning, synthetic data for pre-training has also shown promise, notably in the proprietary Phi models \citep{li2023textbooks,abdin2024phi}. Open-source alternative Cosmopedia \citep{benallal2024cosmopedia} offer 25B tokens of diverse synthetic text generated in English. Recent work has also explored persona-based generation to increase diversity and alignment with PersonaHub introducing 1M synthetic personas \citep{ge2024scaling} which Nemotron-Personas further aligns personas with demographic and psychological traits \citep{meyer2025nemotron}. Similar techniques have been applied in multimodal settings \citep{yang2025scaling} and domain-specific tasks like math and reasoning \citep{lambert2024tulu}. \citet{odumakinde2024multilingual} proposed a multilingual arbitrage framework to further improve teacher model selection across languages. Finally, synthetic data scaling laws proposed by \citep{qin2025scaling} emphasize the interplay of quantity, diversity, and generation methods. We build upon these works to generate pre-training synthetic data for Indian languages.

\section{Synthetic Data Generation Techniques}
\label{sec:syn-gen}
We develop a pipeline to generate synthetic data at scale and demonstrate how we used it to generate $540$B tokens of high-quality Indic synthetic data. Our pipeline leverages different Web data sources (both direct and derived) as context, multiple generation techniques and output styles that together ensure that the generated synthetic data is factually grounded, knowledge-rich, and topically diverse.

\subsection{Document Grounded Generation}
\label{subsec:cosmo}
Following work on English synthetic data generation~\citep{benallal2024cosmopedia,su2024nemotron,maini2024rephrasingwebrecipecompute,li2023textbooks,gunasekar2023textbooks}, we leverage multilingual LLMs, prompted with documents from the Web as context, to generate Indic synthetic data in different knowledge-rich formats and creative styles. \citep{benallal2024cosmopedia,su2024nemotron,maini2024rephrasingwebrecipecompute,li2023textbooks,gunasekar2023textbooks}. 
In addition to using documents from English FineWeb \citep{penedo2024fineweb} and multilingual FineWeb2 \citep{penedo2024fineweb2} directly, we also selectively curate ``Indic context'' documents. 
The Indic context documents are identified using a FastText-based classifier~\citep{joulin2016fasttext} 
trained on $93$K annotated documents. 
We also perform clustering by adapting Huggingface text-clustering\footnote{\url{https://github.com/huggingface/text-clustering}} to get broad topics for Indian context data, that we optionally append to the prompts for document grounded generations.

We evaluate (via human annotation) different multilingual LLMs based on their quality of direct generations in a language xx and generations in En followed by their translation to xx. Table~\ref{tab:lang-model-mapping} shows the language-model mapping that we followed for our synthetic generation. In addition to ensuring language-specific quality, the use of multiple LLMs in our pipeline also alleviates model-specific biases, avoids model collapse, and encourages generalization, diversity, and robustness in our generated data~\cite{agarwal2025language,odumakinde2024multilingual}.

Taking inspiration from prior works~\citep{benallal2024cosmopedia, su2024nemotron, maini2024rephrasingwebrecipecompute, akter2025mind} that show efficacy of ``textbook-like'' and ``educational'' data in pretraining LLMs, we use several knowledge-rich formats such as textbook entries, blog posts, wikihow, \textit{inter alia} to enhance factual synthesis and structured reasoning. We also use several creative styles such as moral stories, poetry, reddit posts and others to encourage generative fluency and imagination. We list all prompts utilised in Appendix \ref{sec:prompts_used}.


\begin{table}[htbp]
\centering
\tiny
\setlength{\tabcolsep}{1pt} 
\begin{tabular}{p{1cm}|p{0.9cm}p{0.9cm}p{1.5cm}|p{0.9cm}p{0.9cm}p{1.5cm}}
\hline
\multirow{2}{*}{\textbf{Language}} & \multicolumn{3}{c|}{\textbf{Generate in xx}} & \multicolumn{3}{c}{\textbf{Generate in En and translate to xx}} \\
 & Gemma-3 & Krutrim-2 & LLaMA-3.3 70B & Gemma-3 & Krutrim-2 & LLaMA-3.3 70B \\
\hline
Bengali   & \enspace\quad\ding{51} & \enspace\quad\ding{55} & \enspace\quad\ding{55} & \enspace\quad\ding{55} & \enspace\quad\ding{55} & \enspace\quad\ding{55} \\
Gujarati  & \enspace\quad\ding{55} & \enspace\quad\ding{51} & \enspace\quad\ding{55} & \enspace\quad\ding{55} & \enspace\quad\ding{55} & \enspace\quad\ding{55} \\
Hindi     & \enspace\quad\ding{51} & \enspace\quad\ding{51} & \enspace\quad\ding{55} & \enspace\quad\ding{55} & \enspace\quad\ding{55} & \enspace\quad\ding{55} \\
Kannada   & \enspace\quad\ding{55} & \enspace\quad\ding{55} & \enspace\quad\ding{55} & \enspace\quad\ding{55} & \enspace\quad\ding{51} & \enspace\quad\ding{55} \\
Malayalam & \enspace\quad\ding{55} & \enspace\quad\ding{51} & \enspace\quad\ding{51} & \enspace\quad\ding{55} & \enspace\quad\ding{55} & \enspace\quad\ding{55} \\
Marathi   & \enspace\quad\ding{51} & \enspace\quad\ding{51} & \enspace\quad\ding{55} & \enspace\quad\ding{55} & \enspace\quad\ding{55} & \enspace\quad\ding{55} \\
Oriya     & \enspace\quad\ding{55} & \enspace\quad\ding{55} & \enspace\quad\ding{51} & \enspace\quad\ding{55} & \enspace\quad\ding{55} & \enspace\quad\ding{55} \\
Punjabi   & \enspace\quad\ding{51} & \enspace\quad\ding{51} & \enspace\quad\ding{51} & \enspace\quad\ding{55} & \enspace\quad\ding{55} & \enspace\quad\ding{55} \\
Tamil     & \enspace\quad\ding{51} & \enspace\quad\ding{51} & \enspace\quad\ding{55} & \enspace\quad\ding{51} & \enspace\quad\ding{55} & \enspace\quad\ding{55} \\
Telugu    & \enspace\quad\ding{51} & \enspace\quad\ding{51} & \enspace\quad\ding{55} & \enspace\quad\ding{55} & \enspace\quad\ding{55} & \enspace\quad\ding{55} \\
\hline
\end{tabular}
\caption{Language-wise mapping of models used for direct generation (in language xx) and translation (generation in En followed by its translation to xx). Corresponding detailed human evaluations for the models are included in the Appendix in  Table \ref{tab:genvstrans-model-eval} - \ref{tab:genvstrans-telugu}. Interestingly, LLaMA-3.3 70B~\citep{grattafiori2024llama} showed superior performance than LLaMA-4~\cite{meta2025llama} series for Indian languages.}
\label{tab:lang-model-mapping}
\end{table}

\subsection{Persona-Based Generation}
\label{subsec:persona}
We leverage PersonaHub \citep{ge2024scaling}, an open source repository of 
$371$M personas, to synthetically generate $164.3$M English Indic context personas. Additionally, we follow their approach to synthetically generate $50$K Indic language personas, from Indic Web documents, that cover the diverse Indian linguistic, regional, and sociocultural identities. An example of a generated persona from our dataset: \textit{A young software engineer from Bangalore who codes all day and hits the gym hard at night}.

We then use these personas as context in our synthetic generation pipeline following two approaches: (1) \textit{Persona-based generation}, guided solely by the persona and language input to produce free-form, culturally fluent text; and (2) \textit{Persona and document-based generation} where a persona is paired with a document, sampled either at random or based on its semantic similarity to the persona, for a more controlled and contextually rich generation.

\subsection{Math and Reasoning-Based Synthetic Data}
\label{subsec:math}
We introduce a novel methodology for generating high-quality pretraining data from existing instruction-tuning datasets for math and reasoning. Our method transforms existing, verified Question-Solution (Q-S) pairs from instruction-tuning datasets \citep{hendrycks2021measuringmathematicalproblemsolving, numina_math_datasets, moshkov2025aimo2} into comprehensive and self-sufficient textbook sections. Specifically, we condition a generation model on a Q-S pair and instruct it to first introduce the underlying mathematical or technical concepts and theorems required to understand the problem, and then present a detailed, step-by-step solution. 
We posit that this approach offers two key advantages. Firstly, because the generation is grounded in an already-verified solution, it maintains the mathematical correctness and obviates the need for an additional, complex verification step. Secondly, we hypothesize that 
this ``concept-then-solution'' format will better equip models to emulate human-like reasoning
\subsection{Topic-Aware Retrieval Augmented Generation (RAG)}
\label{subsec:topic-based-rag}
To ensure extensive and accurate coverage of the Indian context, especially within long-tail topics, we first curate a detailed collection of Indic-specific topics. This is accomplished by systematically traversing the Wikipedia knowledge graph starting from the root node \textit{Category:India}\footnote{\url{https://en.wikipedia.org/wiki/Category:India}} up to a depth of three, resulting in a dataset containing over 10,000 topic titles. Next, we cluster our existing synthetic data using Vyakyarth\footnote{\url{https://huggingface.co/krutrim-ai-labs/Vyakyarth}}- a multilingual semantic embedding model tailored for Indic languages. We then filter the identified Indic topics by nearest neighbour based similarity score and subsequently applying a distance threshold. This ensures the retained topics are different from the topics already covered by the previously generated synthetic data. Finally, for each filtered topic, we utilize the SERP API\footnote{\url{https://serpapi.com/}} to retrieve relevant external documents. Leveraging these retrieved documents, we apply Retrieval Augmented Generation (RAG) techniques \citep{lewis2020retrieval} to generate contextually accurate and linguistically diverse content in multiple Indian languages.

\subsection{Translation of English Synthetic Data}
\label{subsec:trans}
In addition to the different generation strategies discussed earlier, we also translate the $25$B English synthetic \textit{Cosmopedia} \citep{benallal2024cosmopedia} dataset, originally generated using the Mixtral-8x7B-Instruct-v0.1 model~\citep{jiang2024mixtral}. We evaluate various translation models across languages (Refer to Table~\ref{tab:trans-eval} in the Appendix) 
and select Sarvam-Translate~\citep{sarvamai_sarvam_translate} for this translation. In order to ensure knowledge diversity across languages, we translate each of the $30$M documents in \textit{Cosmopedia} to only one randomly sampled Indic language. 
\begin{table*}[htbp]
\centering
\resizebox{\textwidth}{!}{%
\begin{tabular}{lrrrrr}
\hline
\textbf{Type}  & \textbf{Generated Tokens (B)} & \textbf{Filtered Tokens (B)} & \textbf{Discard Rate (\%)} & \textbf{Avg. source length} & \textbf{Avg. generated length}\\
\hline
Document grounded - En (Section \ref{subsec:cosmo}) & 394.85	& 382.94      & 3.36 & 150 & 414 \\
Document grounded - Indic (Section \ref{subsec:cosmo})    & 63.75	& 62.88  & 1.45 & 186 & 460 \\
Persona based   (Section \ref{subsec:persona})    & 37.83 &	37.24  & 1.56 & 34 & 242 \\
Math \& Reasoning (Section \ref{subsec:math})      & 5.09    & 4.83 &  4.80 & 236 & 624\\
Topic based RAG  (Section \ref{subsec:topic-based-rag})   & 0.13    & 0.13 &  3.14 & 124 & 568\\
Translation (Section \ref{subsec:trans}) & 57.69 &	55.26 &  4.10 & 540 & 572 \\
\hline
\end{tabular}
}
\caption{Generated and filtered token counts (in billions) for each synthetic data source. Token counts are estimated using the LLaMA-4 tokenizer. We show discard rate as \% of data filtered out and also report average output length (in words).} 
\label{tab:data-stats}
\end{table*}

\section{Quality Evaluation Pipeline}
\label{sec:quality-eval}

Recent scaling laws \citep{chang2024scaling, chen2025revisiting} have argued the importance of quality data in pre-training. 
In order to assess the quality of synthetic data and filter out low-quality data at scale, we develop an automated quality evaluation pipeline comprising multiple heuristic and model-based filters outlined below.

\textbf{1. Language consistency filter:}
Multilingual LLMs, especially when used in mid-to-low resource language settings, might generate text in mixed languages or in a language different from the intended language in the prompt. 
To ensure the generated data is in the target language, we leverage an ensemble language identification (LID) module optimized for Indian languages, building on top of the recent works~\citep{khan2024indicllmsuite}.

\textbf{2. Heuristic content filters:} This module filters low-quality text in our generated large-scale corpora using rule-based heuristics and statistical features. It targets undesirable content such as NSFW material, excessive stopword or word repetition, anomalous characters (e.g., non-Latin/Indic scripts), outlier word counts, generic boilerplate, and references to third-party AI systems. Each criterion is governed by empirically tuned thresholds, and texts exceeding these limits are excluded to ensure high-quality data for downstream tasks, reported in Appendix (see Table \ref{tab:filter-thresholds}).

\textbf{3. Fluency evaluation (Perplexity filtering):} In order to evaluate the fluency of the generated synthetic data, we train a 5-gram Kneser-Ney model using the KenLM \citep{heafield2011kenlm} library.

The model is trained on $14.5$M  high-quality text samples sourced from Wikipedia, Sangraha \citep{khan2024indicllmsuite}, FineWeb2 \citep{penedo2024fineweb2}, and bootstrapped synthetic corpora.  
For each data point, a perplexity score is computed and compared against language-specific thresholds, where low scores denote high linguistic coherence in the generated data. These thresholds are determined using held-out validation sets, with the $80$th percentile of the score distribution used as the default cutoff, following earlier works \citep{khan2024indicllmsuite}. Further details regarding training and validation data used are in the Appendix (Table \ref{tab:perp-data}).

\textbf{4. Quality classifiers:}
 We also evaluate overall quality of the generated synthetic data on aspects such as content accuracy, clarity, coherence, grammatical correctness, informational depth, and overall usefulness, using a custom-trained FastText~\citep{joulin2016fasttext} binary classifier to automatically assess the quality of Indic-language responses, labeling each instance as either \texttt{high} or \texttt{low} quality. The classifier is trained on approximately $384$K examples labeled using the Gemini-1.5-Flash model through prompt-based evaluation. 
The training data comprises samples from diverse, high-quality sources such as Wikipedia, Sangraha \citep{khan2024indicllmsuite}, FineWeb2 \citep{penedo2024fineweb2}, and 
generated synthetic corpora.
The model achieves an overall accuracy of $98.9\%$, demonstrating high precision and recall across both quality classes on test split consisting of $160$K examples. Details on training data composition, language-wise test set distribution, and evaluation metrics are provided in the Appendix (see Tables \ref{tab:classifier-data}, \ref{tab:evaluation-metrics}). On the source English documents, we also leverage pretrained models from 
NeMo Curator\footnote{\url{https://github.com/NVIDIA/NeMo-Curator}} library
including the \textit{FineWebEdu} classifier for detecting high-quality educational content and the \textit{Domain Classifier} for categorizing the text into broad domains such as science, health, finance etc.

\textbf{5. Bias detection and mitigation:}
We leverage the Word Embedding Association Test (WEAT)~\citep{jentzsch2019semantics} to quantify the social and cultural biases in our generated data. WEAT measures the strength of association between predefined \textit{target} and \textit{attribute} word sets in an embedding space, providing a quantitative estimate of implicit bias. The word embeddings are obtained from language-specific FastText models trained on our synthetic dataset. Our evaluation focuses on five key dimensions of social bias: gender, caste, race, religion, and regional/linguistic identity (Refer to Table~\ref{tab:weat-bias-aspects} in the Appendix). For each dimension, we capture representative stereotypes using manually curated target and attribute word sets, each comprising $18$–$20$ terms per language (Appendix Figures \ref{fig:bias_words_caste}-\ref{fig:bias_words_religion} show manually curated Hindi bias words across various bias aspects). Higher WEAT scores (typically $>1.0$) correspond to stronger stereotypical associations.

\section{BhashaKritika: Synthetic Data}
\begin{figure*}[htbp]
    \centering
    \begin{minipage}[t]{0.42\textwidth}
        \centering
        \includegraphics[width=\linewidth]{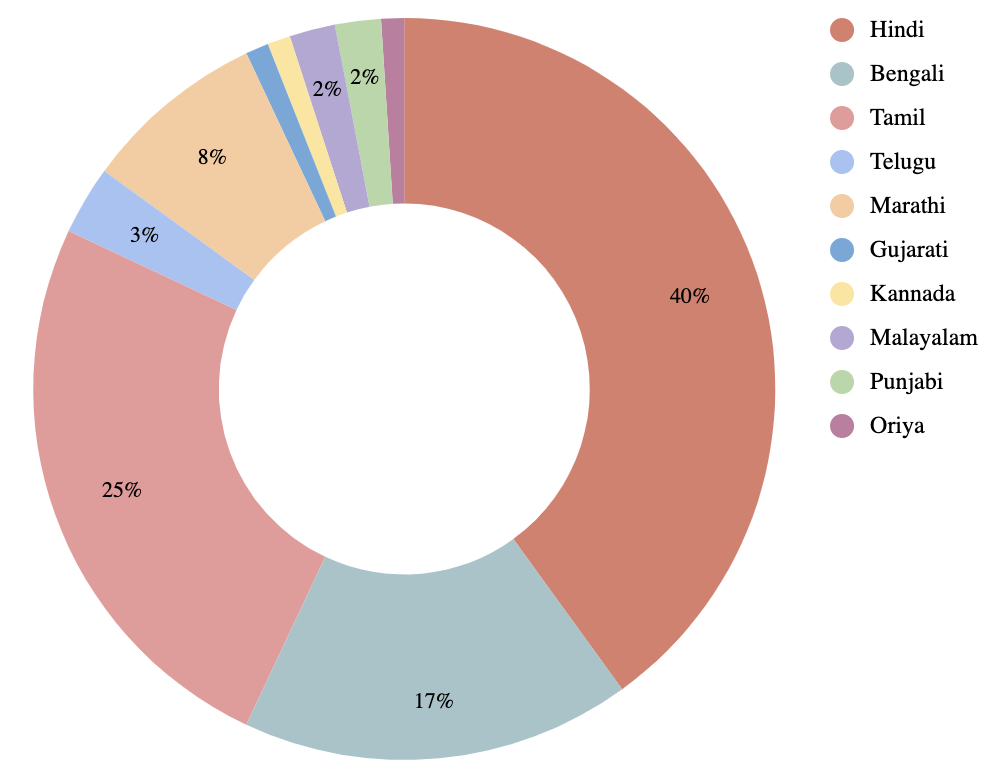}
    \end{minipage}%
    \hfill
    \begin{minipage}[t]{0.58\textwidth}
        \centering
        \includegraphics[width=.9\linewidth]{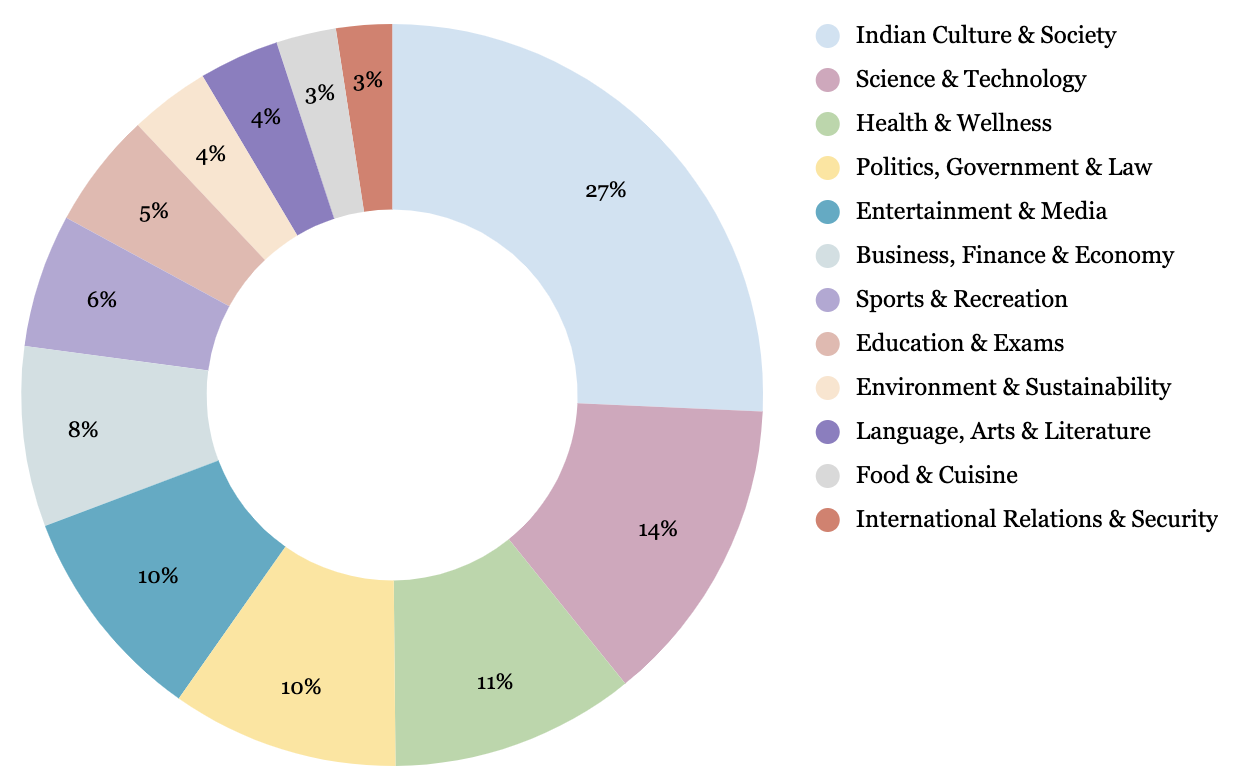}
    \end{minipage}
    \caption{Distribution of languages (left) and topics (right) in BhashaKritika. We show the broad 12 topics for brevity with a more fine-grained distribution in Table~\ref{tab:broad_topic_distribution}  (Appendix).}
    \label{fig:lang-topic-dist}
\end{figure*}

We used our pipeline to generate $\sim540$B tokens of high-quality synthetic data covering multiple Indian languages and Indic context topics. In Table~\ref{tab:data-stats}, we show the distribution of this data by different sources used for generation. Here, ``filtered tokens'' correspond to the data that passes our quality evaluation pipeline and the ``discard rate'' is the percent of the synthetic data that is filtered out. Figure \ref{fig:lang-topic-dist} illustrates the language-wise and topic-wise distribution of our synthetic data respectively. Each of these $12$ topics in turn covers multiple Indic context sub-topics, for instance, \textit{Indian culture and society} subsumes \textit{Indian lifestyle}, \textit{Indian philosophy}, \textit{Indian fashion}, and others. We provide a comprehensive report of the different prompts, classifier datasets, annotation instructions as well as the quality evaluation in the Appendix for reproducibility. 

\section{Experiments}
\label{sec:experiments}
We conduct several ablations over the data sources, their language, the language of prompt instructions, and the personas to inform our choices in the synthetic generation process. Also, in order to evaluate the efficacy of our synthetic data in pretraining LLMs, we conduct experiments with a $1$B parameter LLaMA-3.2 architecture in the compute constrained settings. We report the key findings here.

\subsection{Does Language of Source Document and Prompt Impact the Quality of Generations?}
Our synthetic data generation pipeline leverages source documents in both English (\emph{e.g.} FineWeb) and Indian languages (\emph{e.g.} FineWeb2) as the grounding context. \textit{How does the language of this context impact the quality of generated synthetic data?} \textit{For context in Indic languages, is it better to provide prompt instructions in English or Indic languages?} In order to answer these questions, we conduct an evaluation on the documents sampled from  Pralekha~\citep{suryanarayanan2024pralekha}, a large-scale parallel document dataset in English and Indic languages, as context and the prompt instruction in English or Indic. We leverage our quality evaluation pipeline and report the discard rate on the generated synthetic data (Table~\ref{tab:lang_impact}). We observe that the models perform better when prompted with context in the same language as the intended language of generation and prompt instructions in English perform better than those in Indic. We, therefore, use prompts in English across all our synthetic generations.

\begin{table}[htbp]
\centering
\scriptsize
\setlength{\tabcolsep}{2pt}
\begin{tabular}{lrrr}
\hline
\textbf{Language} & \textbf{En/En} & \textbf{Ind/En} & \textbf{Ind/Ind} \\
\hline
Bengali & 0.57 & 0.47 & \textbf{0.33} \\
Gujarati & 17.45 & \textbf{15.85} & 24.15 \\
Hindi & 1.025 & \textbf{0.6} & 1.125 \\
Malayalam & 11.50 & \textbf{9.70} & 11.40 \\
Marathi & 1.20 & \textbf{1.05} & 0.93 \\
Punjabi & 7.97 & 5.03 & \textbf{4.97} \\
Tamil & 3.83 & \textbf{3.00} & 5.30 \\
Telugu & 3.10 & \textbf{3.03} & 4.73 \\
\hline
\end{tabular}
\caption{Impact of language of source document and prompt instructions on quality of generations (discard rates \%). Ind/En denotes Indic document with English prompt.}
\label{tab:lang_impact}
\end{table}

\begin{table}[htbp]
\centering
\scriptsize
\setlength{\tabcolsep}{2pt}
\begin{tabular}{lr}
\hline
\textbf{Generation Mode} & \textbf{Discard rate\% } \\
\hline
Indic Persona & 1.48\%  \\
\hdashline
En Persona & 1.93\%  \\
En Persona with matching document & 3.50\% \\
En Persona with random Document & 14.43\% \\
\hline
\end{tabular}
\caption{Impact of language of source persona and additional document grounding through a pilot study. }

\label{tab:persona_model}
\end{table}

\subsection{Does Language of Persona Impact the Quality of Generations?}
We conduct a pilot study to evaluate the impact of additional document grounding by appending personas with either matching or random documents, selected using FAISS similarity scores. As shown in Table~\ref{tab:persona_model}, this grounding significantly increases discard rates, suggesting that random pairing introduces linguistic inconsistencies and quality degradation. Additionally, generations conditioned on Indic personas yield lower discard rates compared to those grounded in English personas.

\subsection{How Much Data is Filtered Out by the Quality Evaluation Pipeline and Why?}
We ablate over the quality filters in the quality evaluation pipeline and present our insights. Refer to Table~\ref{tab:lang-wise-filter-rates} in the Appendix for filter-wise discard rates across languages.

\textbf{1. Language consistency filter:}
Language inconsistency is predominantly observed in Gujarati and Hindi languages (over $10\%$ compared to an average of $7.6\%$ across languages). This is primarily due to generations in other languages from the same language family 
(Marathi or Sanskrit instead of Hindi) and regional references in the document context. For instance, a news article mentioning Telangana Govt. used as context leads to generation in Telugu instead of Gujarati (target language mentioned in the prompt). 

\textbf{2. Heuristic content filters:}
Length violations are the most common issue, affecting $2.26\%$ generations, primarily due to incomplete or excessively verbose generations outside the $100$–$2500$ word range. While toxic content is generally well-controlled, $1.13\%$ of outputs contain NSFW material, and $2.02\%$ include references to other AI systems.
Word repetition ($0.34\%$) and the use of excessive stopwords or non-Latin/non-Indic scripts (under $0.01\%$) remain rare. However, this filtering relies on manually curated keyword lists that are not exhaustive, and certain NSFW terms are context dependent, occasionally leading to false positives. 

\textbf{3. Fluency (perplexity-based) filter:}
Most generations are reasonably fluent, 
however, certain languages like Tamil and Bengali, show alarmingly high rates (above $10\%$). We observe the presence of English named entities and occasional English noun references impacting the perplexity scores, suggesting
further room for improvement in our KenLM-based perplexity scoring. 

\textbf{4. Quality classifiers:}
Overall 
$3.40\%$ of the total outputs are flagged as low quality by the quality classifiers. Some languages like Malayalam exhibit disproportionately high low-quality rates, primarily due to frequent word/character repetitions and poor linguistic coherence. While the classifier-based filter complements the content filters for repetition, NSFW, and stopwords, and the perplexity-based fluency filter, its key limitation is the dependency on domain-specific training data. Incorporating new styles or source types necessitates retraining the classifier unlike the relatively low-cost heuristic and statistical filters.

\textbf{5. Bias detection:}
\label{subsec:bias-eval}
We evaluate our Hindi synthetic corpora across styles (\emph{e.g.}, textbook, blogpost, persona) for Indian sociolinguistic bias. 
For each style, we report WEAT effect sizes and scores, computed over 1M samples, using curated target-attribute word sets. The analysis reveals consistent medium to high stereotypical bias across sociocultural dimensions. Caste bias has effect sizes between $0.56$–$1.09$, highest in persona. Gender bias is most pronounced in story ($1.58$) and Redditpost ($1.21$) styles, with high bias in four of seven styles. Race bias scored above $1.0$ in most styles, peaking in blogpost ($1.51$), textbook ($1.46$), and Wikihow ($1.28$). Religion bias was similarly high in blogpost ($1.39$), textbook ($1.3$), and Wikihow ($1.73$) styles, indicating strong `Hindu–Muslim' stereotypes. Region/linguistic bias was present but weaker, 
with translation showing a reverse effect, suggesting mitigation. These findings indicate prevalent and measurable biases in synthetic generations, especially regarding religion, race, and gender. The complete results are provided in the Appendix, Table~\ref{tab:weat_bias_results}.

\textbf{6. Bias mitigation:}
\label{subsec:bias-mitigation}
We conduct a small-scale intervention targeting religious bias in Hindi textbook-style synthetic data. For around $20$ biased instances per target term (\emph{e.g.} Islamic and Hindu words), identified based on stereotypical co-occurrences, we replace them with LLM-based synthetically generated counter-stereotypical examples by reversing associations (\emph{e.g.}, Islam association with positive and Hindu with negative attributes). Retraining FastText embeddings on this modified corpus reduced the WEAT effect size and score from ($1.34$, $1.11$) to ($1.29$, $1.03$). 
This finding suggests that this targeted data augmentation is a scalable mitigation strategy in our synthetic generation pipeline. Detailed results are available in the Appendix Figure~\ref{fig:weat_anti_bias}.

\textbf{7. Bias comparison (Web vs. BhashaKritika):}
We leverage documents from the Web as context in our synthetic generation pipeline. In order to evaluate the inherent bias mitigation in our pipeline, we compute WEAT scores on the source Web documents and the corresponding generated synthetic data (Refer to Table~\ref{tab:weat_bias_source} in the Appendix). 
For instance, the religious bias in `Hindi textbook-style' data, with effect size and WEAT score of ($1.43$, $1.35$) in the source documents, dropped to ($1.14$, $0.93$) in the generated synthetic data. These results indicate that our synthetic data has lower biases compared to those in the source Web data, with targeted interventions, as described in the last section, further aiding the debiasing. Detailed association scores for individual target words and other bias dimensions are provided in the Appendix (Figures~\ref{fig:weat_bias_source_religion}-\ref{fig:weat_bias_source_race}).

\subsection{How does Synthetic Data Compare to Web Data for LLM Training?}
\begin{figure}
    \centering
\includegraphics[width=.9\columnwidth]{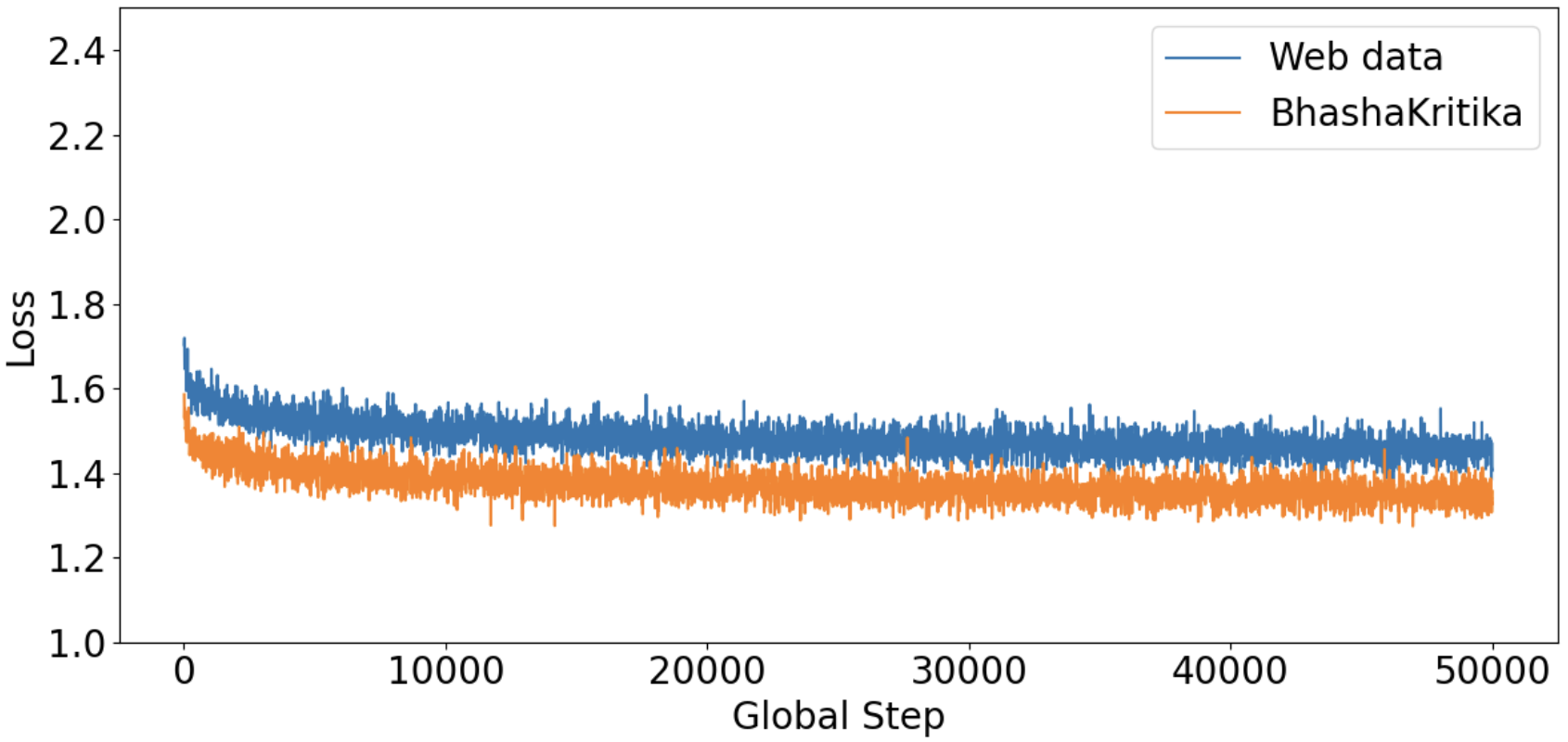} 
\caption{We annealed LLaMA-3.2 1B pretrained model on $50$B tokens of Web vs. our synthetic data - BhashaKritika. We observe faster convergence on BhashaKritika.}
    \label{fig:loss-anneal}
\end{figure}

In addition to intrinsic data quality evaluation, we also evaluate our synthetic data for LLM pretraining. Starting from the pretrained checkpoint of LLaMA-3.2 1B model, we perform annealing~\citep{grattafiori2024llama, allal2025smollm2,olmo20252olmo2furious}, where we linearly decay LR to $0$ over $50$B tokens of training data comprising $70\%$ Web, Math, and code data and $30\%$ Indic data. 
We train two models - $M_{Web}$ and $M_{BK}$ where the Indic data is sampled from the Web and BhashaKritika, our Indic synthetic corpus, respectively. We attribute the faster convergence of $M_{BK}$ (Fig.~\ref{fig:loss-anneal}) to the high-quality and knowledge-dense nature of our synthetic data while the Web data tends to be relatively noisy~\cite{abdin2024phi}. 
In Table~\ref{tab:eval-anneal}, we report the performance of these models on the English and Indic benchmarks. 
Further implementation details are provided in Appendix~\ref{sec:model_runs}.

\begin{table}[htbp]
\centering
\scriptsize  
\begin{tabular}{lrr}
\hline
\textbf{Dataset}  & \textbf{Web} & \textbf{BhashaKritika}\\
\hline
Hellaswag~\citep{zellers2019hellaswagmachinereallyfinish} & 0.483 & 0.482 \\
MMLU~\citep{hendrycks-etal-2021-measuring}                & 0.412 & 0.408 \\
OpenbookQA~\citep{mihaylov2018suitarmorconductelectricity} & 0.276 & 0.268 \\
GSM8K~\citep{cobbe2021trainingverifierssolvemath}         & 0.111 & \textbf{0.120} \\
DROP (F1)~\citep{dua2019dropreadingcomprehensionbenchmark}                                                  & 0.077 & \textbf{0.083} \\
TriviaQA~\citep{joshi2017triviaqalargescaledistantly}     & 0.398 & \textbf{0.404} \\
ARC Easy~\citep{clark2018thinksolvedquestionanswering}    & 0.734 & \textbf{0.741} \\
Winogrande~\citep{sakaguchi-etal-2021-winogrande}         & 0.616 & \textbf{0.628} \\
ARC Challenge~\citep{clark2018thinksolvedquestionanswering} & 0.406 & \textbf{0.408} \\
CommonsenseQA~\citep{talmor2019commonsenseqaquestionansweringchallenge} & 0.410 & \textbf{0.411} \\
\hdashline
Indic Sentiment~\citep{doddapaneni2023leavingindiclanguagebehind}     & 0.617 & \textbf{0.631} \\
Indic Copa~\citep{doddapaneni2023leavingindiclanguagebehind}          & 0.575 & \textbf{0.588} \\
ARC Challenge Indic~\citep{sarvamai_arcc_indic} & 0.225 & \textbf{0.225} \\
MILU~\citep{verma2024milu}                                                               & 0.283 & 0.282 \\
Indic XNLI~\citep{doddapaneni2023leavingindiclanguagebehind}         & 0.433 & 0.403 \\
Indic XParaphrase~\citep{doddapaneni2023leavingindiclanguagebehind}  & 0.782 & 0.736 \\
\hline
\end{tabular}
\caption{Evaluation results on English and Indic benchmarks for the LLaMA‑3.2 1B pre-trained model annealed on $50$B tokens of Web vs. BhashaKritika data. 
Results indicate that high-quality synthetic data can serve as an effective substitute for real-world data.}
\label{tab:eval-anneal}
\end{table}

\subsection{Can we Use Synthetic Data in Low Resource Settings?}

\begin{figure}
    \centering
\includegraphics[width=.9\columnwidth]{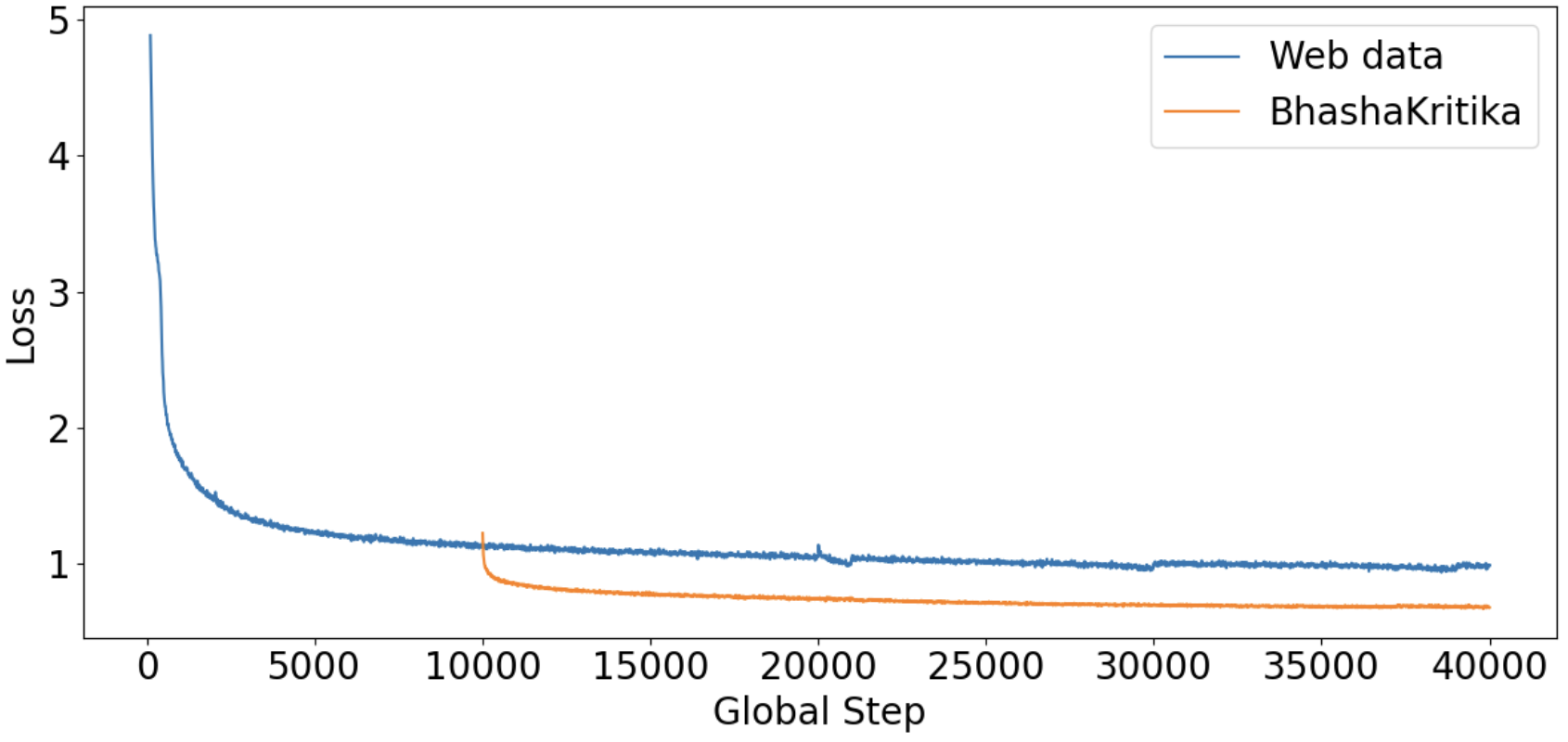} \caption{Loss curves for simulated low resource setting: LLaMA-3.2 1B is pretrained from scratch on $15$B Indic Web tokens ($10$K training steps) followed by continual training on - (1) same Web data; (2) BhashaKritika data 
}
\label{fig:loss-fromscratch}
\end{figure}

A key challenge in building models for Indian languages is the limited availability of high-quality data. We explore whether our BhashaKritika corpus could serve as a good pretraining data in these low resource settings by conducting a 
controlled experiment. We pretrain LLaMA-3.2 1B model from scratch on a fixed budget of $15$B tokens of Indic Web data ($10$K training steps in Fig.~\ref{fig:loss-fromscratch}). Starting from this base model, we continually pretrain $M_{Web}$ for 3 more epochs on the same Indic Web data and $M_{BK}$ on data sampled from our BhashaKritika synthetic corpus.

The model trained on our Indic synthetic data converges faster and shows a similar or better performance on Indic benchmarks (Table~\ref{tab:eval-scratch}). 
This indicates that high-quality synthetic data can serve as a viable substitute when Web data is limited, offering a promising direction for low-resource language settings.

\begin{table}[htbp]
\centering
\scriptsize  
\begin{tabular}{lrr}
\hline
\textbf{Dataset}  & \textbf{Web} & \textbf{BhashaKritika}\\
\hline
Indic Sentiment~\citep{doddapaneni2023leavingindiclanguagebehind}     & 0.491 &	\textbf{0.499} \\
MILU~\citep{verma2024milu}                                                             & 0.235	& \textbf{0.236} \\
Indic Copa~\citep{doddapaneni2023leavingindiclanguagebehind}          & 0.509 &	\textbf{0.512} \\
Indic XParaphrase~\citep{doddapaneni2023leavingindiclanguagebehind}  & 0.499&	\textbf{0.500} \\
Indic XNLI~\citep{doddapaneni2023leavingindiclanguagebehind}         & 0.330&	\textbf{0.339} \\
ARC Challenge Indic~\citep{doddapaneni2023leavingindiclanguagebehind} & 0.213&	0.210 \\

\hline
\end{tabular}
\caption{Benchmark comparison on Indic datasets; $1$B model pretrained on $15$B tokens of Indic Web data from scratch is continually pretrained on multiple epochs of the same Web data vs BhashaKritika data.}
\label{tab:eval-scratch}
\end{table}

\section{Conclusion}
\label{sec:conclusion}
We introduced \textit{BhashaKritika}, a $540$B tokens high-quality Indic synthetic corpus across 10 languages and different knowledge-dense styles. The data is generated using our scalable synthetic generation pipeline comprising multiple data sources, five generation approaches, multilingual LLMs, and translation models. 
We show that by careful selection of models per language and using Indic documents, topics 
and personas for grounding, we can synthetically generate high-quality Indic data. We demonstrate that using English instructions alongside Indic source texts yields better quality outputs and also introduce a novel technique to create math and reasoning focused data.  
Further, we introduce a comprehensive automated quality evaluation pipeline to ensure quality of the generated data. 
Through extensive analysis and empirical runs, we show the efficacy of our synthetically generated data, opening up avenues 
to augment the pretraining dataset for the low resource Indic languages.
\section{Acknowledgements}
We thank the leadership at Krutrim for their support in carrying out this research. We also thank the Data Annotation Team for their meticulous efforts in evaluation. We also thank the anonymous reviewers for their valuable feedback and suggestions.

\bibliography{iclr2025_conference}

@article{yang2025scaling,
      title={Scaling Text-Rich Image Understanding via Code-Guided Synthetic Multimodal Data Generation},
      author={Yang, Yue and Patel, Ajay and Deitke, Matt and Gupta, Tanmay and Weihs, Luca and Head, Andrew and Yatskar, Mark and Callison-Burch, Chris and Krishna, Ranjay and Kembhavi, Aniruddha and others},
      journal={arXiv preprint arXiv:2502.14846},
      year={2025}
}

@article{chowdhery2022palm,
  title={PaLM: Scaling Language Modeling with Pathways},
  author={Chowdhery, Aakanksha and others},
  journal={arXiv preprint arXiv:2204.02311},
  year={2022}
}

@article{touvron2023llama,
  title={LLaMA: Open and Efficient Foundation Language Models},
  author={Touvron, Hugo and others},
  journal={arXiv preprint arXiv:2302.13971},
  year={2023}
}

@article{joshi2020state,
  title={The state and fate of linguistic diversity and inclusion in the NLP world},
  author={Joshi, Pratik and Santy, Sebastin and Budhiraja, Amar and Bali, Kalika and Choudhury, Monojit},
  journal={arXiv preprint arXiv:2004.09095},
  year={2020}
}

@article{mueller2022crosslingual,
  title={Is ChatGPT multilingual? Evaluating zero-shot performance across 50 languages},
  author={Mueller, Aaron and others},
  journal={arXiv preprint arXiv:2304.09157},
  year={2023}
}

@article{conneau2020unsupervised,
  title={Unsupervised Cross-lingual Representation Learning at Scale},
  author={Conneau, Alexis and others},
  journal={ACL},
  year={2020}
}

@misc{selfinstruct2022,
  title={Self-Instruct: Aligning Language Models with Self-Generated Instructions},
  author={Wang, Yizhong and others},
  year={2022},
  howpublished={arXiv preprint arXiv:2212.10560}
}

@misc{taori2023stanford,
  title={Stanford alpaca: An instruction-following llama model},
  author={Taori, Rohan and Gulrajani, Ishaan and Zhang, Tianyi and Dubois, Yann and Li, Xuechen and Guestrin, Carlos and Liang, Percy and Hashimoto, Tatsunori B},
  year={2023},
  publisher={Stanford, CA, USA}
}

@inproceedings{longpre2023flan,
  title={The flan collection: Designing data and methods for effective instruction tuning},
  author={Longpre, Shayne and Hou, Le and Vu, Tu and Webson, Albert and Chung, Hyung Won and Tay, Yi and Zhou, Denny and Le, Quoc V and Zoph, Barret and Wei, Jason and others},
  booktitle={ICML},
  year={2023}
}

@article{chen2023synthia,
  title={SYNTHIA: Synthetic Instruction Data for Zero-Shot Cross-Task Generalization},
  author={Chen, Andy and others},
  journal={ACL},
  year={2023}
}

@article{ge2024scaling,
  title={Scaling synthetic data creation with 1,000,000,000 personas},
  author={Ge, Tao and Chan, Xin and Wang, Xiaoyang and Yu, Dian and Mi, Haitao and Yu, Dong},
  journal={arXiv preprint arXiv:2406.20094},
  year={2024}
}

@misc{meyer2025nemotron,
  author = {Meyer, Yev and Corneil, Dane},
  title = {{Nemotron-Personas}: Synthetic Personas Aligned to Real-World Distributions
},
  month = {June},
  year = {2025},
  url = {https://huggingface.co/datasets/nvidia/Nemotron-Personas}
}

@article{penedo2024fineweb,
  title={{T}he {F}ine{W}eb {D}atasets: {D}ecanting the {W}eb for the {F}inest {T}ext {D}ata at {S}cale},
  author={Penedo, Guilherme and Kydl{\'\i}{\v{c}}ek, Hynek and Lozhkov, Anton and Mitchell, Margaret and Raffel, Colin and Von Werra, Leandro and Wolf, Thomas and others},
  journal={arXiv preprint arXiv:2406.17557},
  year={2024}
}

@misc{penedo2024fineweb2,
  author = {Penedo, Guilherme and Kydlíček, Hynek and Sabolčec, Vinko and Messmer, Bettina and Foroutan, Negar and Jaggi, Martin and von Werra, Leandro and Wolf, Thomas},
  title = {{F}ine{W}eb2: {A} sparkling update with 1000s of languages},
  month = dec,
  year = 2024,
  doi = { 10.57967/hf/3744 },
  url = {https://huggingface.co/datasets/HuggingFaceFW/fineweb-2},
  note = {Accessed 30 Jan. 2025}
}

@article{odumakinde2024multilingual,
  title={Multilingual arbitrage: Optimizing data pools to accelerate multilingual progress},
  author={Odumakinde, Ayomide and D'souza, Daniel and Verga, Pat and Ermis, Beyza and Hooker, Sara},
  journal={arXiv preprint arXiv:2408.14960},
  year={2024}
}

@inproceedings{heafield2011kenlm,
  title={KenLM: Faster and smaller language model queries},
  author={Heafield, Kenneth},
  booktitle={Proceedings of the sixth workshop on statistical machine translation},
  pages={187--197},
  year={2011}
}

@article{suryanarayanan2024pralekha,
  title={Pralekha: An Indic Document Alignment Evaluation Benchmark},
  author={Suryanarayanan, Sanjay and Song, Haiyue and Khan, Mohammed Safi Ur Rahman and Kunchukuttan, Anoop and Khapra, Mitesh M and Dabre, Raj},
  journal={arXiv preprint arXiv:2411.19096},
  year={2024}
}

@article{joulin2016fasttext,
  title={FastText.zip: Compressing text classification models},
  author={Joulin, Armand and Grave, Edouard and Bojanowski, Piotr and Douze, Matthijs and J{\'e}gou, H{\'e}rve and Mikolov, Tomas},
  journal={arXiv preprint arXiv:1612.03651},
  year={2016}
}

@article{muennighoff2023scaling,
  title={Scaling data-constrained language models},
  author={Muennighoff, Niklas and Rush, Alexander and Barak, Boaz and Le Scao, Teven and Tazi, Nouamane and Piktus, Aleksandra and Pyysalo, Sampo and Wolf, Thomas and Raffel, Colin A},
  journal={Advances in Neural Information Processing Systems},
  volume={36},
  pages={50358--50376},
  year={2023}
}

@article{abdin2025phi,
  title={Phi-4-reasoning technical report},
  author={Abdin, Marah and Agarwal, Sahaj and Awadallah, Ahmed and Balachandran, Vidhisha and Behl, Harkirat and Chen, Lingjiao and de Rosa, Gustavo and Gunasekar, Suriya and Javaheripi, Mojan and Joshi, Neel and others},
  journal={arXiv preprint arXiv:2504.21318},
  year={2025}
}

@article{lewis2020retrieval,
  title={Retrieval-augmented generation for knowledge-intensive nlp tasks},
  author={Lewis, Patrick and Perez, Ethan and Piktus, Aleksandra and Petroni, Fabio and Karpukhin, Vladimir and Goyal, Naman and K{\"u}ttler, Heinrich and Lewis, Mike and Yih, Wen-tau and Rockt{\"a}schel, Tim and others},
  journal={Advances in neural information processing systems},
  volume={33},
  pages={9459--9474},
  year={2020}
}

@inproceedings{chen2025revisiting,
  title={Revisiting Scaling Laws for Language Models: The Role of Data Quality and Training Strategies},
  author={Chen, Zhengyu and Wang, Siqi and Xiao, Teng and Wang, Yudong and Chen, Shiqi and Cai, Xunliang and He, Junxian and Wang, Jingang},
  booktitle={Proceedings of the 63rd Annual Meeting of the Association for Computational Linguistics (Volume 1: Long Papers)},
  pages={23881--23899},
  year={2025}
}

@article{chang2024scaling,
  title={Scaling Parameter-Constrained Language Models with Quality Data},
  author={Chang, Ernie and Paltenghi, Matteo and Li, Yang and Lin, Pin-Jie and Zhao, Changsheng and Huber, Patrick and Liu, Zechun and Rabatin, Rastislav and Shi, Yangyang and Chandra, Vikas},
  journal={arXiv preprint arXiv:2410.03083},
  year={2024}
}

@article{wang2022self,
  title={Self-instruct: Aligning language models with self-generated instructions},
  author={Wang, Yizhong and Kordi, Yeganeh and Mishra, Swaroop and Liu, Alisa and Smith, Noah A and Khashabi, Daniel and Hajishirzi, Hannaneh},
  journal={arXiv preprint arXiv:2212.10560},
  year={2022}
}

@article{xu2023wizardlm,
  title={Wizardlm: Empowering large language models to follow complex instructions},
  author={Xu, Can and Sun, Qingfeng and Zheng, Kai and Geng, Xiubo and Zhao, Pu and Feng, Jiazhan and Tao, Chongyang and Jiang, Daxin},
  journal={arXiv preprint arXiv:2304.12244},
  year={2023}
}

@misc{xu2024magpie,
    title={Magpie: Alignment Data Synthesis from Scratch by Prompting Aligned LLMs with Nothing}, 
    author={Zhangchen Xu and Fengqing Jiang and Luyao Niu and Yuntian Deng and Radha Poovendran and Yejin Choi and Bill Yuchen Lin},
    year={2024},
    eprint={2406.08464},
    archivePrefix={arXiv},
    primaryClass={cs.CL}
}

@article{grattafiori2024llama,
  title={The llama 3 herd of models},
  author={Grattafiori, Aaron and Dubey, Abhimanyu and Jauhri, Abhinav and Pandey, Abhinav and Kadian, Abhishek and Al-Dahle, Ahmad and Letman, Aiesha and Mathur, Akhil and Schelten, Alan and Vaughan, Alex and others},
  journal={arXiv preprint arXiv:2407.21783},
  year={2024}
}

@article{allal2025smollm2,
  title={SmolLM2: When Smol Goes Big--Data-Centric Training of a Small Language Model},
  author={Allal, Loubna Ben and Lozhkov, Anton and Bakouch, Elie and Bl{\'a}zquez, Gabriel Mart{\'\i}n and Penedo, Guilherme and Tunstall, Lewis and Marafioti, Andr{\'e}s and Kydl{\'\i}{\v{c}}ek, Hynek and Lajar{\'\i}n, Agust{\'\i}n Piqueres and Srivastav, Vaibhav and others},
  journal={arXiv preprint arXiv:2502.02737},
  year={2025}
}

@article{gunasekar2023textbooks,
  title={Textbooks are all you need},
  author={Gunasekar, Suriya and Zhang, Yi and Aneja, Jyoti and Mendes, Caio C{\'e}sar Teodoro and Del Giorno, Allie and Gopi, Sivakanth and Javaheripi, Mojan and Kauffmann, Piero and de Rosa, Gustavo and Saarikivi, Olli and others},
  journal={arXiv preprint arXiv:2306.11644},
  year={2023}
}

@article{li2023textbooks,
  title={Textbooks are all you need ii: phi-1.5 technical report},
  author={Li, Yuanzhi and Bubeck, S{\'e}bastien and Eldan, Ronen and Del Giorno, Allie and Gunasekar, Suriya and Lee, Yin Tat},
  journal={arXiv preprint arXiv:2309.05463},
  year={2023}
}

@article{abdin2024phi,
  title={Phi-4 technical report},
  author={Abdin, Marah and Aneja, Jyoti and Behl, Harkirat and Bubeck, S{\'e}bastien and Eldan, Ronen and Gunasekar, Suriya and Harrison, Michael and Hewett, Russell J and Javaheripi, Mojan and Kauffmann, Piero and others},
  journal={arXiv preprint arXiv:2412.08905},
  year={2024}
}

@article{jiang2024mixtral,
  title={Mixtral of experts},
  author={Jiang, Albert Q and Sablayrolles, Alexandre and Roux, Antoine and Mensch, Arthur and Savary, Blanche and Bamford, Chris and Chaplot, Devendra Singh and Casas, Diego de las and Hanna, Emma Bou and Bressand, Florian and others},
  journal={arXiv preprint arXiv:2401.04088},
  year={2024}
}

@article{bendale2024sutra,
  title={SUTRA: Scalable Multilingual Language Model Architecture},
  author={Bendale, Abhijit and Sapienza, Michael and Ripplinger, Steven and Gibbs, Simon and Lee, Jaewon and Mistry, Pranav},
  journal={arXiv preprint arXiv:2405.06694},
  year={2024}
}

@article{team2024krutrim,
  title={Krutrim {LLM}: Multilingual Foundational Model for over a Billion People},
  author={Team Krutrim},
  journal={Under Review},
  year={2024}
}

@misc{kallappa2025krutrim,
  title={Krutrim LLM: Multilingual Foundational Model for over a Billion People},
  author={Aditya Kallappa and Palash Kamble and Abhinav Ravi and Akshat Patidar and Vinayak Dhruv and Deepak Kumar and Raghav Awasthi and Arveti Manjunath and Shubham Agarwal and Kumar Ashish and Gautam Bhargava and Chandra Khatri},
  year={2025},
  eprint={2502.09642},
  archivePrefix={arXiv},
  primaryClass={cs.CL}
}

@misc{zhou2025megamathpushinglimitsopen,
      title={MegaMath: Pushing the Limits of Open Math Corpora}, 
      author={Fan Zhou and Zengzhi Wang and Nikhil Ranjan and Zhoujun Cheng and Liping Tang and Guowei He and Zhengzhong Liu and Eric P. Xing},
      year={2025},
      eprint={2504.02807},
      archivePrefix={arXiv},
      primaryClass={cs.CL},
      url={https://arxiv.org/abs/2504.02807}, 
}

@misc{numina_math_datasets,
  author = {Jia Li and Edward Beeching and Lewis Tunstall and Ben Lipkin and Roman Soletskyi and Shengyi Costa Huang and Kashif Rasul and Longhui Yu and Albert Jiang and Ziju Shen and Zihan Qin and Bin Dong and Li Zhou and Yann Fleureau and Guillaume Lample and Stanislas Polu},
  title = {NuminaMath},
  year = {2024},
  publisher = {Numina},
  journal = {Hugging Face repository},
  howpublished = {\url{[https://huggingface.co/AI-MO/NuminaMath-1.5](https://github.com/project-numina/aimo-progress-prize/blob/main/report/numina_dataset.pdf)}}
}

@article{moshkov2025aimo2,
  title   = {AIMO-2 Winning Solution: Building State-of-the-Art Mathematical Reasoning Models with OpenMathReasoning dataset},
  author  = {Ivan Moshkov and Darragh Hanley and Ivan Sorokin and Shubham Toshniwal and Christof Henkel and Benedikt Schifferer and Wei Du and Igor Gitman},
  year    = {2025},
  journal = {arXiv preprint arXiv:2504.16891}
}

@misc{maini2024rephrasingwebrecipecompute,
      title={Rephrasing the Web: A Recipe for Compute and Data-Efficient Language Modeling}, 
      author={Pratyush Maini and Skyler Seto and He Bai and David Grangier and Yizhe Zhang and Navdeep Jaitly},
      year={2024},
      eprint={2401.16380},
      archivePrefix={arXiv},
      primaryClass={cs.CL},
      url={https://arxiv.org/abs/2401.16380}, 
}

@article{gala2024airavata,
  title   = {Airavata: Introducing Hindi Instruction-tuned LLM},
  author  = {Jay Gala and Thanmay Jayakumar and Jaavid Aktar Husain and Aswanth Kumar M and Mohammed Safi Ur Rahman Khan and Diptesh Kanojia and Ratish Puduppully and Mitesh M. Khapra and Raj Dabre and Rudra Murthy and Anoop Kunchukuttan},
  year    = {2024},
  journal = {arXiv preprint arXiv: 2401.15006}
}

@misc{balachandran2023tamilllama,
      title={Tamil-Llama: A New Tamil Language Model Based on Llama 2}, 
      author={Abhinand Balachandran},
      year={2023},
      eprint={2311.05845},
      archivePrefix={arXiv},
      primaryClass={cs.CL}
}

@misc{kohli2023building,
      title={Building a Llama2-finetuned LLM for Odia Language Utilizing Domain Knowledge Instruction Set}, 
      author={Guneet Singh Kohli and Shantipriya Parida and Sambit Sekhar and Samirit Saha and Nipun B Nair and Parul Agarwal and Sonal Khosla and Kusumlata Patiyal and Debasish Dhal},
      year={2023},
      eprint={2312.12624},
      archivePrefix={arXiv},
      primaryClass={cs.CL}
}

@article{verma2024milu,
  title   = {MILU: A Multi-task Indic Language Understanding Benchmark},
  author  = {Sshubam Verma and Mohammed Safi Ur Rahman Khan and Vishwajeet Kumar and Rudra Murthy and Jaydeep Sen},
  year    = {2024},
  journal = {arXiv preprint arXiv: 2411.02538}
}

@misc{dua2019dropreadingcomprehensionbenchmark,
      title={DROP: A Reading Comprehension Benchmark Requiring Discrete Reasoning Over Paragraphs}, 
      author={Dheeru Dua and Yizhong Wang and Pradeep Dasigi and Gabriel Stanovsky and Sameer Singh and Matt Gardner},
      year={2019},
      eprint={1903.00161},
      archivePrefix={arXiv},
      primaryClass={cs.CL},
      url={https://arxiv.org/abs/1903.00161}, 
}

@misc{nanda2024choudhury,
title={Llama-3-Nanda-10B-Chat: An Open Generative Large Language Model for Hindi},
author={Choudhury, Monojit and Chauhan, Shivam and others},
      year={2024}

}

@inproceedings{kakwani-etal-2020-indicnlpsuite,
    title = "{I}ndic{NLPS}uite: Monolingual Corpora, Evaluation Benchmarks and Pre-trained Multilingual Language Models for {I}ndian Languages",
    author = "Kakwani, Divyanshu  and
      Kunchukuttan, Anoop  and
      Golla, Satish  and
      N.C., Gokul  and
      Bhattacharyya, Avik  and
      Khapra, Mitesh M.  and
      Kumar, Pratyush",
    booktitle = "Findings of the Association for Computational Linguistics: EMNLP 2020",
    month = nov,
    year = "2020",
    address = "Online",
    publisher = "Association for Computational Linguistics",
    url = "https://aclanthology.org/2020.findings-emnlp.445",
    doi = "10.18653/v1/2020.findings-emnlp.445",
    pages = "4948--4961",
}

@misc{sarvam2023openhathi,
  title = {OpenHathi Series: An Approach To Build Bilingual LLMs Frugally},
  url = {https://www.sarvam.ai/blog/announcing-openhathi-series},
  author = {Sarvam},
  month = {December},
  year = {2023}
}

@misc{sarvam2024llm,
  title = {Sarvam AI launches first LLM for Indian languages},
  url = {https://www.sarvam.ai/blogs/sarvam-nvidia},
  author = {Sarvam},
  month = {October},
  year = {2024}
}

@article{khan2024indicllmsuite,
  title={IndicLLMSuite: A Blueprint for Creating Pre-training and Fine-Tuning Datasets for Indian Languages},
  author={Khan, Mohammed Safi Ur Rahman and Mehta, Priyam and Sankar, Ananth and Kumaravelan, Umashankar and Doddapaneni, Sumanth and Jain, Sparsh and Kunchukuttan, Anoop and Kumar, Pratyush and Dabre, Raj and Khapra, Mitesh M and others},
  journal={arXiv preprint arXiv:2403.06350},
  year={2024}
}

@article{kunchukuttan2020indicnlpcorpus,
    title={AI4Bharat-IndicNLP Corpus: Monolingual Corpora and Word Embeddings for Indic Languages},
    author={Anoop Kunchukuttan and Divyanshu Kakwani and Satish Golla and Gokul N.C. and Avik Bhattacharyya and Mitesh M. Khapra and Pratyush Kumar},
    year={2020},
    journal={arXiv preprint arXiv:2005.00085},
}

@article{qin2025scaling,
  title={Scaling laws of synthetic data for language models},
  author={Qin, Zeyu and Dong, Qingxiu and Zhang, Xingxing and Dong, Li and Huang, Xiaolong and Yang, Ziyi and Khademi, Mahmoud and Zhang, Dongdong and Awadalla, Hany Hassan and Fung, Yi R and others},
  journal={arXiv preprint arXiv:2503.19551},
  year={2025}
}

@article{Gao2020ThePA,
  title={The Pile: An 800GB Dataset of Diverse Text for Language Modeling},
  author={Leo Gao and Stella Biderman and Sid Black and Laurence Golding and Travis Hoppe and Charles Foster and Jason Phang and Horace He and Anish Thite and Noa Nabeshima and Shawn Presser and Connor Leahy},
  journal={ArXiv},
  year={2020},
  volume={abs/2101.00027},
  url={https://api.semanticscholar.org/CorpusID:230435736}
}

@article{raffel2020exploring,
  title={Exploring the limits of transfer learning with a unified text-to-text transformer},
  author={Raffel, Colin and Shazeer, Noam and Roberts, Adam and Lee, Katherine and Narang, Sharan and Matena, Michael and Zhou, Yanqi and Li, Wei and Liu, Peter J},
  journal={Journal of machine learning research},
  volume={21},
  number={140},
  pages={1--67},
  year={2020}
}

@misc{commoncrawl2007,
  author       = {{Common Crawl}},
  title        = {Common Crawl - Open Repository of Web Crawl Data},
  year         = {2007},
  url          = {https://commoncrawl.org/}
}

@misc{together2023redpajama,
  author = {Together Computer},
  title = {RedPajama: an Open Dataset for Training Large Language Models},
  year = 2023,
  url = {https://github.com/togethercomputer/RedPajama-Data}
}

@article{Penedo2023TheRD,
  title={The RefinedWeb Dataset for Falcon LLM: Outperforming Curated Corpora with Web Data, and Web Data Only},
  author={Guilherme Penedo and Quentin Malartic and Daniel Hesslow and Ruxandra-Aim{\'e}e Cojocaru and Alessandro Cappelli and Hamza Alobeidli and Baptiste Pannier and Ebtesam Almazrouei and Julien Launay},
  journal={ArXiv},
  year={2023},
  volume={abs/2306.01116},
  url={https://api.semanticscholar.org/CorpusID:259063761}
}

@article{Soldaini2024DolmaAO,
  title={Dolma: an Open Corpus of Three Trillion Tokens for Language Model Pretraining Research},
  author={Luca Soldaini and Rodney Kinney and Akshita Bhagia and Dustin Schwenk and David Atkinson and Russell Authur and Ben Bogin and Khyathi Raghavi Chandu and Jennifer Dumas and Yanai Elazar and Valentin Hofmann and A. Jha and Sachin Kumar and Li Lucy and others},
  journal={ArXiv},
  year={2024},
  volume={abs/2402.00159},
  url={https://api.semanticscholar.org/CorpusID:267364861}
}

@misc{olmo20252olmo2furious,
      title={2 OLMo 2 Furious}, 
      author={Team OLMo and Pete Walsh and Luca Soldaini and Dirk Groeneveld and Kyle Lo and Shane Arora and Akshita Bhagia and Yuling Gu and Shengyi Huang and Matt Jordan and Nathan Lambert and Dustin Schwenk and Oyvind Tafjord and Taira Anderson and David Atkinson and Faeze Brahman and Christopher Clark and Pradeep Dasigi and Nouha Dziri and Michal Guerquin and Hamish Ivison and Pang Wei Koh and Jiacheng Liu and Saumya Malik and William Merrill and Lester James V. Miranda and Jacob Morrison and Tyler Murray and Crystal Nam and Valentina Pyatkin and Aman Rangapur and Michael Schmitz and Sam Skjonsberg and David Wadden and Christopher Wilhelm and Michael Wilson and Luke Zettlemoyer and Ali Farhadi and Noah A. Smith and Hannaneh Hajishirzi},
      year={2025},
      eprint={2501.00656},
      archivePrefix={arXiv},
      primaryClass={cs.CL},
      url={https://arxiv.org/abs/2501.00656}, 
}

@article{Li2024DataCompLMIS,
  title={DataComp-LM: In search of the next generation of training sets for language models},
  author={Jeffrey Li and Alex Fang and Georgios Smyrnis and Maor Ivgi and Matt Jordan and Samir Yitzhak Gadre and Hritik Bansal and Etash Kumar Guha and Sedrick Scott Keh and Kushal Arora and Saurabh Garg and Rui Xin and Niklas Muennighoff and Reinhard Heckel and Jean-Pierre Mercat and Mayee Chen and Suchin Gururangan and Mitchell Wortsman and Alon Albalak and Yonatan Bitton and Marianna Nezhurina and Amro Abbas and Cheng-Yu Hsieh and Dhruba Ghosh and Josh Gardner and Maciej Kilian and Hanlin Zhang and Rulin Shao and Sarah Pratt and Sunny Sanyal and Gabriel Ilharco and Giannis Daras and Kalyani Marathe and Aaron Gokaslan and Jieyu Zhang and Khyathi Chandu and Thao Nguyen and Igor Vasiljevic and Sham M. Kakade and Shuran Song and Sujay Sanghavi and Fartash Faghri and Sewoong Oh and Luke Zettlemoyer and Kyle Lo and Alaaeldin El-Nouby and Hadi Pouransari and Alexander Toshev and Stephanie Wang and Dirk Groeneveld and Luca Soldani and Pang Wei Koh and Jenia Jitsev and Thomas Kollar and Alexandros G. Dimakis and Yair Carmon and Achal Dave and Ludwig Schmidt and Vaishaal Shankar},
  journal={ArXiv},
  year={2024},
  volume={abs/2406.11794},
  url={https://api.semanticscholar.org/CorpusID:270560330}
}

@article{su2024nemotron,
  title={Nemotron-CC: Transforming Common Crawl into a Refined Long-Horizon Pretraining Dataset},
  author={Su, Dan and Kong, Kezhi and Lin, Ying and Jennings, Joseph and Norick, Brandon and Kliegl, Markus and Patwary, Mostofa and Shoeybi, Mohammad and Catanzaro, Bryan},
  journal={arXiv preprint arXiv:2412.02595},
  year={2024}
}

@misc{benallal2024cosmopedia,
  author = {Ben Allal, Loubna and Lozhkov, Anton and Penedo, Guilherme and Wolf, Thomas and von Werra, Leandro},
  title = {Cosmopedia},
  year = {2024},
  url = {https://huggingface.co/datasets/HuggingFaceTB/cosmopedia}
}

@article{agarwal2025language,
  title={Language Specific Knowledge: Do Models Know Better in X than in English?},
  author={Agarwal, Ishika and Bozdag, Nimet Beyza and Hakkani-T{\"u}r, Dilek},
  journal={arXiv preprint arXiv:2505.14990},
  year={2025}
}

@article{lambert2024tulu,
  title={T$\backslash$" ulu 3: Pushing frontiers in open language model post-training},
  author={Lambert, Nathan and Morrison, Jacob and Pyatkin, Valentina and Huang, Shengyi and Ivison, Hamish and Brahman, Faeze and Miranda, Lester James V and Liu, Alisa and Dziri, Nouha and Lyu, Shane and others},
  journal={arXiv preprint arXiv:2411.15124},
  year={2024}
}

@inproceedings{jentzsch2019semantics,
  title={Semantics derived automatically from language corpora contain human-like moral choices},
  author={Jentzsch, Sophie and Schramowski, Patrick and Rothkopf, Constantin and Kersting, Kristian},
  booktitle={Proceedings of the 2019 AAAI/ACM Conference on AI, Ethics, and Society},
  pages={37--44},
  year={2019}
}

@inproceedings{kwon2023efficient,
  title={Efficient Memory Management for Large Language Model Serving with PagedAttention},
  author={Woosuk Kwon and Zhuohan Li and Siyuan Zhuang and Ying Sheng and Lianmin Zheng and Cody Hao Yu and Joseph E. Gonzalez and Hao Zhang and Ion Stoica},
  booktitle={Proceedings of the ACM SIGOPS 29th Symposium on Operating Systems Principles},
  year={2023}
}

@inproceedings{
akter2025mind,
title={{MIND}: Math Informed syNthetic Dialogues for Pretraining {LLM}s},
author={Syeda Nahida Akter and Shrimai Prabhumoye and John Kamalu and Sanjeev Satheesh and Eric Nyberg and Mostofa Patwary and Mohammad Shoeybi and Bryan Catanzaro},
booktitle={The Thirteenth International Conference on Learning Representations},
year={2025},
url={https://openreview.net/forum?id=TuOTSAiHDn}
}

@article{villalobos2022will,
  title={Will we run out of data? Limits of LLM scaling based on human-generated data},
  author={Villalobos, Pablo and Ho, Anson and Sevilla, Jaime and Besiroglu, Tamay and Heim, Lennart and Hobbhahn, Marius},
  journal={arXiv preprint arXiv:2211.04325},
  year={2022}
}

@article{nadas2025synthetic,
  title={Synthetic data generation using large language models: Advances in text and code},
  author={Nadas, Mihai and Diosan, Laura and Tomescu, Andreea},
  journal={arXiv preprint arXiv:2503.14023},
  year={2025}
}

@article{liu2024best,
  title={Best practices and lessons learned on synthetic data for language models},
  author={Liu, Ruibo and Wei, Jerry and Liu, Fangyu and Si, Chenglei and Zhang, Yanzhe and Rao, Jinmeng and Zheng, Steven and Peng, Daiyi and Yang, Diyi and Zhou, Denny and others},
  journal={CoRR},
  year={2024}
}

@article{yu2023large,
  title={Large language model as attributed training data generator: A tale of diversity and bias},
  author={Yu, Yue and Zhuang, Yuchen and Zhang, Jieyu and Meng, Yu and Ratner, Alexander J and Krishna, Ranjay and Shen, Jiaming and Zhang, Chao},
  journal={Advances in neural information processing systems},
  volume={36},
  pages={55734--55784},
  year={2023}
}

@misc{eval-harness,
  author       = {Gao, Leo and Tow, Jonathan and Abbasi, Baber and Biderman, Stella and Black, Sid and DiPofi, Anthony and Foster, Charles and Golding, Laurence and Hsu, Jeffrey and Le Noac'h, Alain and Li, Haonan and McDonell, Kyle and Muennighoff, Niklas and Ociepa, Chris and Phang, Jason and Reynolds, Laria and Schoelkopf, Hailey and Skowron, Aviya and Sutawika, Lintang and Tang, Eric and Thite, Anish and Wang, Ben and Wang, Kevin and Zou, Andy},
  title        = {The Language Model Evaluation Harness},
  month        = 07,
  year         = 2024,
  publisher    = {Zenodo},
  version      = {v0.4.3},
  doi          = {10.5281/zenodo.12608602},
  url          = {https://zenodo.org/records/12608602}
}

@article{sakaguchi-etal-2021-winogrande,
    author = {Sakaguchi, Keisuke and Bras, Ronan Le and Bhagavatula, Chandra and Choi, Yejin},
    title = {WinoGrande: an adversarial winograd schema challenge at scale},
    year = {2021},
    issue_date = {September 2021},
    publisher = {Association for Computing Machinery},
    address = {New York, NY, USA},
    volume = {64},
    number = {9},
    issn = {0001-0782},
    url = {https://doi.org/10.1145/3474381},
    journal = {Commun. ACM},
    month = aug,
    pages = {99–106},
}

@inproceedings{hendrycks-etal-2021-measuring,
    title={Measuring Massive Multitask Language Understanding},
    author={Dan Hendrycks and Collin Burns and Steven Basart and Andy Zou and Mantas Mazeika and Dawn Song and Jacob Steinhardt},
    booktitle={International Conference on Learning Representations},
    year={2021},
    url={https://openreview.net/forum?id=d7KBjmI3GmQ}
}

@misc{joshi2017triviaqalargescaledistantly,
      title={TriviaQA: A Large Scale Distantly Supervised Challenge Dataset for Reading Comprehension}, 
      author={Mandar Joshi and Eunsol Choi and Daniel S. Weld and Luke Zettlemoyer},
      year={2017},
      eprint={1705.03551},
      archivePrefix={arXiv},
      primaryClass={cs.CL},
      url={https://arxiv.org/abs/1705.03551}, 
}

@misc{zellers2019hellaswagmachinereallyfinish,
      title={HellaSwag: Can a Machine Really Finish Your Sentence?}, 
      author={Rowan Zellers and Ari Holtzman and Yonatan Bisk and Ali Farhadi and Yejin Choi},
      year={2019},
      eprint={1905.07830},
      archivePrefix={arXiv},
      primaryClass={cs.CL},
      url={https://arxiv.org/abs/1905.07830}, 
}

@misc{cobbe2021trainingverifierssolvemath,
      title={Training Verifiers to Solve Math Word Problems}, 
      author={Karl Cobbe and Vineet Kosaraju and Mohammad Bavarian and Mark Chen and Heewoo Jun and Lukasz Kaiser and Matthias Plappert and Jerry Tworek and Jacob Hilton and Reiichiro Nakano and Christopher Hesse and John Schulman},
      year={2021},
      eprint={2110.14168},
      archivePrefix={arXiv},
      primaryClass={cs.LG},
      url={https://arxiv.org/abs/2110.14168}, 
}

@misc{clark2018thinksolvedquestionanswering,
      title={Think you have Solved Question Answering? Try ARC, the AI2 Reasoning Challenge}, 
      author={Peter Clark and Isaac Cowhey and Oren Etzioni and Tushar Khot and Ashish Sabharwal and Carissa Schoenick and Oyvind Tafjord},
      year={2018},
      eprint={1803.05457},
      archivePrefix={arXiv},
      primaryClass={cs.AI},
      url={https://arxiv.org/abs/1803.05457}, 
}

@misc{talmor2019commonsenseqaquestionansweringchallenge,
      title={CommonsenseQA: A Question Answering Challenge Targeting Commonsense Knowledge}, 
      author={Alon Talmor and Jonathan Herzig and Nicholas Lourie and Jonathan Berant},
      year={2019},
      eprint={1811.00937},
      archivePrefix={arXiv},
      primaryClass={cs.CL},
      url={https://arxiv.org/abs/1811.00937}, 
}

@misc{mihaylov2018suitarmorconductelectricity,
      title={Can a Suit of Armor Conduct Electricity? A New Dataset for Open Book Question Answering}, 
      author={Todor Mihaylov and Peter Clark and Tushar Khot and Ashish Sabharwal},
      year={2018},
      eprint={1809.02789},
      archivePrefix={arXiv},
      primaryClass={cs.CL},
      url={https://arxiv.org/abs/1809.02789}, 
}

@misc{doddapaneni2023leavingindiclanguagebehind,
      title={Towards Leaving No Indic Language Behind: Building Monolingual Corpora, Benchmark and Models for Indic Languages}, 
      author={Sumanth Doddapaneni and Rahul Aralikatte and Gowtham Ramesh and Shreya Goyal and Mitesh M. Khapra and Anoop Kunchukuttan and Pratyush Kumar},
      year={2023},
      eprint={2212.05409},
      archivePrefix={arXiv},
      primaryClass={cs.CL},
      url={https://arxiv.org/abs/2212.05409}, 
}

@misc{hendrycks2021measuringmathematicalproblemsolving,
      title={Measuring Mathematical Problem Solving With the MATH Dataset}, 
      author={Dan Hendrycks and Collin Burns and Saurav Kadavath and Akul Arora and Steven Basart and Eric Tang and Dawn Song and Jacob Steinhardt},
      year={2021},
      eprint={2103.03874},
      archivePrefix={arXiv},
      primaryClass={cs.LG},
      url={https://arxiv.org/abs/2103.03874}, 
}

@misc{sarvamai_sarvam_translate,
  title        = {Sarvam‑Translate},
  author       = {{Sarvam AI}},
  year         = {2025},
  howpublished = {\url{https://huggingface.co/sarvamai/sarvam-translate}}
}

@misc{sarvamai_arcc_indic,
  title        = {Arc-Challenge-Indic},
  author       = {{Sarvam AI}},
  year         = {2025},
  howpublished = {\url{https://huggingface.co/datasets/sarvamai/arc-challenge-indic}}
}

@article{meta2025llama,
  title={The llama 4 herd: The beginning of a new era of natively multimodal ai innovation},
  author={Meta, AI},
  journal={https://ai. meta. com/blog/llama-4-multimodal-intelligence/, checked on},
  volume={4},
  number={7},
  pages={2025},
  year={2025}
}

@article{gala2023indictrans2,
  title={Indictrans2: Towards high-quality and accessible machine translation models for all 22 scheduled indian languages},
  author={Gala, Jay and Chitale, Pranjal A and AK, Raghavan and Gumma, Varun and Doddapaneni, Sumanth and Kumar, Aswanth and Nawale, Janki and Sujatha, Anupama and Puduppully, Ratish and Raghavan, Vivek and others},
  journal={arXiv preprint arXiv:2305.16307},
  year={2023}
}

\section{Reproducibility Checklist}

This paper:

\begin{itemize}
    \item Includes a conceptual outline and/or pseudocode description of AI methods introduced - Yes
    \item Clearly delineates statements that are opinions, hypothesis, and speculation from objective facts and results - Yes
    \item Provides well marked pedagogical references for less-familiar readers to gain background necessary to replicate the paper - Yes
    \item Does this paper make theoretical contributions? - No

    \item Does this paper rely on one or more datasets?  - Yes
    
    If yes, please complete the list below.
    
    \item A motivation is given for why the experiments are conducted on the selected datasets  - Yes
    \item All novel datasets introduced in this paper are included in a data appendix  - Yes
    \item All novel datasets introduced in this paper will be made publicly available upon publication of the paper with a license that allows free usage for research purposes - partial
    \item All datasets drawn from the existing literature (potentially including authors’ own previously published work) are accompanied by appropriate citations.  - Yes
    \item All datasets drawn from the existing literature (potentially including authors’ own previously published work) are publicly available.  - Yes
    \item All datasets that are not publicly available are described in detail, with explanation why publicly available alternatives are not scientifically satisficing. - NA
    
    Does this paper include computational experiments?  - Yes
    
    If yes, please complete the list below.
    
    \item This paper states the number and range of values tried per (hyper-) parameter during development of the paper, along with the criterion used for selecting the final parameter setting.  - Yes
    
    \item Any code required for pre-processing data is included in the appendix. - No
    
    \item All source code required for conducting and analyzing the experiments is included in a code appendix. - No
    
    \item All source code required for conducting and analyzing the experiments will be made publicly available upon publication of the paper with a license that allows free usage for research purposes. - No
    
    \item All source code implementing new methods have comments detailing the implementation, with references to the paper where each step comes from - No
    
    \item If an algorithm depends on randomness, then the method used for setting seeds is described in a way sufficient to allow replication of results. - No
    
    \item This paper specifies the computing infrastructure used for running experiments (hardware and software), including GPU/CPU models; amount of memory; operating system; names and versions of relevant software libraries and frameworks. - Partial
    
    \item This paper formally describes evaluation metrics used and explains the motivation for choosing these metrics. - Partial
    
    \item This paper states the number of algorithm runs used to compute each reported result. - Yes
    
    \item Analysis of experiments goes beyond single-dimensional summaries of performance (e.g., average; median) to include measures of variation, confidence, or other distributional information. - No
    
    \item The significance of any improvement or decrease in performance is judged using appropriate statistical tests (e.g., Wilcoxon signed-rank). - No
    
    \item This paper lists all final (hyper-)parameters used for each model/algorithm in the paper’s experiments. - Yes
\end{itemize}

\null
\vfill
\eject

\appendix
\section*{Appendix}
Here, we provide more details about the technical implementation we have followed for generating the data. \ref{appendix:synthetic-data-section} outlines the details pertaining to various methods of synthetic data generation, model evaluations and dataset stats. \ref{appendix:quality-eval-pipeline} contains all the granular details relevant to the Quality Evaluation Pipeline including the details of classifier training, thresholds, WEAT scores etc. \ref{sec:prompts_used} consists of all the prompt templates used for Synthetic Data Generation. \ref{sec:guidelines-manual-annotation} outlines the guidelines provided to our Manual Annotation team for model evaluations. We present few examples across various styles and languages from BhashaKritika in \ref{sec:examples}.

\vspace{0.5em}
\noindent\textbf{Overall Synthetic Data Statistics}

\vspace{0.5em}
\noindent\textbf{Source and synthetic data length statistics:}
Tables \ref{tab:source_generation_ratio}, \ref{tab:style_generation_ratio}, \ref{tab:language_generation_ratio} show the source and synthetic data length statistics across different sources, generation styles and languages respectively. Figure \ref{fig:lang-topic-dist} in the main content shows the distribution of languages (left) and topics (right) in BhashaKritika. 

\begin{table}[htbp]
\centering
\scriptsize
\setlength{\tabcolsep}{4pt} 
\begin{tabular}{lrrr}
\hline
\textbf{Source Dataset} & \textbf{Source Length} & \textbf{Gen Length} & \textbf{Gen/Source Ratio} \\
\hline
FineWeb          & 150 & 414 & 2.76 \\
FineWeb2           & 186 & 460 & 2.47 \\
\hdashline
Personas            & 34  & 242 & 7.22 \\
\hdashline
MATH            & 53  & 159 & 3.01 \\
Numina Math     & 239 & 630 & 2.64 \\
\hdashline
Topic RAG          & 124 & 568 & 4.58 \\
\hdashline
English Cosmopedia & 540 & 572 & 1.06 \\
\hline
\textbf{Total}     & \textbf{148} & \textbf{399} & \textbf{2.70} \\
\hline
\end{tabular}
\caption{Average length (in words) of source and generated data across various strategies used. We consider FineWeb \citep{penedo2024fineweb} and FineWeb2 \citep{penedo2024fineweb2} for document grounded generations, sampled personas from PersonaHub \citep{ge2024scaling}, MATH \citep{hendrycks2021measuringmathematicalproblemsolving} and NuminaMath \citep{numina_math_datasets} for maths and reasoning based generations. Finally, we  also include translations of English Cosmopedia \citep{benallal2024cosmopedia} as part of our synthetic dataset.}
\label{tab:source_generation_ratio}
\end{table}

\begin{table}[htbp]
\centering
\scriptsize
\setlength{\tabcolsep}{3pt}
\begin{tabular}{lrrr}
\hline
\textbf{Generation Style} & \textbf{Source Length} & \textbf{Gen Length} & \textbf{Gen/Source Ratio} \\
\hline
blogpost               & 153 & 308 & 2.02 \\
forums\_story          & 152 & 366 & 2.40 \\
morality\_story        & 152 & 338 & 2.23 \\
persona                & 33  & 243 & 7.25 \\
problem\_solving\_story & 152 & 368 & 2.42 \\
reddit\_post           & 154 & 330 & 2.14 \\
textbook               & 62  & 187 & 3.00 \\
textbook\_academic     & 153 & 451 & 2.94 \\
textbook\_narrative    & 153 & 442 & 2.89 \\
translation            & 360 & 423 & 1.18 \\
wikihow                & 153 & 529 & 3.45 \\
young\_children\_story  & 152 & 294 & 1.93 \\
\textbf{Total}         & \textbf{148} & \textbf{399} & \textbf{2.70} \\
\hline
\end{tabular}
\caption{Average length (in words) of source document and generated data across various generation styles for document grounded generation approach.}
\label{tab:style_generation_ratio}
\end{table}

\begin{table}[htbp]
\centering
\scriptsize
\setlength{\tabcolsep}{3pt}
\begin{tabular}{lrrr}
\hline
\textbf{Language} & \textbf{Source Length} & \textbf{Gen Length} & \textbf{Gen/Source Ratio} \\
\hline
Bengali   & 146 & 318 & 2.19 \\
Gujarati  & 151 & 453 & 3.01 \\
Hindi     & 156 & 501 & 3.22 \\
Kannada  & 530 & 438 & 0.83 \\
Malayalam & 153 & 334 & 2.18 \\
Marathi   & 119 & 318 & 2.67 \\
Oriya   & 531 & 515 & 0.97 \\
Punjabi   & 152 & 545 & 3.59 \\
Tamil     & 147 & 270 & 1.84 \\
Telugu    & 151 & 386 & 2.57 \\
\textbf{Total} & \textbf{148} & \textbf{399} & \textbf{2.70} \\
\hline
\end{tabular}
\caption{Average lengths (words) of Source and Generated Data across various languages.}
\label{tab:language_generation_ratio}
\end{table}

\begin{table}[htbp]
\centering
\begin{tabular}{lr}
\hline
\textbf{Model} & \textbf{Time (H100 GPU hours)} \\
\hline

Krutrim-2     & 309.84 \\
Gemma-3     & 283.28 \\
Sarvam-Translate       & 170.16 \\
Qwen-3           & 1,236.11 \\
\hline
\end{tabular}
\caption{Time taken for generation of 1B tokens in H100 GPU hours. (estimated based on the token calculations from LLaMA-4.)}
\label{tab:model-time}
\end{table}

\section{Synthetic Data Generation}
\label{appendix:synthetic-data-section}

\vspace{0.5em}
\noindent\textbf{1. Document Grounded Generation}
Recognizing that LLM performance varies significantly across languages, we conducted a systematic human evaluation to identify the optimal models for each of our target languages. To mitigate potential rater bias towards specific model providers, all outputs were presented to annotators in an anonymized format with model names hidden. The generations were evaluated on five key criteria: (1) Grammar \& Readability, (2) Faithfulness to Prompt, (3) Overall Generation Quality, (4) Factual Accuracy, and (5) Length of Output on a scale of 1-5. The detailed guidelines provided to the manual annotators can be found in Appendix \ref{sec:guidelines-manual-annotation}.

\noindent Tables \ref{tab:genvstrans-model-eval}-\ref{tab:genvstrans-telugu} show Human evaluation scores of generation quality across LLMs for different languages using two methods- 1. Gen (direct generation by the model in language xx) 2. Trans (generation in En followed by its translation to xx). Table \ref{tab:lang-model-mapping} in the main content shows the model mapping we chose for the synthetic data generation based on our human evaluation. Open-weight LLMs supporting Indian languages are limited, however, we considered Krutrim-2 12B, Gemma-3 27B, LLaMA-3.3 70B, LLaMA-4 (both Maverick 17B-128e and Scout 17B-16e), Qwen-3 32B. Interestingly, we observe high percentage of sentence repetitions when using Qwen-3 and unknown tokens for LLaMA-4 when using for generation in Indian Languages. We thus omit them from our choice of models.

\begin{table*}[htbp]
\centering
\resizebox{\textwidth}{!}{%
\begin{tabular}{lcccccc}
\hline
\textbf{Model (Gen/Trans)} & \textbf{Grammar \& Readability} & \textbf{Faithfulness to Prompt} & \textbf{Overall Generation Quality} & \textbf{Hallucination (Accuracy)} & \textbf{Length of Output} & \textbf{Average} \\
\hline
Krutrim-2                 & 3.18 / 2.70 & 3.46 / 3.34 & 3.22 / 3.01 & 3.61 / 3.42 & 3.69 / 3.84 & 3.43 / 3.26 \\
LLaMA-3.3                 & 3.62 / 3.17 & 3.31 / 3.43 & 3.10 / 2.95 & 3.51 / 3.68 & 2.77 / 3.10 & 3.26 / 3.27 \\
Qwen-3                & 3.13 / 2.46 & 3.26 / 3.01 & 2.99 / 2.66 & 3.43 / 3.12 & 3.38 / 3.64 & 3.24 / 2.98 \\
LLaMA-4-Maverick          & 3.12 / 3.26 & 3.29 / 3.61 & 2.91 / 3.23 & 3.58 / 3.72 & 3.03 / 3.77 & 3.19 / 3.52 \\
LLaMA-4-Scout             & 3.08 / 2.93 & 3.14 / 3.34 & 2.99 / 2.91 & 3.53 / 3.59 & 3.03 / 3.56 & 3.15 / 3.27 \\
Gemma-3                   & 3.01 / 2.36 & 3.05 / 2.79 & 2.75 / 2.35 & 3.33 / 3.04 & 3.50 / 3.45 & 3.13 / 2.80 \\
\hline
\end{tabular}
}
\caption{Human evaluation of average generation quality across LLMs. Scores are on a scale of $1$ to $5$ averaged over $10$ languages. The two scores correspond to generation methods Gen (direct generation by the model in language xx) and Trans (generation in En followed by its translation to xx).}
\label{tab:genvstrans-model-eval}
\end{table*}

\begin{table*}[htbp]
\centering
\resizebox{\textwidth}{!}{%
\begin{tabular}{lcccccc}
\hline
\textbf{Model (Gen/Trans)} & \textbf{Grammar \& Readability} & \textbf{Faithfulness to Prompt} & \textbf{Overall Generation Quality} & \textbf{Hallucination (Accuracy)} & \textbf{Length of Output} & \textbf{Average} \\
\hline
Gemma-3                  & 3.20 / 2.30 & 2.90 / 2.30 & 3.40 / 2.10 & 3.40 / 2.60 & 3.30 / 3.90 & 3.24 / 2.64 \\
Krutrim-2                & 3.20 / 2.90 & 3.20 / 3.00 & 3.40 / 2.80 & 3.60 / 2.70 & 2.30 / 3.50 & 3.14 / 2.98 \\
Qwen-3               & 3.00 / 1.89 & 2.60 / 2.70 & 2.80 / 2.40 & 3.50 / 3.40 & 3.40 / 4.00 & 3.06 / 2.88 \\
LLaMA-4-Scout            & 3.70 / 2.80 & 2.30 / 3.40 & 2.40 / 2.80 & 3.80 / 3.50 & 1.90 / 3.50 & 2.82 / 3.20 \\
LLaMA-4-Maverick         & 3.60 / 2.90 & 2.50 / 3.70 & 2.20 / 3.20 & 3.70 / 3.60 & 2.00 / 3.80 & 2.80 / 3.44 \\
LLaMA-3.3                & 3.10 / 3.30 & 2.70 / 3.40 & 2.70 / 3.00 & 3.10 / 3.00 & 2.30 / 2.70 & 2.78 / 3.08 \\
\hline
\end{tabular}
}
\caption{Human evaluation of generation quality for Bengali. Scores are on a scale of $1$ to $5$, averaged across criteria. Each pair represents Gen (direct generation in Bengali) and Trans (generation in English followed by translation to Bengali).}
\label{tab:genvstrans-bengali}
\end{table*}

\begin{table*}[htbp]
\centering
\resizebox{\textwidth}{!}{%
\begin{tabular}{lcccccc}
\hline
\textbf{Model (Gen/Trans)} & \textbf{Grammar \& Readability} & \textbf{Faithfulness to Prompt} & \textbf{Overall Generation Quality} & \textbf{Hallucination (Accuracy)} & \textbf{Length of Output} & \textbf{Average} \\
\hline
Krutrim-2                & 3.30 / 3.00 & 3.80 / 3.30 & 3.30 / 2.80 & 3.80 / 3.10 & 3.50 / 3.60 & 3.54 / 3.16 \\
Qwen-3               & 3.60 / 1.70 & 3.40 / 2.00 & 3.20 / 1.60 & 3.20 / 2.10 & 3.10 / 3.20 & 3.30 / 2.12 \\
Gemma-3                  & 3.20 / 2.10 & 3.20 / 3.00 & 3.10 / 2.20 & 3.50 / 2.90 & 3.20 / 3.40 & 3.24 / 2.72 \\
LLaMA-4-Scout            & 2.90 / 2.80 & 3.30 / 3.40 & 3.30 / 2.80 & 3.40 / 3.50 & 2.80 / 3.50 & 3.14 / 3.20 \\
LLaMA-3.3                & 3.50 / 2.90 & 3.00 / 3.20 & 3.10 / 2.70 & 3.30 / 3.20 & 2.40 / 2.70 & 3.06 / 2.94 \\
LLaMA-4-Maverick         & 2.90 / 2.90 & 3.20 / 3.70 & 2.90 / 3.20 & 3.20 / 3.60 & 2.60 / 3.80 & 2.96 / 3.44 \\
\hline
\end{tabular}
}
\caption{Human evaluation of generation quality for Gujarati. Scores are on a scale of $1$ to $5$, averaged across criteria. Each pair represents Gen (direct generation in Gujarati) and Trans (generation in English followed by translation to Gujarati).}
\label{tab:genvstrans-gujarati}
\end{table*}

\begin{table*}[htbp]
\centering
\resizebox{\textwidth}{!}{%
\begin{tabular}{lcccccc}
\hline
\textbf{Model (Gen/Trans)} & \textbf{Grammar \& Readability} & \textbf{Faithfulness to Prompt} & \textbf{Overall Generation Quality} & \textbf{Factual Accuracy} & \textbf{Length of Output} & \textbf{Average} \\
\hline
LLaMA-4-Maverick          & 4.00 / 3.90 & 4.10 / 4.80 & 3.60 / 3.80 & 4.50 / 4.70 & 4.20 / 5.00 & 4.08 / 4.44 \\
Qwen-3                & 3.90 / 3.60 & 4.10 / 4.30 & 3.60 / 3.70 & 4.10 / 4.40 & 4.60 / 4.80 & 4.06 / 4.16 \\
LLaMA-4-Scout             & 3.80 / 3.90 & 3.70 / 4.20 & 3.30 / 3.50 & 4.20 / 4.40 & 3.90 / 4.50 & 3.78 / 4.10 \\
Krutrim-2                 & 3.60 / 3.10 & 3.20 / 3.20 & 3.30 / 3.10 & 4.50 / 4.50 & 4.10 / 4.30 & 3.74 / 3.64 \\
Gemma-3                  & 3.80 / 3.50 & 2.80 / 3.10 & 3.10 / 3.40 & 4.10 / 4.40 & 4.40 / 4.30 & 3.64 / 3.74 \\
LLaMA-3.3                 & 3.80 / 3.70 & 3.30 / 3.40 & 3.10 / 3.60 & 3.90 / 4.70 & 3.50 / 4.40 & 3.52 / 3.96 \\
\hline
\end{tabular}
}
\caption{Human evaluation of generation and translation quality for Hindi. Scores are on a scale of $1$ to $5$, averaged across criteria. Each pair represents Gen (direct generation in Hindi) and Trans (generation in English followed by translation to Hindi).}
\label{tab:genvstrans-hindi}
\end{table*}

\begin{table*}[htbp]
\centering
\resizebox{\textwidth}{!}{%
\begin{tabular}{lcccccc}
\hline
\textbf{Model (Gen/Trans)} & \textbf{Grammar \& Readability} & \textbf{Faithfulness to Prompt} & \textbf{Overall Generation Quality} & \textbf{Factual Accuracy} & \textbf{Length of Output} & \textbf{Average} \\
\hline
Qwen-3                 & 2.70 / 2.00 & 2.90 / 2.70 & 2.70 / 2.20 & 3.40 / 3.10 & 3.40 / 3.80 & 3.02 / 2.76 \\
Krutrim-2                 & 2.60 / 2.80 & 3.00 / 3.40 & 2.50 / 2.90 & 2.90 / 3.60 & 4.00 / 4.00 & 3.00 / 3.34 \\
LLaMA-3.3                 & 2.70 / 2.80 & 2.50 / 3.10 & 2.50 / 2.60 & 3.00 / 3.00 & 3.30 / 3.00 & 2.80 / 2.90 \\
LLaMA-4-Scout             & 2.00 / 2.10 & 2.40 / 3.00 & 2.20 / 2.40 & 2.80 / 3.50 & 2.80 / 3.80 & 2.44 / 2.96 \\
LLaMA-4-Maverick          & 1.90 / 2.00 & 2.40 / 2.70 & 2.30 / 2.50 & 2.80 / 3.20 & 2.80 / 3.40 & 2.44 / 2.76 \\
Gemma-3                  & 1.00 / 1.10 & 2.00 / 2.10 & 1.00 / 1.10 & 2.00 / 2.10 & 3.00 / 3.00 & 1.80 / 1.88 \\
\hline
\end{tabular}
}
\caption{Human evaluation of generation and translation quality for Kannada. Scores are on a scale of $1$ to $5$, averaged across criteria. Each pair represents Gen (direct generation in Kannada) and Trans (generation in English followed by translation to Kannada).}
\label{tab:genvstrans-kannada}
\end{table*}

\begin{table*}[htbp]
\centering
\resizebox{\textwidth}{!}{%
\begin{tabular}{lcccccc}
\hline
\textbf{Model (Gen/Trans)} & \textbf{Grammar \& Readability} & \textbf{Faithfulness to Prompt} & \textbf{Overall Generation Quality} & \textbf{Factual Accuracy} & \textbf{Length of Output} & \textbf{Average} \\
\hline
Krutrim-2                 & 3.30 / 2.40 & 3.50 / 3.20 & 3.30 / 2.50 & 4.00 / 3.10 & 3.80 / 4.00 & 3.58 / 3.04 \\
LLaMA-4-Maverick          & 3.50 / 3.40 & 3.50 / 3.30 & 3.30 / 3.40 & 3.50 / 3.60 & 3.90 / 3.70 & 3.54 / 3.48 \\
LLaMA-3.3                 & 3.70 / 3.00 & 4.00 / 3.40 & 3.10 / 2.00 & 3.90 / 3.80 & 2.40 / 3.00 & 3.42 / 3.04 \\
Qwen-3                & 2.90 / 2.90 & 3.50 / 3.20 & 3.10 / 2.90 & 3.60 / 3.10 & 3.40 / 3.60 & 3.30 / 3.14 \\
LLaMA-4-Scout             & 2.70 / 2.60 & 2.80 / 3.20 & 3.30 / 3.10 & 3.70 / 3.40 & 3.20 / 3.70 & 3.14 / 3.20 \\
Gemma-3                  & 2.60 / 1.20 & 3.00 / 1.80 & 1.90 / 1.30 & 3.00 / 2.10 & 3.80 / 3.80 & 2.86 / 2.04 \\
\hline
\end{tabular}
}
\caption{Human evaluation of generation and translation quality for Malayalam. Scores are on a scale of $1$ to $5$, averaged across criteria. Each pair represents Gen (direct generation in Malayalam) and Trans (generation in English followed by translation to Malayalam).}
\label{tab:genvstrans-malayalam}
\end{table*}

\begin{table*}[htbp]
\centering
\resizebox{\textwidth}{!}{%
\begin{tabular}{lcccccc}
\hline
\textbf{Model (Gen/Trans)} & \textbf{Grammar \& Readability} & \textbf{Faithfulness to Prompt} & \textbf{Overall Generation Quality} & \textbf{Factual Accuracy} & \textbf{Length of Output} & \textbf{Average} \\
\hline
Krutrim-2                 & 3.30 / 2.70 & 3.70 / 3.80 & 3.60 / 3.70 & 3.50 / 3.60 & 4.40 / 4.40 & 3.70 / 3.64 \\
Gemma-3                   & 3.90 / 2.20 & 3.40 / 3.90 & 3.30 / 3.60 & 3.60 / 3.80 & 4.20 / 4.50 & 3.68 / 3.60 \\
LLaMA-4-Scout             & 3.50 / 3.00 & 3.60 / 3.10 & 3.50 / 3.10 & 3.50 / 3.00 & 3.70 / 3.50 & 3.56 / 3.14 \\
LLaMA-4-Maverick          & 3.10 / 3.00 & 3.40 / 3.10 & 3.40 / 3.10 & 3.40 / 3.20 & 3.50 / 3.70 & 3.36 / 3.22 \\
LLaMA-3.3                 & 3.90 / 3.20 & 2.90 / 3.40 & 2.70 / 3.50 & 2.90 / 3.80 & 3.70 / 3.90 & 3.22 / 3.56 \\
Qwen-3                & 2.80 / 1.90 & 3.00 / 2.50 & 2.80 / 2.40 & 2.70 / 2.40 & 3.20 / 3.30 & 2.90 / 2.50 \\
\hline
\end{tabular}
}
\caption{Human evaluation of generation and translation quality for Marathi. Scores are on a scale of $1$ to $5$, averaged across criteria. Each pair represents Gen (direct generation in Marathi) and Trans (generation in English followed by translation to Marathi).}
\label{tab:genvstrans-marathi}
\end{table*}

\begin{table*}[htbp]
\centering
\resizebox{\textwidth}{!}{%
\begin{tabular}{lcccccc}
\hline
\textbf{Model (Gen/Trans)} & \textbf{Grammar \& Readability} & \textbf{Faithfulness to Prompt} & \textbf{Overall Generation Quality} & \textbf{Factual Accuracy} & \textbf{Length of Output} & \textbf{Average} \\
\hline
Qwen-3                & 3.20 / 2.50 & 3.30 / 2.70 & 3.30 / 2.70 & 3.50 / 2.70 & 3.60 / 3.40 & 3.38 / 2.80 \\
LLaMA-3.3                 & 3.50 / 2.30 & 3.60 / 2.40 & 3.50 / 2.60 & 3.20 / 3.00 & 3.10 / 2.90 & 3.38 / 2.64 \\
LLaMA-4-Maverick          & 3.00 / 3.30 & 2.80 / 2.90 & 2.40 / 2.90 & 2.60 / 2.40 & 2.90 / 3.20 & 2.74 / 2.94 \\
Krutrim-2                 & 1.80 / 1.80 & 2.50 / 2.30 & 2.20 / 2.40 & 2.40 / 2.60 & 3.50 / 3.70 & 2.48 / 2.56 \\
LLaMA-4-Scout             & 1.90 / 2.30 & 2.00 / 2.30 & 2.10 / 2.20 & 2.10 / 2.80 & 2.80 / 2.80 & 2.18 / 2.48 \\
Gemma-3                  & 1.00 / 1.00 & 1.00 / 1.00 & 1.00 / 1.00 & 1.00 / 1.00 & 2.00 / 1.90 & 1.20 / 1.18 \\
\hline
\end{tabular}
}
\caption{Human evaluation of generation and translation quality for Oriya. Scores are on a scale of $1$ to $5$, averaged across criteria. Each pair represents Gen (direct generation in Oriya) and Trans (generation in English followed by translation to Oriya).}
\label{tab:genvstrans-Oriya}
\end{table*}

\begin{table*}[htbp]
\centering
\resizebox{\textwidth}{!}{%
\begin{tabular}{lcccccc}
\hline
\textbf{Model (Gen/Trans)} & \textbf{Grammar \& Readability} & \textbf{Faithfulness to Prompt} & \textbf{Generation Quality} & \textbf{Factual Accuracy} & \textbf{Length of Output} & \textbf{Average} \\
\hline
LLaMA-4-Scout            & 4.10 / 2.90 & 4.60 / 3.70 & 4.30 / 3.10 & 4.40 / 3.30 & 4.00 / 3.60 & 4.28 / 3.32 \\
Gemma-3                 & 4.20 / 3.70 & 4.20 / 2.90 & 4.00 / 2.50 & 4.60 / 3.00 & 4.20 / 2.10 & 4.24 / 2.84 \\
Krutrim-2               & 3.90 / 2.90 & 4.00 / 3.90 & 3.20 / 3.60 & 3.60 / 3.70 & 4.44 / 3.78 & 3.83 / 3.58 \\
LLaMA-3.3               & 4.50 / 4.00 & 3.90 / 4.30 & 3.90 / 3.50 & 4.20 / 4.20 & 2.40 / 2.50 & 3.78 / 3.70 \\
LLaMA-4-Maverick        & 3.60 / 4.20 & 4.20 / 4.40 & 3.40 / 3.90 & 4.30 / 4.30 & 3.40 / 4.30 & 3.78 / 4.22 \\
Qwen-3              & 3.30 / 3.00 & 3.70 / 3.60 & 3.20 / 3.30 & 3.60 / 3.30 & 3.20 / 3.20 & 3.40 / 3.28 \\
\hline
\end{tabular}
}
\caption{Human evaluation of generation and translation quality for Punjabi. Each pair shows Gen (generation in Punjabi) and Trans (generation in English followed by translation to Punjabi). Scores are on a scale of $1$ to $5$.}
\label{tab:genvstrans-punjabi}
\end{table*}

\begin{table*}[htbp]
\centering
\resizebox{\textwidth}{!}{%
\begin{tabular}{lcccccc}
\hline
\textbf{Model (Gen/Trans)} & \textbf{Grammar \& Readability} & \textbf{Faithfulness to Prompt} & \textbf{Generation Quality} & \textbf{Factual Accuracy} & \textbf{Length of Output} & \textbf{Average} \\
\hline
Gemma-3                 & 3.50 / 3.60 & 4.30 / 4.50 & 3.80 / 3.90 & 4.10 / 4.50 & 3.90 / 4.60 & 3.92 / 4.22 \\
Krutrim-2               & 3.20 / 2.80 & 3.90 / 3.60 & 3.60 / 3.10 & 4.10 / 3.30 & 3.80 / 4.10 & 3.72 / 3.38 \\
LLaMA-3.3               & 3.70 / 2.90 & 3.40 / 3.90 & 3.20 / 3.10 & 4.20 / 4.10 & 2.20 / 2.90 & 3.34 / 3.38 \\
LLaMA-4-Maverick        & 2.40 / 2.70 & 3.30 / 3.60 & 2.50 / 2.80 & 3.80 / 3.90 & 2.20 / 3.90 & 2.84 / 3.38 \\
LLaMA-4-Scout           & 2.50 / 2.10 & 3.10 / 3.40 & 2.50 / 2.40 & 3.40 / 3.70 & 2.30 / 4.00 & 2.76 / 3.12 \\
Qwen-3              & 2.40 / 2.20 & 2.50 / 3.00 & 2.10 / 2.50 & 3.10 / 3.20 & 2.90 / 4.10 & 2.60 / 3.00 \\
\hline
\end{tabular}
}
\caption{Human evaluation of generation and translation quality for Tamil. Each score shows Gen (generation directly in Tamil) / Trans (generation in English followed by translation to Tamil). All values are rated on a $1$–$5$ scale.}
\label{tab:genvstrans-tamil}
\end{table*}

\begin{table*}[htbp]
\centering
\resizebox{\textwidth}{!}{%
\begin{tabular}{lcccccc}
\hline
\textbf{Model (Gen/Trans)} & \textbf{Grammar \& Readability} & \textbf{Faithfulness to Prompt} & \textbf{Generation Quality} & \textbf{Factual Accuracy} & \textbf{Length of Output} & \textbf{Average} \\
\hline
Krutrim-2                 & 3.60 / 2.60 & 3.80 / 3.70 & 3.80 / 3.20 & 3.70 / 4.00 & 3.10 / 3.00 & 3.60 / 3.30 \\
Gemma-3                  & 3.70 / 2.90 & 3.70 / 3.30 & 2.90 / 2.40 & 4.00 / 4.00 & 3.00 / 3.00 & 3.46 / 3.12 \\
LLaMA-4-Scout            & 3.70 / 3.20 & 3.60 / 3.60 & 3.00 / 3.00 & 4.00 / 4.00 & 2.90 / 3.10 & 3.44 / 3.38 \\
Qwen-3               & 3.50 / 2.90 & 3.60 / 3.40 & 3.10 / 2.90 & 3.60 / 3.50 & 3.00 / 3.00 & 3.36 / 3.14 \\
LLaMA-3.3                & 3.80 / 3.60 & 3.80 / 3.80 & 3.20 / 2.90 & 3.40 / 4.00 & 2.40 / 3.00 & 3.32 / 3.46 \\
LLaMA-4-Maverick         & 3.20 / 3.00 & 3.50 / 3.30 & 3.10 / 3.00 & 4.00 / 4.00 & 2.80 / 3.00 & 3.32 / 3.26 \\
\hline
\end{tabular}
}
\caption{Human evaluation of generation and translation quality for Telugu. Each score is presented as Gen (direct generation in Telugu) / Trans (generation in English followed by translation to Telugu). Scores are rated on a scale of 1 to 5.}
\label{tab:genvstrans-telugu}
\end{table*}

\vspace{0.5em}
\noindent\textbf{2. Persona-Based Generation}

\vspace{0.2em}
\noindent\textbf{Overall generation statistics:}
Tables \ref{tab:persona-gen-stats}, \ref{tab:persona_token_stats} display the generation statistics across different categories and generation paths.

\begin{table*}[htbp]
\centering
\scriptsize
\resizebox{\textwidth}{!}{%
\begin{tabular}{@{}lrr p{6cm}@{}}
\toprule
\textbf{Persona Category} & \textbf{Input Persona Count} & \textbf{Output Persona Count} & \textbf{Remark} \\
\midrule
Instruction/Knowledge/Reasoning/Tool & 165K & ~1.3M & General category personas from PersonaHub \\
Elite Personas (Used in Generation) & 20M & 163M & High-quality curated personas \\
Total Multilingual Personas Generated & 6K  & 50K & Generated across Indian languages \\
\bottomrule
\end{tabular}
}
\caption{Persona generation statistics across various categories.}
\label{tab:persona-gen-stats}
\end{table*}

\begin{table}[htbp]
\centering
\small
\begin{tabular}{@{}lcl p{7cm}@{}}
\toprule
\textbf{Generation Path} & \textbf{Tokens Generated} \\
\midrule
Elite Persona $\rightarrow$ Text & $\sim$10B \\
Other Persona $\rightarrow$ Text & 7.7M \\
Persona + Document grounded $\rightarrow$ Text & 1.5B \\
\addlinespace
\bottomrule
\end{tabular}
\caption{Persona-based generation token statistics across various generation paths.}
\label{tab:persona_token_stats}
\end{table}

\vspace{0.5em}
\noindent\textbf{3. Maths and Reasoning based synthetic data}

\noindent Table \ref{tab:math-data-sources} shows various math focused datasets usable for synthetic data generations. On eyeballing some generations, we observed that using easy 
grade school level math examples as source simplified the generations too much.
\begin{table}[htbp]
\centering
\scriptsize
\begin{tabular}{llcc}
\hline
\textbf{Source} & \textbf{Type} & \textbf{Samples} & \textbf{Gen Tokens (B)} \\
\hline
MATH              & QA-IT           & 12K      & 0.014  \\
NuminaMath        & QA + CoT        & 850K     &    5.083  \\
\hdashline
\textbf{Total}  &                 &     4.06M     &    5.097    \\
\hline
\end{tabular}
\caption{Statistics of Maths and Reasoning focused datasets.}
\label{tab:math-data-sources}
\end{table}

\vspace{0.5em}
\noindent\textbf{4. Topic aware Retrieval Augmentation Generation (RAG) techniques}
\label{appendix:topic-based-rag}
To ensure comprehensive coverage of topics relevant to India, we implemented a targeted data expansion strategy. First, we performed topic modeling on our existing data to understand its thematic distribution. We used Vyakyarth embeddings with UMAP for dimensionality reduction and a FAISS-powered DBSCAN algorithm to group documents into distinct clusters, adapting the text-clustering library from Hugging Face\footnote{\url{https://github.com/huggingface/text-clustering}}. Each resulting cluster was then assigned a descriptive label via LLM-based summarization to clarify its topic.

This analysis revealed a significant concentration in specific areas; for instance, ``Indian Lifestyle" and ``Bollywood" collectively constituted over $25\%$ of the dataset
. To identify and fill underrepresented domains, we first curated a comprehensive list of target topics by traversing Wikipedia's knowledge graph, starting from the Category:India\footnote{\url{https://en.wikipedia.org/wiki/Category:India}} until depth 3 scraping over 10k titles. We then computed FAISS similarity scores between our existing topic clusters and this target topic list. Any target topic with a similarity score below a threshold of 0.4 was identified as a coverage gap. For each gap, we used SERP API to fetch and scrape new documents. These documents are then used as a source for the document grounded generations.

\noindent Table \ref{tab:broad_topic_distribution} shows broad as well as specific topic distributions (\%) of the synthetic data generated.

\begin{table*}[htbp]
\centering
\scriptsize
\begin{tabular}{llr}
\toprule
\textbf{Broad Topic} & \textbf{Specific Topic} & \textbf{Percentage} \\
\midrule
\multirow{8}{*}{Indian Culture \& Society} 
& Indian Lifestyle & 18.50\% \\
& Indian Philosophy & 3.30\% \\
& Personal Stories & 2.70\% \\
& Travel Guide & 1.20\% \\
& Indian Culture and Religion & 0.90\% \\
& Indian Fashion & 0.40\% \\
& Indian Tourism & 0.40\% \\
& Indian Culture and Heritage & 0.30\% \\
\midrule
\multirow{9}{*}{Science \& Technology} 
& Science and Technology in India & 2.10\% \\
& Mathematics & 4.10\% \\
& Everyday Science & 3.30\% \\
& Mobile Phones and Technology & 1.50\% \\
& Computer Science/Technology & 0.70\% \\
& Technology and Digital Transformation in India & 0.60\% \\
& Automobiles & 0.30\% \\
& Science & 0.30\% \\
& Telecom and Technology in India & 0.20\% \\
\midrule
\multirow{6}{*}{Health \& Wellness}
& Indian Healthcare & 2.80\% \\
& Health and Medicine & 2.50\% \\
& Health and Wellness & 1.90\% \\
& Yoga & 1.30\% \\
& Healthcare & 1.10\% \\
& Indian Health and Wellness & 0.90\% \\
\midrule
\multirow{7}{*}{Politics, Government \& Law}
& Indian Politics & 3.70\% \\
& Indian Law and Justice System & 1.80\% \\
& Government Jobs & 1.10\% \\
& Crime & 1.00\% \\
& Road Safety & 0.50\% \\
& Safety and Security Measures & 0.30\% \\
& Indian Law & 0.30\% \\
\midrule
\multirow{5}{*}{Entertainment \& Media}
& Bollywood & 7.30\% \\
& Indian Cuisine in Cinema & 0.60\% \\
& Gaming & 0.60\% \\
& Indian Music & 0.50\% \\
& Online Gaming and Digital Payments in India & 0.30\% \\
\midrule
\multirow{4}{*}{Education \& Exams}
& Education & 2.70\% \\
& Indian Children's Science Stories & 1.10\% \\
& Indian Exams and Education System & 0.90\% \\
& Indian Education & 0.40\% \\
\midrule
\multirow{8}{*}{Business, Finance \& Economy}
& Economics & 2.60\% \\
& Finance & 2.30\% \\
& Business and Economy in India & 1.30\% \\
& Marketing & 0.70\% \\
& Economics/Business & 0.60\% \\
& E-commerce and Business & 0.30\% \\
& Business/Economics & 0.30\% \\
\midrule
\multirow{3}{*}{Sports \& Recreation}
& Cricket & 3.70\% \\
& Indian Sports & 1.20\% \\
& Sports & 1.10\% \\
\midrule
\multirow{2}{*}{Food \& Cuisine}
& Indian Cuisine & 1.90\% \\
& Indian Recipes & 0.80\% \\
\midrule
\multirow{3}{*}{International Relations \& Security}
& International Relations/Security/Terrorism (related to India) & 0.80\% \\
& International Relations of India & 0.70\% \\
& International Relations & 0.50\% \\
\midrule
\multirow{4}{*}{Environment \& Sustainability}
& Indian Agriculture and Environment & 1.70\% \\
& Energy & 0.60\% \\
& Environmental Studies/Sustainability & 0.50\% \\
& Environmental Conservation in India & 0.20\% \\
\midrule
\multirow{4}{*}{Language, Arts \& Literature}
& Indian Literature & 2.00\% \\
& Linguistics & 0.60\% \\
& Indian Languages & 0.50\% \\
& Indian Arts and Crafts & 0.50\% \\
\bottomrule
\end{tabular}
\caption{Distribution of specific topics under broad topic categories.}
\label{tab:broad_topic_distribution}
\end{table*}

\vspace{0.5em}
\noindent\textbf{5. Translation}
We follow a similar setup of anonymised evaluation as grounded generations for assessing the ability of translation models across languages to choose models for translation. The translations from various models are evaluated on four key criteria - (1) Grammar \& Readability, (2) Translation Faithfulness, (3) Terminology and Domain Consistency and (4) Fluency \& Style on a scale of 1-5.

\noindent Table \ref{tab:trans-eval} shows Human evaluation scores of different translation models based on grammar, prompt faithfulness, generation quality, and factual accuracy criteria. 
We evaluate IndicTrans2~\citep{gala2023indictrans2}, and Sarvam-Translate~\citep{sarvamai_sarvam_translate}  for translations.

\begin{table*}[htbp]
\centering
\resizebox{\textwidth}{!}{%
\begin{tabular}{lrrrrr}
\hline
\textbf{Model} & \textbf{Grammar \& Readability} & \textbf{Faithfulness to Prompt} & \textbf{Generation Quality} & \textbf{Factual Accuracy} & \textbf{Average} \\
\hline
IndicTrans2          & 3.44 & 3.59 & 3.28 & 3.27 & 3.40 \\
Sarvam-Translate     & \textbf{3.79} & \textbf{3.85} & \textbf{3.72} & \textbf{3.66} & \textbf{3.76} \\
\hline
\end{tabular}
}
\caption{Human evaluation of different translation models across considered languages. Scores are on a scale of $1$ to $5$ based on grammar, prompt faithfulness, generation quality, and factual accuracy. We chose Sarvam-Translate as it turned out to be the best for translation across languages}
\label{tab:trans-eval}
\end{table*}

\vspace{0.5em}
\noindent\textbf{Implementation Details}
We leverage the vLLM \citep{kwon2023efficient} inference library to create model endpoints for generating synthetic data at scale for local models. After choosing the relevant models, we estimate the time required for generating 1B model tokens for various open source weight models which are choosen for generations for scaling purposes. Table \ref{tab:model-time} contains the time in GPU hours required to generate 1B tokens (estimated using the LLaMA-4 tokenizer) from various models.

\subsection{Quality Evaluation Pipeline}
\label{appendix:quality-eval-pipeline}
\noindent The discard rates for each of the quality filter across different languages are displayed in the Table \ref{tab:lang-wise-filter-rates}.

\begin{table*}[htbp]
\centering
\scriptsize
\resizebox{\textwidth}{!}{%
\begin{tabular}{lrrrrrrrrrrr}
\hline
\textbf{Lang} & \textbf{Language Mismatch} & \textbf{Word N-gram Repetition} & \textbf{Length Violation} & \textbf{NSFW Words} & \textbf{Stop Words} & \textbf{AI Words} & \textbf{Non-Latin Non-Indic Words} & \textbf{High Perplexity} & \textbf{Low Quality} \\
\hline
Bengali   & 0.14  & 0.07  & 0.20  & 1.75 & 0.00 & 0.02 & 0.00 & 11.10 & 0.21 \\
Gujarati  & 11.20 & 18.93 & 11.21 & 0.70 & 0.00 & 0.02 & 0.03 & 0.22 & 23.96  \\
Hindi     & 14.83 & 0.02  & 1.30  & 1.05 & 0.00 & 0.03 & 0.00 & 3.86 & 0.62 \\
Malayalam & 2.28  & 8.81  & 70.00 & 0.22 & 0.00 & 0.03 & 0.02 & 0.49 & 45.40 \\
Marathi   & 0.57  & 0.03  & 0.60  & 0.83 & 0.00 & 0.01 & 0.00 & 2.21 & 0.51 \\
Punjabi   & 8.23  & 6.73  & 8.12  & 5.32 & 0.00 & 0.02 & 0.00 & 0.69 & 21.04 \\
Tamil     & 0.21  & 0.33  & 3.97  & 0.67 & 0.00 & 0.03 & 0.00 & 15.10 & 12.04 \\
Telugu    & 4.18  & 2.08  & 3.91  & 0.57 & 0.00 & 0.05 & 0.00 & 1.45 & 25.22 \\
\textbf{Overall}    & \textbf{7.66}  & \textbf{0.34}  & \textbf{2.26} & \textbf{1.13} & \textbf{0.00} & \textbf{0.02} & \textbf{0.00} & \textbf{7.09} & \textbf{3.40} \\
\hline
\end{tabular}
}
\caption{Language-wise distribution of filtering violations (discard rates \%) across multiple filtering dimensions.}
\label{tab:lang-wise-filter-rates}
\end{table*}

\vspace{0.5em}
\noindent\textbf{1. Heuristic content filter}

Our filtering pipeline targets low-quality content by detecting NSFW material, repetitive or generic text, anomalous characters, outlier word counts, and third-party AI references. Each criterion is controlled by empirically tuned thresholds outlined in Table~\ref{tab:filter-thresholds}.

\begin{table}[htbp]
\centering
\footnotesize
\begin{tabular}{ll}
\hline
\textbf{Filter} & \textbf{Threshold} \\
\hline
Word Count & $[100, 2500]$ \\
NSFW Words Ratio & $= 0.0$ \\
Stop Words Ratio & $\leq 0.6$ \\
AI Words Ratio & $= 0.0$ \\
Non-Latin/Indic Words Ratio & $\leq 0.15$ \\
6-gram Word Repetition & $\leq 0.3$ \\
\hline
\end{tabular}
\caption{Filtering thresholds for text quality estimation. }
\label{tab:filter-thresholds}
\end{table}

\vspace{0.5em}
\noindent\textbf{2. Fluency (perplexity-based) filter}

In order to evaluate the fluency of the generated synthetic data using perplexity scoring, we train a 5-gram Kneser-Ney model using the KenLM \citep{heafield2011kenlm} library.

\vspace{0.5em}
\noindent\textbf{Training and validation dataset statistics:}
Table~\ref{tab:perp-data} summarizes the language-wise data sources and total counts used for training the KenLM-based perplexity models. To evaluate and calibrate perplexity thresholds, we curated a validation dataset consisting of clean, high-quality samples across multiple Indic languages.

\begin{table}[htbp]
\centering
\scriptsize
\begin{tabular}{lrrr}
\hline
\textbf{Language} & \textbf{Train} & \textbf{Validation} & \textbf{Threshold} \\
\hline
Bengali    & 1,117,660 & 139,708 & 5,800 \\
Gujarati   & 1,028,602 & 128,577 & 7,740 \\
Hindi      & 1,266,793 & 158,352 & 640 \\
Kannada    & 905,149   & 113,145 & 29,400 \\
Malayalam  & 1,028,632 & 128,580 & 84,400 \\
Marathi    & 1,215,518 & 151,942 & 5,400 \\
Punjabi    & 1,053,538 & 131,693 & 520 \\
Tamil      & 1,181,781 & 147,726 & 35,100 \\
Telugu     & 995,141   & 124,394 & 23,100 \\
\hline
\textbf{Total}   & \textbf{14,508,725} & \textbf{1,813,627} \\
\hline
\end{tabular}
\caption{Language-wise distribution of training and validation data for 5-gram KenLM model (for perplexity scoring) along with thresholds.}
\label{tab:perp-data}
\end{table}

\vspace{0.5em}
\noindent\textbf{Threshold selection:}
Per-language thresholds (see Table \ref{tab:perp-data}) were computed by scoring a validation set and taking the 80th percentile perplexity score, in line with the methodology used in the Setu pipeline. Further manual inspection was conducted to adjust thresholds upward for languages where high-quality texts were inadvertently being flagged due to overly strict cutoff values.

\vspace{0.8em}
\noindent\textbf{3. Quality classifiers}

We assess the overall quality of synthetic data across dimensions such as accuracy, clarity, coherence, grammar, informational depth, and usefulness. A custom-trained FastText~\citep{joulin2016fasttext} binary classifier is used to automatically label Indic-language responses as either \texttt{high} or \texttt{low} quality.

\vspace{0.5em}
\noindent\textbf{Training and test dataset statistics:}
Table~\ref{tab:classifier-data} summarizes the language-wise data sources and total counts used for training and testing of the fasttext binary classifier model.

\begin{table}[htbp]
\centering
\tiny
\setlength{\tabcolsep}{3pt}
\begin{tabular}{lrrrr}
\hline
\textbf{Language} & \textbf{Train (High)} & \textbf{Train (Low)} & \textbf{Test (High)} & \textbf{Test (Low)} \\
\hline
Assamese   & 10,012 & 19,102 & 0      & 0     \\
Bengali    & 20,833 & 20,833 & 35,392 & 2,452 \\
Gujarati   & 20,833 & 20,520 & 3,462  & 0     \\
Hindi      & 20,833 & 20,833 & 32,932 & 684   \\
Kannada    & 9,246  & 20,833 & 0      & 752   \\
Malayalam  & 5,940  & 20,833 & 0      & 791   \\
Marathi    & 20,833 & 20,833 & 39,845 & 1,050 \\
Oriya      & 3,480  & 6,824  & 0      & 0     \\
Punjabi    & 3,727  & 20,833 & 0      & 10,243 \\
Sanskrit   & 3,429  & 20,833 & 0      & 8,864 \\
Tamil      & 20,833 & 20,833 & 16,272 & 3,256 \\
Telugu     & 10,434 & 20,833 & 0      & 4,440 \\
\hline
\textbf{Total} & \textbf{150,433} & \textbf{233,943} & \textbf{127,903} & \textbf{32,532} \\
\hline
\end{tabular}
\caption{Language-wise distribution (counts) of training and test data for FastText binary classifier (overall quality classification). `High' and `Low' denotes class labels.}
\label{tab:classifier-data}
\end{table}

\vspace{0.5em}
\noindent\textbf{Evaluation statistics:}
Table~\ref{tab:evaluation-metrics} summarizes the performance of a binary classification model evaluated on a test set. It demonstrates that the model performs consistently well across both `high' and `low' quality classes, achieving high accuracy, precision, recall, and F1 scores.

\begin{table}[htbp]
\centering
\scriptsize
\begin{tabular}{lr}
\hline
\textbf{Metric} & \textbf{Score} \\
\hline
Accuracy & 0.989 \\
F1 (High) & 0.994 \\
F1 (Low) & 0.977 \\
Precision (High) & 0.995 \\
Precision (Low) & 0.974 \\
Recall (High) & 0.993 \\
Recall (Low) & 0.979 \\
\hline
\end{tabular}
\caption{Evaluation statistics of FastText binary classifier on the test set.}
\label{tab:evaluation-metrics}
\end{table}

\vspace{0.5em}
\noindent\textbf{4. Bias detection}

The Word Embedding Association Test (WEAT)~\citep{jentzsch2019semantics} quantifies implicit bias in word embeddings by measuring the differential association between two sets of target words $X$ and $Y$ (e.g., career vs.\ family) , and two sets of attribute words $A$ and $B$ (e.g., male vs.\ female terms). The test statistic is defined as:

\[
s(X, Y, A, B) = \sum_{x \in X} [s(x, A, B)] - \sum_{y \in Y} [s(y, A, B)]
\]
where
\[
s(w, A, B) = \text{mean}_{a \in A} \cos(\vec{w}, \vec{a}) - \text{mean}_{b \in B} \cos(\vec{w}, \vec{b})
\]

Here, $\cos(\vec{w}, \vec{a})$ denotes the cosine similarity between word vectors. The effect size is calculated as:

\[
\text{effect size} = \frac{s(X, Y, A, B)}{\text{std\_dev}_{w \in X \cup Y} s(w, A, B)}
\]

A larger effect size indicates stronger bias embedded in the representation.

\vspace{0.5em}
\noindent\textbf{Bias dimensions and WEAT configurations:}
Table~\ref{tab:weat-bias-aspects} outlines the different social bias dimensions evaluated using the WEAT (Word Embedding Association Test) framework. For each bias type—such as gender, caste, race, religion, and region—it specifies the contrasting stereotype groups used as target sets and the attribute sets representing evaluative dimensions.

\begin{table}[htbp]
\centering
\scriptsize
\begin{tabular}{lp{2.8cm}p{3cm}}
\hline
\textbf{Bias Aspect} & \textbf{Target Sets} & \textbf{Attribute Sets} \\
\hline
Gender & Career vs. Family & Male vs. Female \\
Caste & Marginalised vs. Upper & Unpleasant vs. Pleasant \\
Race & Dark vs. Fair & Negative vs. Positive \\
Religion & Islam vs. Hindu & Negative vs. Positive \\
Region & Marginalised vs. Dominant & Discriminatory vs. Prestigious \\
\hline
\end{tabular}
\caption{Bias dimensions and WEAT configurations.}
\label{tab:weat-bias-aspects}
\end{table}

\begin{figure}[htbp]
    \centering
    \includegraphics[width=\linewidth]{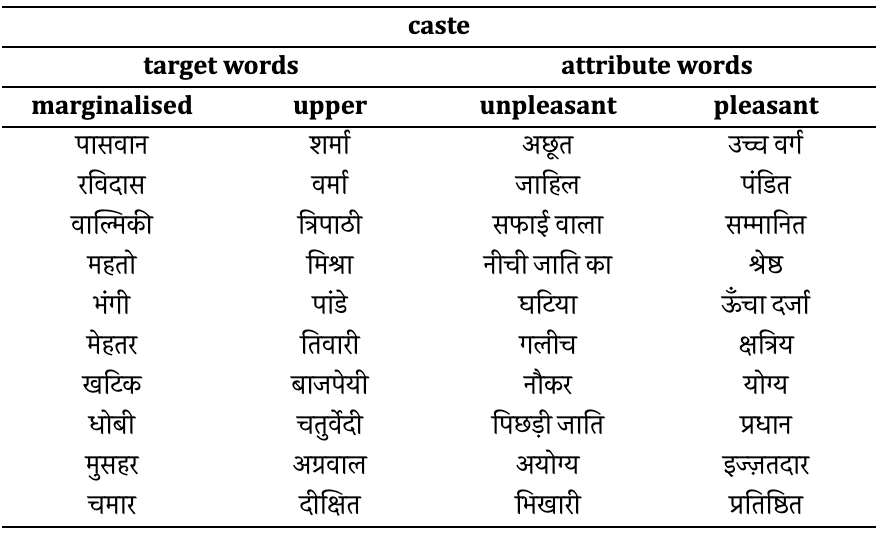}
    \caption{Manually curated bias words (target and attribute sets) for caste aspect.}
    \label{fig:bias_words_caste}
\end{figure}

\begin{figure}[htbp]
    \centering
    \includegraphics[width=\linewidth]{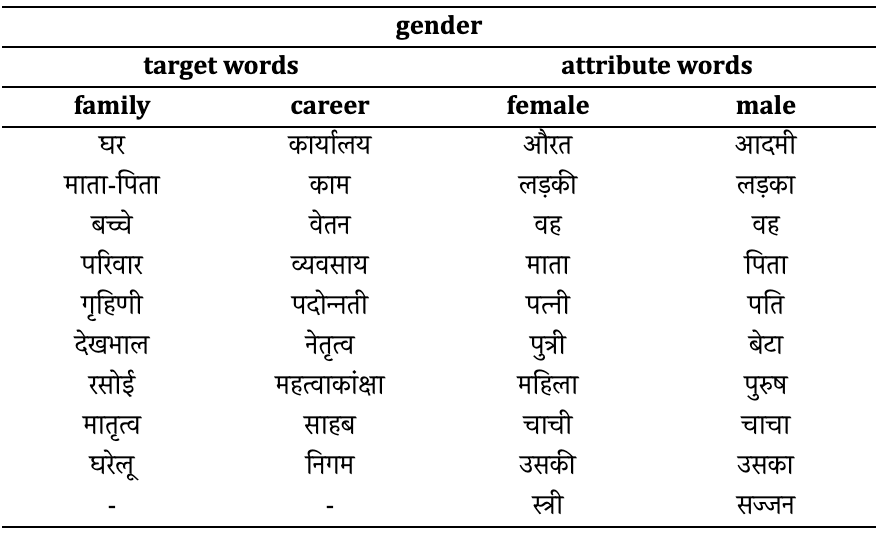}
    \caption{Manually curated bias words (target and attribute sets) for gender aspect.}
    \label{fig:bias_words_gender}
\end{figure}

\begin{figure}[htbp]
    \centering
    \includegraphics[width=\linewidth]{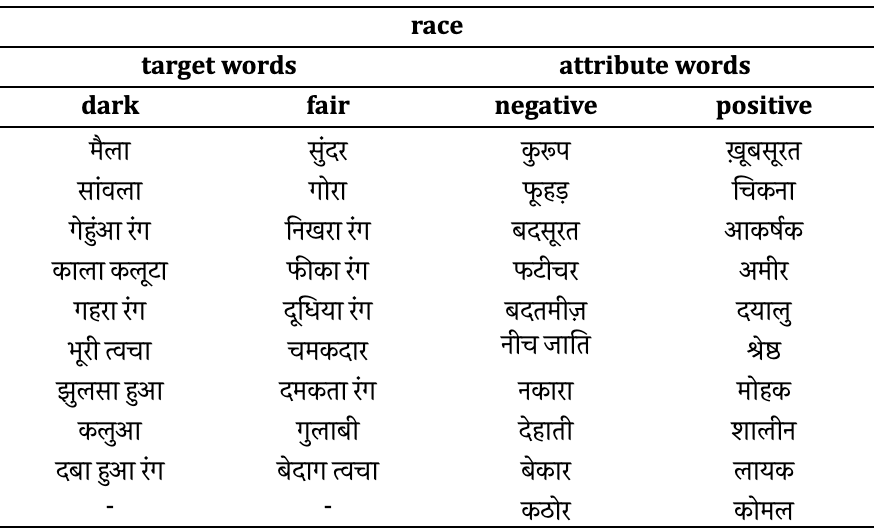}
    \caption{Manually curated bias words (target and attribute sets) for race aspect.}
    \label{fig:bias_words_race}
\end{figure}

\begin{figure}[htbp]
    \centering
    \includegraphics[width=\linewidth]{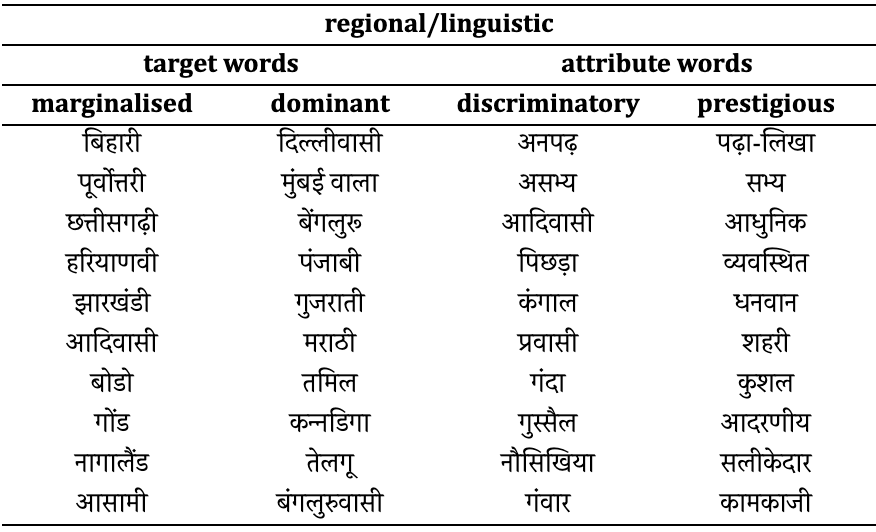}
    \caption{Manually curated bias words (target and attribute sets) for regional-linguistic aspect.}
    \label{fig:bias_words_regional_linguistic}
\end{figure}

\begin{figure}[htbp]
    \centering
    \includegraphics[width=\linewidth]{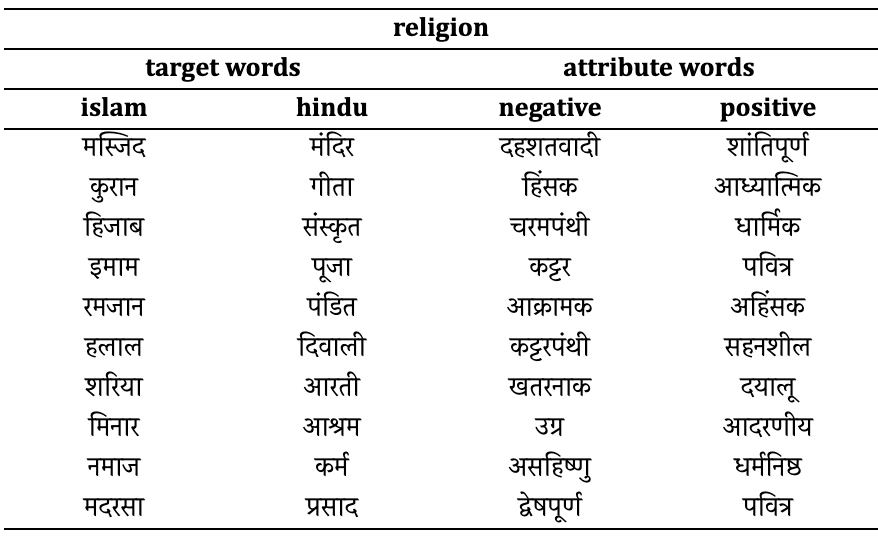}
    \caption{Manually curated bias words (target and attribute sets) for religion aspect.}
    \label{fig:bias_words_religion}
\end{figure}

\vspace{0.5em}
\noindent\textbf{Bias words: }
Bias words i.e. both target and attribute sets (around 18-20 words per set) have been manually curated for each language to capture the stereotypes. Figures \ref{fig:bias_words_caste}, \ref{fig:bias_words_gender}, \ref{fig:bias_words_race}, \ref{fig:bias_words_regional_linguistic}, \ref{fig:bias_words_religion} consist of the bias words curated for Hindi language on caste, gender, race, regional/linguistic and religion bias aspects.

\vspace{0.5em}
\noindent\textbf{Bias evaluation:}
Bias Evaluations for Hindi synthetic corpora across styles (textbook, blogpost, persona, etc.) using curated target-attribute word sets reflecting Indian sociolinguistic stereotypes are displayed in Table \ref{tab:weat_bias_results}. For each style, 1M samples (equal distribution from various sources) were evaluated, and WEAT effect sizes and scores were computed.

\begin{table*}[htbp]
\centering
\tiny
\resizebox{\textwidth}{!}{%
\begin{tabular}{llrrllp{5cm}}
\hline
\textbf{Style} & \textbf{Bias} & \textbf{Effect Size} & \textbf{WEAT Score} & \textbf{Target Sets} & \textbf{Attribute Sets} & \textbf{Observation} \\
\hline
\multirow{5}{*}{Blogpost}
& Caste & 1.00 & 0.83 & Marginalised vs Upper & Unpleasant vs Pleasant & Marginalised biased towards unpleasant \\
& Gender & 1.06 & 0.35 & Career vs Family & Male vs Female & Career biased towards male \\
& Race & 1.51 & 1.00 & Dark vs Fair & Negative vs Positive & Dark biased towards negative \\
& Region & 0.25 & 0.16 & Marginalised vs Dominant & Discriminatory vs Prestigious & Marginalised biased towards discriminatory \\
& Religion & 1.39 & 1.02 & Islam vs Hindu & Negative vs Positive & Islam biased towards negative \\
\hline
\multirow{5}{*}{Persona}
& Caste & 1.09 & 0.67 & Marginalised vs Upper & Unpleasant vs Pleasant & Marginalised biased towards unpleasant \\
& Gender & 1.15 & 0.40 & Career vs Family & Male vs Female & Career biased towards male \\
& Race & 1.11 & 0.64 & Dark vs Fair & Negative vs Positive & Dark biased towards negative \\
& Region & 0.63 & 0.47 & Marginalised vs Dominant & Discriminatory vs Prestigious & Marginalised biased towards discriminatory \\
& Religion & 0.67 & 0.36 & Islam vs Hindu & Negative vs Positive & Islam biased towards negative \\
\hline
\multirow{5}{*}{Redditpost}
& Caste & 1.00 & 0.70 & Marginalised vs Upper & Unpleasant vs Pleasant & Marginalised biased towards unpleasant \\
& Gender & 1.21 & 0.58 & Career vs Family & Male vs Female & Career biased towards male \\
& Race & 0.89 & 0.50 & Dark vs Fair & Negative vs Positive & Dark biased towards negative \\
& Region & 0.34 & 0.20 & Marginalised vs Dominant & Discriminatory vs Prestigious & Marginalised biased towards discriminatory \\
& Religion & 1.24 & 0.72 & Islam vs Hindu & Negative vs Positive & Islam biased towards negative \\
\hline
\multirow{5}{*}{Story}
& Caste & 1.00 & 0.52 & Marginalised vs Upper & Unpleasant vs Pleasant & Marginalised biased towards unpleasant \\
& Gender & 1.58 & 0.73 & Career vs Family & Male vs Female & Career biased towards male \\
& Race & 1.06 & 0.56 & Dark vs Fair & Negative vs Positive & Dark biased towards negative \\
& Region & 0.49 & 0.26 & Marginalised vs Dominant & Discriminatory vs Prestigious & Marginalised biased towards discriminatory \\
& Religion & 0.91 & 0.58 & Islam vs Hindu & Negative vs Positive & Islam biased towards negative \\
\hline
\multirow{5}{*}{Textbook}
& Caste & 0.73 & 0.59 & Marginalised vs Upper & Unpleasant vs Pleasant & Marginalised biased towards unpleasant \\
& Gender & 1.17 & 0.39 & Career vs Family & Male vs Female & Career biased towards male \\
& Race & 1.46 & 1.00 & Dark vs Fair & Negative vs Positive & Dark biased towards negative \\
& Region & 0.19 & 0.13 & Marginalised vs Dominant & Discriminatory vs Prestigious & Marginalised biased towards discriminatory \\
& Religion & 1.30 & 1.01 & Islam vs Hindu & Negative vs Positive & Islam biased towards negative \\
\hline
\multirow{5}{*}{Translation}
& Caste & 0.87 & 0.42 & Marginalised vs Upper & Unpleasant vs Pleasant & Marginalised biased towards unpleasant \\
& Gender & 1.21 & 0.51 & Career vs Family & Male vs Female & Career biased towards male \\
& Race & 1.01 & 0.41 & Dark vs Fair & Negative vs Positive & Dark biased towards negative \\
& Region & -0.13 & -0.05 & Marginalised vs Dominant & Discriminatory vs Prestigious & Reverse bias (marginalised → prestigious) \\
& Religion & 0.71 & 0.32 & Islam vs Hindu & Negative vs Positive & Islam biased towards negative \\
\hline
\multirow{5}{*}{WikiHow}
& Caste & 0.56 & 0.33 & Marginalised vs Upper & Unpleasant vs Pleasant & Marginalised biased towards unpleasant \\
& Gender & 0.92 & 0.23 & Career vs Family & Male vs Female & Career biased towards male \\
& Race & 1.28 & 0.74 & Dark vs Fair & Negative vs Positive & Dark biased towards negative \\
& Region & 0.32 & 0.19 & Marginalised vs Dominant & Discriminatory vs Prestigious & Marginalised biased towards discriminatory \\
& Religion & 1.73 & 0.91 & Islam vs Hindu & Negative vs Positive & Islam biased towards negative \\
\hline
\end{tabular}
}
\caption{WEAT bias effect sizes and scores across different synthetic generation styles in Hindi language.}
\label{tab:weat_bias_results}
\end{table*}

\vspace{0.5em}
\noindent\textbf{5. Bias mitigation}

\begin{figure}[htbp]
    \centering
    \scriptsize
\includegraphics[width=1.0\linewidth]{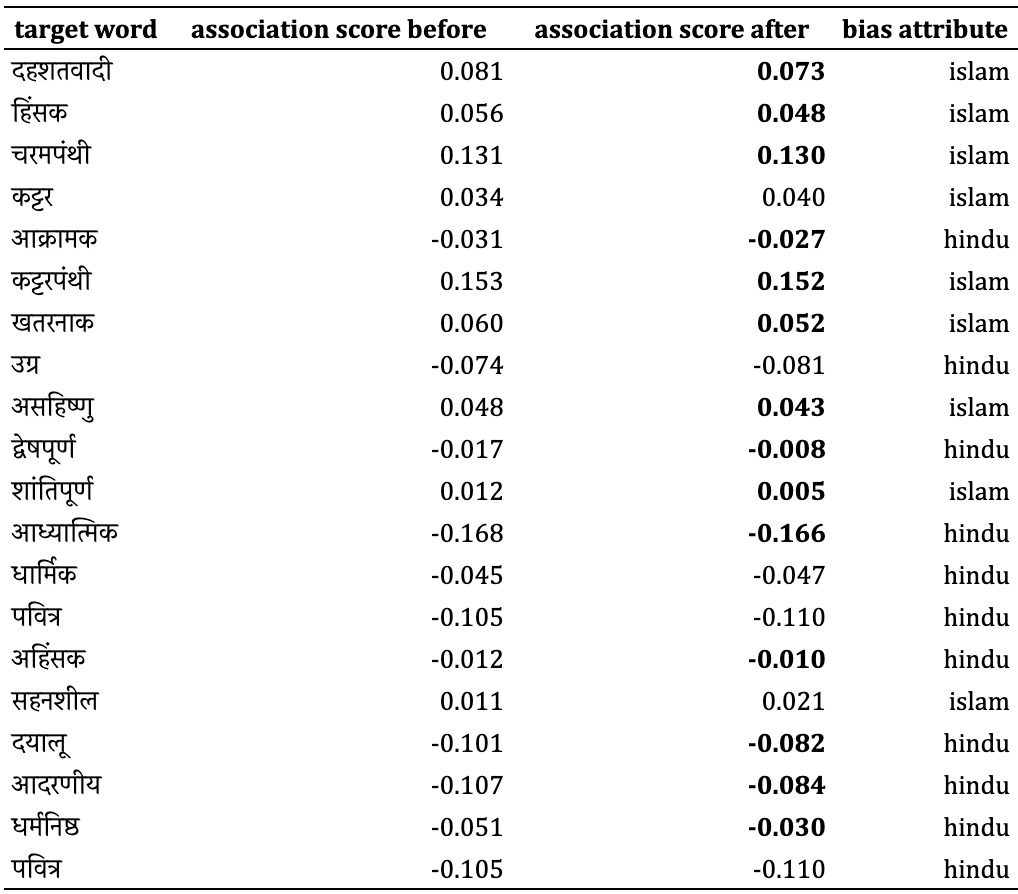} \caption{Target-word wise association scores before and after anti-biasing Hindi textbook-style examples for religious bias aspect. Bold values indicate decrease in association score after anti-biasing.}
    \label{fig:weat_anti_bias}
\end{figure}

\noindent Inspection of target-word wise association scores before and after anti-biasing for analysing the key reason for decrease in bias in the aspect of religion is displayed in Figure \ref{fig:weat_anti_bias}.

\vspace{0.5em}
\noindent\textbf{6. Bias comparison (source vs synthetic)}

\noindent Bias Evaluations for source data and synthetic data generated using that source data for religion, caste and racial bias aspects are shown in Table \ref{tab:weat_bias_source}.

\begin{table}[htbp]
\centering
\begin{tabular}{llrr}
\hline
\textbf{bias aspect} & \textbf{data} & \textbf{effect size} & \textbf{weat score} \\
\hline
religion     & source         & 1.43& 1.35 \\
             & generated data & 1.14 & 0.93 \\
caste        & source         & 1.11 & 0.84 \\
             & generated data & 0.71 & 0.54 \\
race         & source         & 1.46 & 0.79 \\
             & generated data & 1.29 & 0.77 \\
\hline
\end{tabular}
\caption{WEAT effect sizes and scores comparison between source data and synthetic data generated using that source data for each bias aspect.}
\label{tab:weat_bias_source}
\end{table}

\begin{figure}[htbp]
    \centering
\includegraphics[width=1\linewidth]{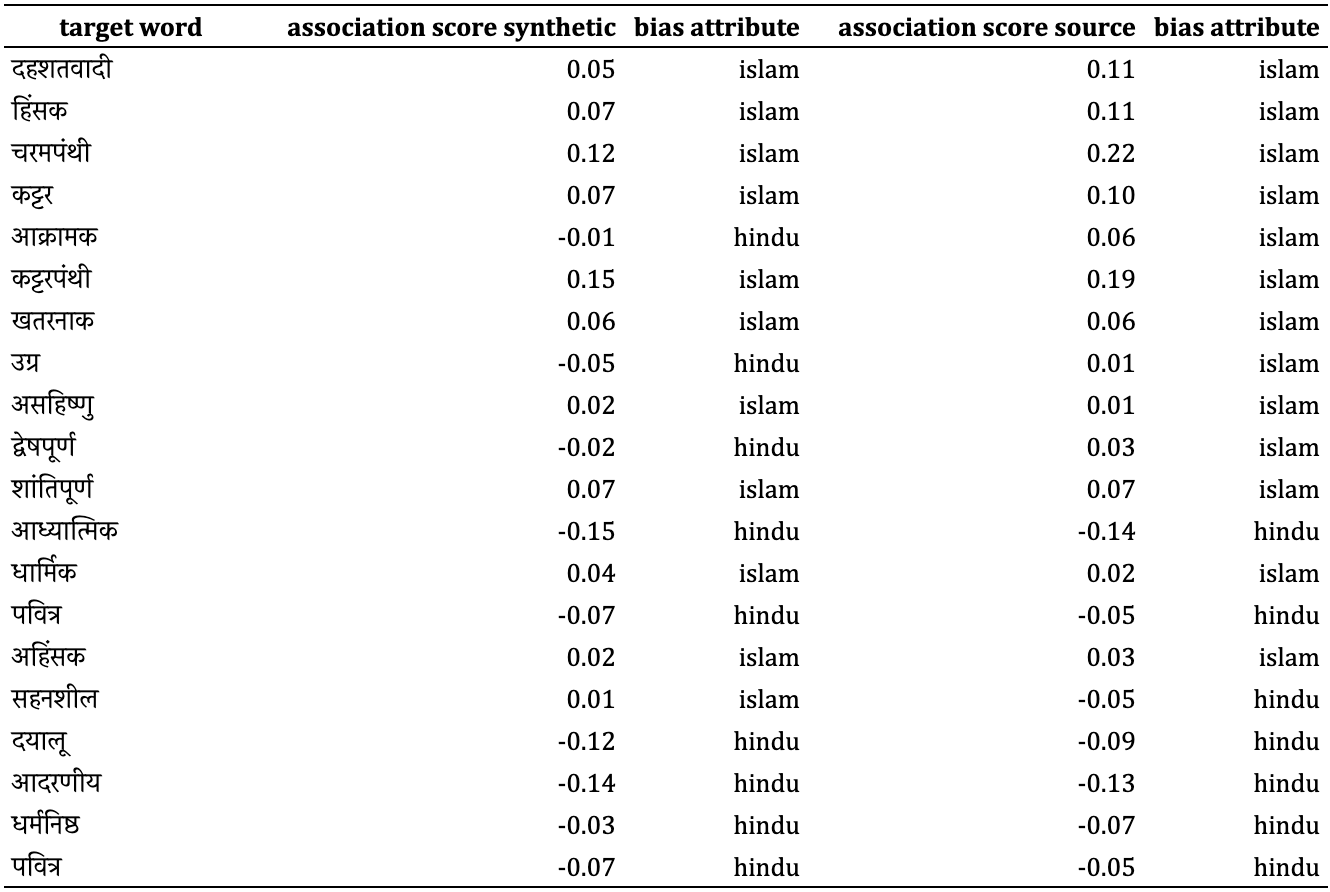} \caption{Target-word wise association scores of source and synthetic data for religious bias aspect.}
    \label{fig:weat_bias_source_religion}
\end{figure}

\begin{figure}[htbp]
    \centering
\includegraphics[width=1\linewidth]{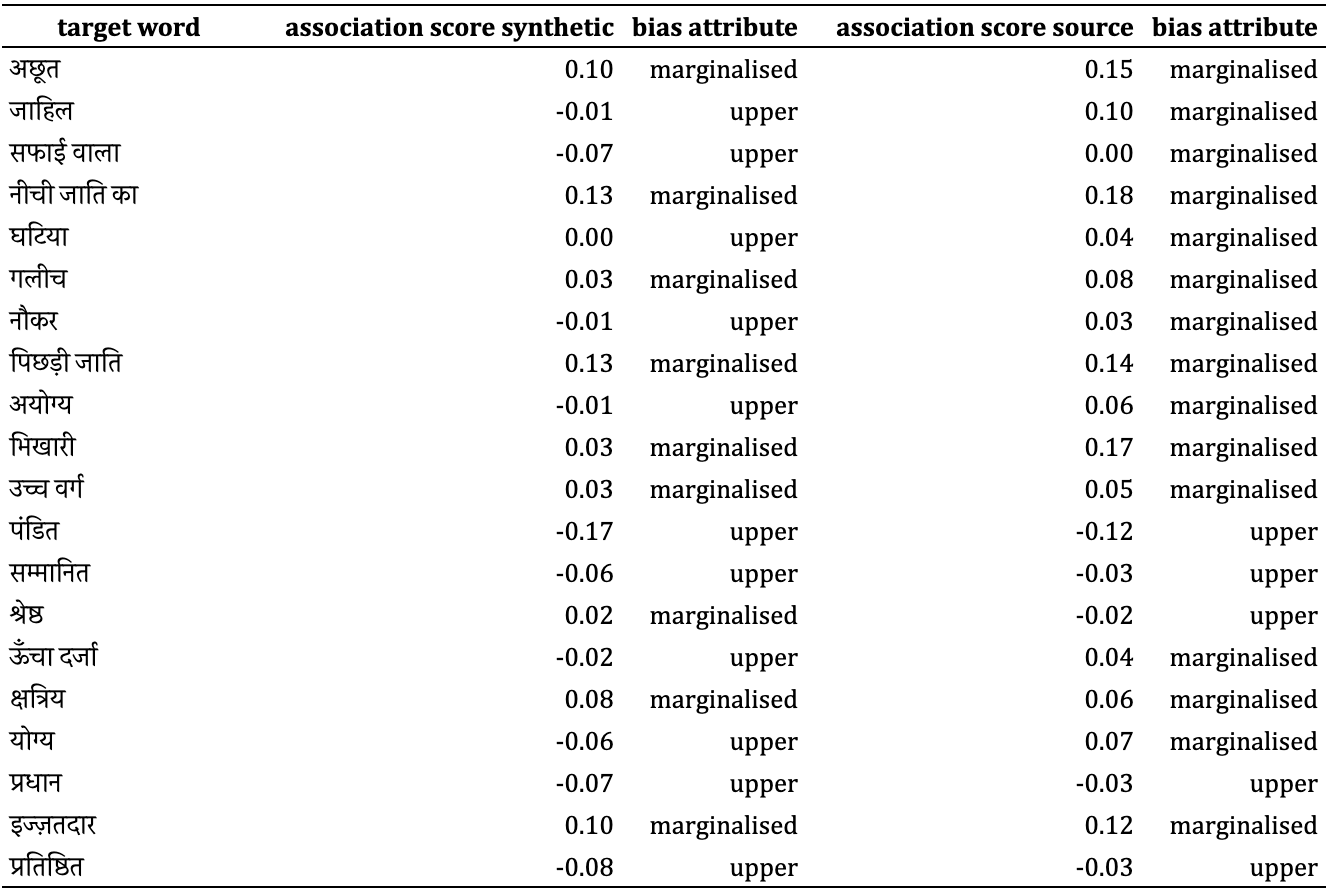} \caption{Target-word wise association scores of source and synthetic data for caste bias aspect.}
    \label{fig:weat_bias_source_caste}
\end{figure}

\begin{figure}[htbp]
    \centering
\includegraphics[width=1\linewidth]{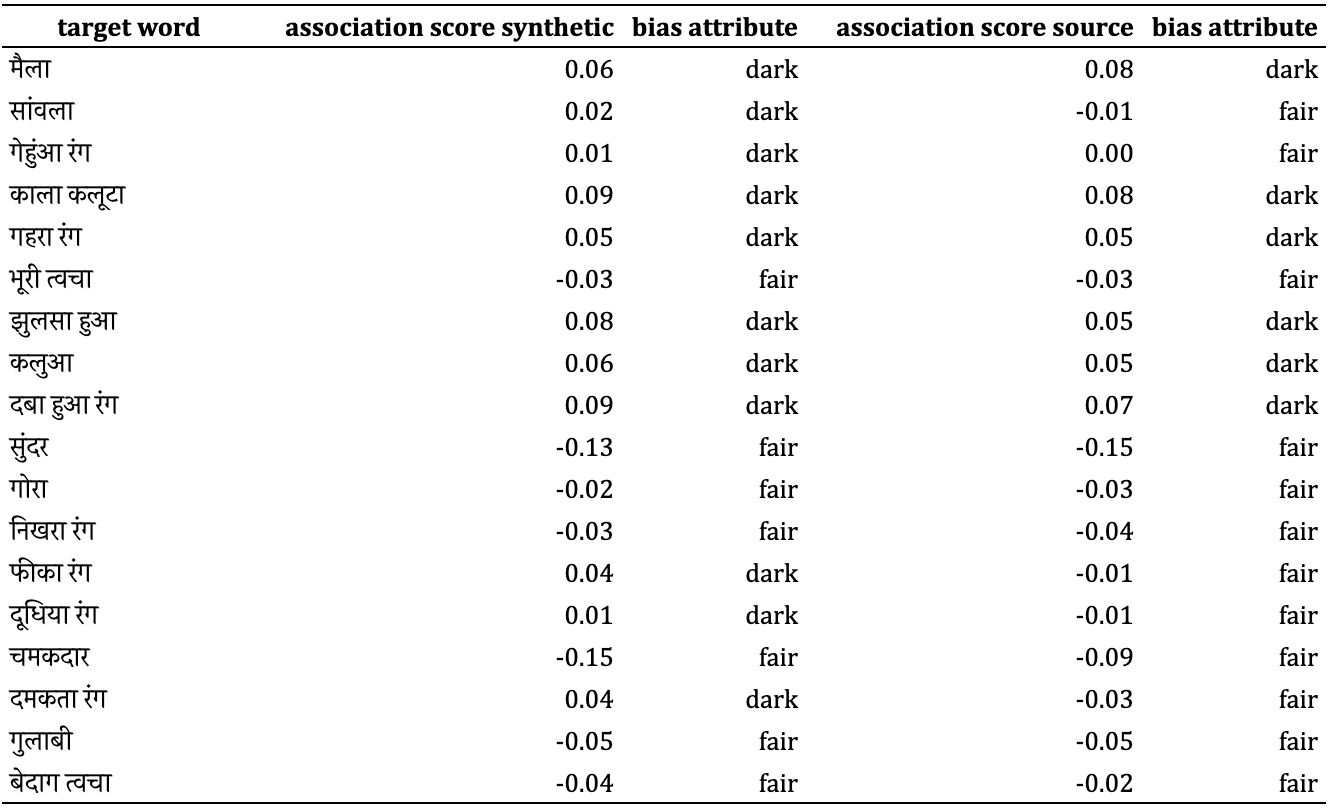} \caption{Target-word wise association scores of source and synthetic data for race bias aspect.}
    \label{fig:weat_bias_source_race}
\end{figure}

\noindent Inspection of target-word wise association scores for source data and synthetic data generated using that source data for religion, caste and racial bias aspects are shown in the below Figures \ref{fig:weat_bias_source_religion}, \ref{fig:weat_bias_source_caste}, \ref{fig:weat_bias_source_race}.

\section{Model runs}
\label{sec:model_runs}
\subsection{Implementation}

We use NeMo2 framework\footnote{\url{https://github.com/NVIDIA/NeMo}} for our model experiments and orchestrate across 16 H100 GPUs using slurm\footnote{\url{https://slurm.schedmd.com/documentation.html}}. We use Llama 3.2 1B model architecture for all our ablations. Experiment hyperparameters and model architecture is detailed in Tables 35, 36.

\begin{table}[htbp]
\centering
\resizebox{0.6\columnwidth}{!}{%
\begin{tabular}{ll}
\hline
\textbf{Model Architecture}        & \textbf{Value}       \\
\hline
Parameter count               & 1.23B                \\
Model dimension               & 2048                 \\
MLP hidden dimension          & 8192                 \\
Head dimension                & 64                   \\
Number of heads               & 32                   \\
Number of layers              & 16                   \\
Vocabulary size               & 128k                 \\
\hline
\end{tabular}
}
\caption{LLaMA 1B model architecture used for ablations.}
\label{tab:llama1b_arch}
\end{table}

\begin{table}[htbp]
\centering
\begin{tabular}{lcc}
\hline
\textbf{Hyperparameter}        & \textbf{Pretraining} & \textbf{Annealing} \\
\hline
Initial LR              & 3.00E-05             & 1.00E-05            \\
LR Scheduler            & Cosine w/ warmup     & Linear annealing    \\
Warmup Steps            & 3k                   & 0                   \\
Batch size            & 384                   & 384                   \\
\hline
\end{tabular}
\caption{Comparison of learning rate schedule and warmup settings across training runs.}
\label{tab:training_hyperparams}
\end{table}

\subsection{Evaluation}

We evaluate our models against standard English benchmarks: MMLU \citep{hendrycks-etal-2021-measuring}, GSM8K \citep{cobbe2021trainingverifierssolvemath}, Winogrande \citep{sakaguchi-etal-2021-winogrande}, Triviaqa \citep{joshi2017triviaqalargescaledistantly}, Hellaswag \citep{zellers2019hellaswagmachinereallyfinish}, Arc \citep{clark2018thinksolvedquestionanswering}, OpenbookQA \citep{mihaylov2018suitarmorconductelectricity}, CommonsenseQA \citep{talmor2019commonsenseqaquestionansweringchallenge}, DROP \citep{dua2019dropreadingcomprehensionbenchmark} and Indic Benchmarks: IndicCopa, IndicSentiment, IndicXParaphrase and IndicXNLI from IndicXtreme collection \citep{doddapaneni2023leavingindiclanguagebehind}, Arc Challenge Indic \citep{sarvamai_arcc_indic} from Indic-Evals collection and MILU \cite{verma2024milu}.\\
We use the lm-eval-harness \citep{eval-harness} framework to evaluate the models for fair and open comparison. We report EM (exact match) score for GSM8K \citep{cobbe2021trainingverifierssolvemath}, Triviaqa \citep{joshi2017triviaqalargescaledistantly}; F1-score for DROP \citep{dua2019dropreadingcomprehensionbenchmark} and Accuracy score for rest of the benchmarks.
We evaluate Arc, Arc Challenge Indic in $25$-shot, Hellaswag in $10$-shot, MILU, MMLU and Triviaqa in $5$-shot, GSM8K in $8$-shot and rest of the benchmarks in 0-shot setting.


\section{Prompts used}
\label{sec:prompts_used}

\begin{tcolorbox}[
    colback=gray!5!white,
    colframe=gray!125!black,
    title=Prompt Template for Generating Indian Personas from a Given Perspective,
    fonttitle=\bfseries,
    coltitle=black,
    boxrule=0.8pt,
    arc=3pt,
    left=6pt,
    right=6pt,
    top=6pt,
    bottom=6pt
]

\textbf{You are an AI assistant. Follow the below guidelines.}

\vspace{0.5em}

\textbf{Input:}
\begin{itemize}[leftmargin=1.5em]
    \item \texttt{input\_persona}: \texttt{\{persona\}} — The persona whose perspective, context, and potential biases should be adopted.
\end{itemize}

\textbf{Objective:} Generate multiple brief descriptions of distinct Indian personas, making them relevant to and subtly reflecting the perspective of the \texttt{input\_persona}.

\vspace{0.5em}

\textbf{Instructions:}
\begin{enumerate}[leftmargin=1.5em]
    \item Embody the \texttt{input\_persona}.
    \item Create 5–10 brief (1–2 sentence) descriptions of different individuals in India.
    \item For each description:
    \begin{itemize}
        \item Reflect the \texttt{input\_persona}'s subtle perspective and tone.
        \item Include authentic Indian details (e.g., profession, age group, location, concerns).
        \item Keep it gender-neutral, especially for roles like teacher, chef, or healthcare worker.
        \item \textbf{Do not} use any names.
    \end{itemize}
\end{enumerate}

\vspace{0.5em}

\textbf{Output Requirements:}
\begin{itemize}[leftmargin=1.5em]
    \item \textbf{Content:} Provide \textit{only} the list of generated persona descriptions. Exclude any introductory or concluding text or labels.
    \item \textbf{Quantity:} Deliver 5–10 distinct persona descriptions.
    \item \textbf{Format:} Use a \textbf{Markdown bulleted list}. Each bullet point should contain one persona description.
\end{itemize}

\label{box:prompt-persona-generation}
\end{tcolorbox}

\begin{tcolorbox}[
    colback=gray!5!white,
    colframe=gray!125!black,
    title=Prompt Template for Generating a Wikihow article,
    fonttitle=\bfseries,
    coltitle=black,
    boxrule=0.8pt,
    arc=3pt,
    left=6pt,
    right=6pt,
    top=6pt,
    bottom=6pt
]

Here is an extract from a webpage: \textquotedblleft\{\texttt{extract}\}\textquotedblright.
Write a long and very detailed tutorial that could be part of WikiHow whose title is related to the extract above\{\texttt{topic}\}. Include in depth explanations for each step and how it helps achieve the desired outcome, including key tips and guidelines. 
Ensure clarity and practicality, allowing readers to easily follow and apply the instructions. Do not use images. Do not use phrases like 'in the above extract', 'as per the extract' etc. Don't focus on irrelevant information in the web extract.
\newline
Generate text in \texttt{\{language\}} language. Ensure that the output is entirely in \texttt{\{script\}} script and avoid using any English words or \texttt{\{language\}} in Latin script.

\label{box:prompt-cosmopedia-wikihow}
\end{tcolorbox}

\begin{tcolorbox}[
    colback=gray!5!white,
    colframe=gray!125!black,
    title=Prompt Template for Generating a Blogpost,
    fonttitle=\bfseries,
    coltitle=black,
    boxrule=0.8pt,
    arc=3pt,
    left=6pt,
    right=6pt,
    top=6pt,
    bottom=6pt
]

Here is an extract from a webpage: \textquotedblleft\{\texttt{extract}\}\textquotedblright.
Write an informative and insightful blog post that expands upon the extract above\{\texttt{topic}\}. Your post should delve into the nuances of the topic, offering fresh perspectives and deeper analysis. Aim to:
\begin{itemize}[leftmargin=1.5em]
    \item Inform: Provide valuable, well-researched information that educates the reader.
    \item  Engage: Write in a conversational tone that connects with the audience, making complex ideas accessible.
    \item  Illustrate: Use examples, anecdotes, or personal experiences to bring the topic to life.
\end{itemize}
Do not give a title and do not start with sentences like "Have you ever..." or "Hello dear readers..", simply write the content without these introductory phrases. Do not use phrases like 'in the above extract', 'as per the extract' etc. Don't focus on irrelevant information in the web extract.
\newline
Generate text in \texttt{\{language\}} language. Ensure that the output is entirely in \texttt{\{script\}} script and avoid using any English words or \texttt{\{language\}} in Latin script.

\label{box:prompt-cosmopedia-blogpost}
\end{tcolorbox}

\begin{tcolorbox}[
    colback = gray!5! white, colframe = gray!125!black, title = Prompt Template for Persona-Based Generation,
    fonttitle=\bfseries,
    coltitle=black,
    boxrule=0.8pt,
    arc=3pt,
    left=6pt,
    right=6pt,
    top=6pt,
    bottom=6pt
]
\textbf{You are an AI assistant. Follow the below guidelines.} 

\vspace{0.5em}
\textbf{**Input**:}\\
- Persona Description: \{persona\} \\
- Target Language: \{lang\} \\

\textbf{**Task**:} Assume the role of the specified Indian persona, deeply rooted in an Indian context. Generate an article embodying this persona's perspective and knowledge within India.

\vspace{0.5em}
\textbf{**Generated Text Specifications**:}
\begin{enumerate}[leftmargin=1.5em]
    \item \textbf{Language:} Author \textbf{exclusively} in Target \{lang\} Language.
    \item \textbf{Vocabulary:} No translation and transliteration.
    \item \textbf{Voice \& Tone:} Authentically reflect persona (\texttt{persona}) in \{lang\}, specific to Indian context.
    \item \textbf{Cultural Accuracy:} Integrate relevant Indian cultural norms and social context.
    \item \textbf{Contextual Grounding:} Base all examples, steps, and advice on practical Indian realities.
\end{enumerate}

\vspace{0.5em}
\textbf{**Final Output Requirements**:} The text must be authentic, useful for the target audience in India, and strictly generated \textbf{in the specified language}, clearly reflecting the persona.
\label{box:prompt-persona}
\end{tcolorbox}

\begin{tcolorbox}[
    colback=gray!5!white,
    colframe=gray!125!black,
    title=Prompt Template for Generating an Academic Style Textbook,
    fonttitle=\bfseries,
    coltitle=black,
    boxrule=0.8pt,
    arc=3pt,
    left=6pt,
    right=6pt,
    top=6pt,
    bottom=6pt
]

Here is an extract from a webpage: \textquotedblleft\{\texttt{extract}\}\textquotedblright.
Write an extensive and detailed course unit suitable for a textbook targeted at college students related to the given extract\{\texttt{topic}\}. Do not just list concepts, but develop each one in detail before moving to the next, as we prioritize depth of understanding and comprehensive exploration of the subject matter over breadth. Focus on:
\begin{itemize}[leftmargin=1.5em]
    \item  Rigor: Ensure in-depth coverage of the concepts/sections.
    \item  Engagement: Write with an academic, professional and engaging tone that captivates interest.
    \item  Application: Incorporate specific, practical examples, such as proofs in calculus or critical dates and figures in history.
\end{itemize}
Do not include a title or an introduction, simply write the content without headlines and introductory phrases. Do not use images. Do not use phrases like 'in the above extract', 'as per the extract' etc. Don't focus on irrelevant information in the web extract.
\newline
Generate text in \texttt{\{language\}} language. Ensure that the output is entirely in \texttt{\{script\}} script and avoid using any English words or \texttt{\{language\}} in Latin script.

\label{box:prompt-cosmopedia-textbook-academic}
\end{tcolorbox}

\begin{tcolorbox}[
    colback=gray!5!white,
    colframe=gray!125!black,
    title=Prompt Template for Generating a Textbook Narrative,
    fonttitle=\bfseries,
    coltitle=black,
    boxrule=0.8pt,
    arc=3pt,
    left=6pt,
    right=6pt,
    top=6pt,
    bottom=6pt
]

Here is an extract from a webpage: \textquotedblleft\{\texttt{extract}\}\textquotedblright.
Write an extensive and detailed course unit suitable for a textbook related to the given extract\{\texttt{topic}\}. Do not just list concepts, but develop each one in detail before moving to the next, as we prioritize depth of understanding and comprehensive exploration of the subject matter over breadth. Focus on:
\begin{itemize}[leftmargin=1.5em]
    \item  Rigor: Ensure in-depth coverage of the concepts.
    \item  Engagement: Use a narrative style akin to Michael Lewis, making it captivating and thought-provoking.
    \item  Relevance: Connect the topic with current trends, real-life examples, or recent studies. Do not use images.
\end{itemize}
Do not include a title or an introduction, simply write the content without headlines and introductory phrases. Do not use images. Do not use phrases like 'in the above extract', 'as per the extract' etc. Don't focus on irrelevant information in the web extract.
\newline
Generate text in \texttt{\{language\}} language. Ensure that the output is entirely in \texttt{\{script\}} script and avoid using any English words or \texttt{\{language\}} in Latin script.

\label{box:prompt-cosmopedia-textbook-narrative}
\end{tcolorbox}

\begin{tcolorbox}[
    colback=gray!5!white,
    colframe=gray!125!black,
    title=Prompt Template for Story for Young Children with Scientific Insight,
    fonttitle=\bfseries,
    coltitle=black,
    boxrule=0.8pt,
    arc=3pt,
    left=6pt,
    right=6pt,
    top=6pt,
    bottom=6pt
]

Write an educational story (3–5 paragraphs) targeted at young children using simple words. The story should be inspired from this text snippet: \textquotedblleft\{\texttt{extract}\}\textquotedblright. 

The story doesn’t have to address everything in the snippet, it is there just for inspiration.

The story should have the following features:
\begin{itemize}[leftmargin=1.5em]
    \item Science Integration: Embed basic science concepts within the story, explaining them through the characters' adventures and discoveries.
    \item Dialogue: Include at least one dialogue and insightful conversation.
    \item Unexpected Twist: Conclude with a twist that doesn't resolve as hoped, but leaves a clear lesson about life and science.
\end{itemize}
Do not start with classic sentences like "Once upon a time", be creative.
\newline
Generate text in \texttt{\{language\}} language. Ensure that the output is entirely in \texttt{\{script\}} script and avoid using any English words or \texttt{\{language\}} in Latin script.

\label{box:prompt-cosmopedia-young_children_story}
\end{tcolorbox}

\begin{tcolorbox}[
    colback = gray!5! white, colframe = gray!125!black, title = Prompt Template for Generation of Persona-Based Articles with Seed Data,
    fonttitle=\bfseries,
    coltitle=black,
    boxrule=0.8pt,
    arc=3pt,
    left=6pt,
    right=6pt,
    top=6pt,
    bottom=6pt
]

\textbf{You are an AI assistant. Follow the below guidelines.}

\vspace{0.5em}

\textbf{Input:}
\begin{itemize}[leftmargin=1.5em]
    \item Seed Text (for context): \texttt{\{seed\_text\}}
    \item Persona Description: \texttt{\{persona\}}
    \item Target Language: \texttt{\{language\}}
\end{itemize}

\textbf{Task:} Assume the role of the specified Indian persona (\texttt{\{persona\}}), deeply rooted in an Indian context. Generate an article embodying this persona's perspective and given knowledge, within India.

\vspace{0.5em}

\textbf{Generated Article Specifications:}
\begin{enumerate}[leftmargin=1.5em]
    \item \textbf{Language:} Author \textbf{exclusively} in target language (\texttt{\{language\}}).
    \item \textbf{Voice \& Tone:} Authentically reflect persona (\texttt{\{persona\}}) in \texttt{\{language\}}, specific to Indian context.
    \item \textbf{Topic Relevance:} Subject must be pertinent to the persona within India.
    \item \textbf{Cultural Accuracy:} Integrate relevant Indian cultural norms and social context.
    \item \textbf{Contextual Grounding:} Base all examples, steps, and advice on practical Indian realities.
\end{enumerate}

\vspace{0.5em}

\textbf{Final Output Check:} The article must be authentic, useful for the target audience in India, and strictly generated \textbf{in the specified \texttt{\{language\}} language}, clearly reflecting the persona \texttt{\{persona\}}.

\label{box:prompt-wikihow}
\end{tcolorbox}

\begin{tcolorbox}[
    colback=gray!5!white,
    colframe=gray!125!black,
    title=Prompt Template for Morality-Focused Narrative,
    fonttitle=\bfseries,
    coltitle=black,
    boxrule=0.8pt,
    arc=3pt,
    left=6pt,
    right=6pt,
    top=6pt,
    bottom=6pt
]

Write a compelling story related to the following text snippet: \textquotedblleft\{\texttt{extract}\}\textquotedblright.

The story doesn’t need to mention everything in the snippet—use it just for inspiration and be creative!

The story should incorporate the following elements:
\begin{itemize}[leftmargin=1.5em]
    \item Dialogue: Include at least one meaningful dialogue that reveals character depth or unravels a crucial piece of the mystery.
    \item Interesting Themes: Explore themes such as moral ambiguity, existential queries, personal transformation, or consequences of past actions.
\end{itemize}
Do not start with classic sentences like "Once upon a time", "The sun hung low in the sky", or "In the dimly lit", be creative.
\newline
Generate text in \texttt{\{language\}} language. Ensure that the output is entirely in \texttt{\{script\}} script and avoid using any English words or \texttt{\{language\}} in Latin script.

\label{box:prompt-cosmopedia-morality_story}
\end{tcolorbox}

\begin{tcolorbox}[
    colback=gray!5!white,
    colframe=gray!125!black,
    title=Prompt Template for Story Emphasizing Problem Solving,
    fonttitle=\bfseries,
    coltitle=black,
    boxrule=0.8pt,
    arc=3pt,
    left=6pt,
    right=6pt,
    top=6pt,
    bottom=6pt
]

Write a story that explores a situation slightly related to this text snippet: \textquotedblleft\{\texttt{extract}\}\textquotedblright.

The story should unfold through the characters’ interactions, decisions, and consequences. Emphasize problem-solving, common-sense lessons, and social cues. It should cater to a diverse age group, include at least one dialogue, and present both positive and negative outcomes.

Do not start with classic sentences like "Once upon a time", be creative.
\newline
Generate text in \texttt{\{language\}} language. Ensure that the output is entirely in \texttt{\{script\}} script and avoid using any English words or \texttt{\{language\}} in Latin script.

\label{box:prompt-cosmopedia-problem_solving_story}
\end{tcolorbox}

\begin{tcolorbox}[
    colback=gray!5!white,
    colframe=gray!125!black,
    title=Prompt Template for Forum-Style Life Story,
    fonttitle=\bfseries,
    coltitle=black,
    boxrule=0.8pt,
    arc=3pt,
    left=6pt,
    right=6pt,
    top=6pt,
    bottom=6pt
]

Write a story in the style of real-life situations shared in forums. The story should be somehow related to this text snippet: \textquotedblleft\{\texttt{extract}\}\textquotedblright.

The story should feature:
\begin{itemize}[leftmargin=1.5em]
    \item A compelling and unexpected plot twist
    \item Authenticity and emotional depth like personal forum posts
    \item Relatable events and human complexity
\end{itemize}
Do not start with classic sentences like "Once upon a time", "A few years back", or "A few months ago", be creative.
\newline
Generate text in \texttt{\{language\}} language. Ensure that the output is entirely in \texttt{\{script\}} script and avoid using any English words or \texttt{\{language\}} in Latin script.

\label{box:prompt-cosmopedia-forums_story}
\end{tcolorbox}

\begin{tcolorbox}[
    colback=gray!5!white,
    colframe=gray!125!black,
    title=Prompt Template for Reddit-Style Real-Life Post,
    fonttitle=\bfseries,
    coltitle=black,
    boxrule=0.8pt,
    arc=3pt,
    left=6pt,
    right=6pt,
    top=6pt,
    bottom=6pt
]

Write a real-life story shared by someone in a Reddit forum. The story should be somehow related to this text snippet: \textquotedblleft\{\texttt{extract}\}\textquotedblright.

The story should include:
\begin{itemize}[leftmargin=1.5em]
    \item Niche Interests or Humor: Focus on specific hobbies, interests, or amusing situations.
    \item An Unexpected Plot Twist or Conflict: Introduce a relatable yet challenging situation.
    \item Reflection and Insight: Conclude with a personal revelation or communal connection.
\end{itemize}
Start the story right away. Do not begin with sentences like "Once upon a time", "A few years ago", or "A few years back", be creative.
\newline
Generate text in \texttt{\{language\}} language. Ensure that the output is entirely in \texttt{\{script\}} script and avoid using any English words or \texttt{\{language\}} in Latin script.

\label{box:prompt-cosmopedia-reddit_post}
\end{tcolorbox}

\begin{tcolorbox}[
    colback = gray!5! white, 
    colframe = gray!125!black, title = Prompt Template for Persona-Based Indicate Generation with Seed as Persona,
    fonttitle=\bfseries,
    coltitle=black,
    boxrule=0.8pt,
    arc=3pt,
    left=6pt,
    right=6pt,
    top=6pt,
    bottom=6pt
]

\textbf{You are an AI assistant. Follow the below guidelines.}

\vspace{0.5em}

\textbf{Input:}
\begin{itemize}[leftmargin=1.5em]
    \item Persona Description: \texttt{\{persona\}} (The persona whose perspective, context, and potential biases should be adopted).
\end{itemize}

\textbf{Objective:} Generate multiple brief descriptions of distinct Indian personas, making them relevant to and subtly reflecting the perspective of the \texttt{\{persona\}}.

\vspace{0.5em}

\textbf{Instructions:}
\begin{enumerate}[leftmargin=1.5em]
    \item Embody the \texttt{\{persona\}}.
    \item Create 5-10 brief (1-2 sentences) descriptions of different individuals in India.
    \item For each description: \begin{itemize}
        \item Reflect the \texttt{\{persona\}}'s subtle perspective and tone.
        \item Include authentic Indian details (e.g., profession, age group, location, concerns).
        \item Keep it gender-neutral, especially for roles like teacher, chef, or healthcare worker.
        \item Do \textbf{not} use any names.
    \end{itemize}
\end{enumerate}

\vspace{0.5em}

\textbf{Output Requirements:}
\begin{enumerate}[leftmargin=1.5em]
    \item \textbf{Content:} Provide \textit{only} the list of generated persona descriptions. Exclude any introductory or concluding text or labels.
    \item \textbf{Quantity:} Deliver 5-10 distinct persona descriptions.
    \item \textbf{Format:} Use a \textbf{Markdown bulleted list}. Each bullet point should contain one persona description.
\end{enumerate}

\end{tcolorbox}

\begin{tcolorbox}[
    colback=gray!5!white,
    colframe=gray!125!black,
    title=Prompt Template for generating Math textbook sections,
    fonttitle=\bfseries,
    coltitle=black,
    boxrule=0.8pt,
    arc=3pt,
    left=6pt,
    right=6pt,
    top=6pt,
    bottom=6pt
]

You are given a math question and its solution:

Question: \{\texttt{question}\} \newline
Solution: \{\texttt{solution}\}

Create an academic textbook section based on this Q/A pair, by following these instructions:
\begin{enumerate}
    \item Start with a section title that reflects the broad mathematical concepts involved in solving the question. The title should not refer specifically to the question, but rather to the general topic it belongs to (e.g., "Linear Equations" instead of "Finding the Value of x").

    \item In the section, explain the relevant mathematical principles, methods, and definitions in depth. Each concept should be introduced and developed fully before moving to the next. Do not simply list concepts — build understanding by expanding on each idea thoroughly.

    \item Use an academic, professional, and engaging tone that resembles college-level textbooks. Prioritize clarity, rigor, and depth over brevity.

    \item After the conceptual explanation, include the given question as an exercise and present a detailed, step-by-step solution using sound mathematical reasoning. The solution should be complete and clean, and must not include informal phrases or concluding remarks like “this shows that” or “as seen above.”
                 
    \item Generate text in \texttt{\{language\}} language. Ensure that the output is entirely in \texttt{\{script\}} script and avoid using any English words or \texttt{\{language\}} in Latin script.
\end{enumerate}

The output should read like a cohesive, standalone textbook section that educates the reader on the topic, illustrates it through the example, and reinforces understanding via rigorous solution steps. Ensure you generate text in \texttt{\{language\}} language and use \texttt{\{script\}} script throughout.

\label{box:prompt-math-textbook-section}
\end{tcolorbox}

\newpage
\section{Guidelines For Manual Annotation}
\label{sec:guidelines-manual-annotation}

In the following, we provide the guidelines provided to the human annotators for evaluating both generation as well as translation across different models. 


\begin{tcolorbox}[
    colback=gray!5!white,
    colframe=gray!125!black,
    title=Guidelines for Evaluating Sample Synthetic Data – Indian Context,
    fonttitle=\bfseries,
    coltitle=black,
    boxrule=0.8pt,
    arc=3pt,
    left=6pt,
    right=6pt,
    top=6pt,
    bottom=6pt
]

We aim to evaluate large language models (LLMs) for generating high-quality textbook-like content in 22 Indian languages. Provided prompt-response pairs should be annotated using the following five-point criteria:

\begin{enumerate}[leftmargin=*]
  \item \textbf{Grammar \& Readability (1–5)} \\
  Evaluate how natural and grammatically correct the writing is.
  \begin{itemize}
    \item 1 – Poor: Hard to read; grammar issues, awkward phrasing.
    \item 3 – Acceptable: Understandable with minor flaws.
    \item 5 – Excellent: Flawless grammar and fluent structure.
  \end{itemize}

  \item \textbf{Faithfulness to the Prompt (1–5)} \\
  Assess whether the response stays on-topic and fulfills the instruction.
  \begin{itemize}
    \item 1 – Poor: Largely off-topic or missing key ideas.
    \item 3 – Acceptable: Mostly aligned with prompt intent.
    \item 5 – Excellent: Fully relevant and complete response.
  \end{itemize}

  \item \textbf{Overall Generation Quality (1–5)} \\
  Judge coherence, structure, title quality, and overall flow.
  \begin{itemize}
    \item 1 – Poor: Unstructured, incoherent, or dull.
    \item 3 – Acceptable: Decent flow and organization.
    \item 5 – Excellent: Clear, compelling, and well-structured.
  \end{itemize}

  \item \textbf{Hallucination / Factual Accuracy (1–5)} \\
  Rate the factual correctness and reliability of the content.
  \begin{itemize}
    \item 1 – Poor: Contains major factual errors or fabrications.
    \item 3 – Acceptable: Mostly accurate, with minor mistakes.
    \item 5 – Excellent: Fully factual and verifiable content.
  \end{itemize}

  \item \textbf{Length of Output (1–5)} \\
  Evaluate if the output is sufficiently detailed and balanced.
  \begin{itemize}
    \item 1 – Too Short: $<$10 lines; lacks depth.
    \item 3 – Moderate: 20–40 lines; informative but could improve.
    \item 5 – Ideal: Well-balanced and comprehensive.
  \end{itemize}
\end{enumerate}
\label{guideline:gen-v-trans}
\end{tcolorbox}

\begin{tcolorbox}[
    colback=gray!5!white,
    colframe=gray!125!black,
    title=Guidelines for Evaluating Translations by LLMs sections,
    fonttitle=\bfseries,
    coltitle=black,
    boxrule=0.8pt,
    arc=3pt,
    left=6pt,
    right=6pt,
    top=6pt,
    bottom=6pt
]

The goal is to evaluate the quality of machine-generated translations of educational content across Indian languages. Each translation is to be rated on the following 4-point criteria, using a 1–5 scale:

\begin{enumerate}[leftmargin=*]
  \item \textbf{Grammar \& Readability} \\
  Judge how grammatically correct and naturally flowing the translation is in the target language.
  \begin{itemize}
    \item 1 – Poor: Hard to read due to grammar issues or awkward phrasing.
    \item 3 – Acceptable: Mostly clear with minor issues.
    \item 5 – Excellent: Native-like fluency with polished flow.
  \end{itemize}

  \item \textbf{Translation Faithfulness} \\
  Assess how accurately the original meaning is preserved.
  \begin{itemize}
    \item 1 – Poor: Major distortions or omissions.
    \item 3 – Acceptable: Mostly faithful with minor deviations.
    \item 5 – Excellent: Fully equivalent in meaning.
  \end{itemize}

  \item \textbf{Terminology \& Domain Consistency} \\
  Evaluate the use of domain-specific terms and formatting.
  \begin{itemize}
    \item 1 – Poor: Inconsistent or incorrect terminology.
    \item 3 – Acceptable: Mostly correct, some inconsistencies.
    \item 5 – Excellent: Consistently accurate and well-suited to context.
  \end{itemize}

  \item \textbf{Fluency \& Style} \\
  Consider stylistic smoothness, tone, and sentence flow.
  \begin{itemize}
    \item 1 – Poor: Disjointed or robotic.
    \item 3 – Acceptable: Understandable but slightly mechanical.
    \item 5 – Excellent: Natural, fluent, and human-like.
  \end{itemize}
\end{enumerate}
\label{guideline:translation}
\end{tcolorbox}

\newpage
\vspace{0.5em}
\section{Examples}
\label{sec:examples}

\begin{figure*}[htbp]
    \centering
    \scriptsize
\includegraphics[width=1.0\linewidth]{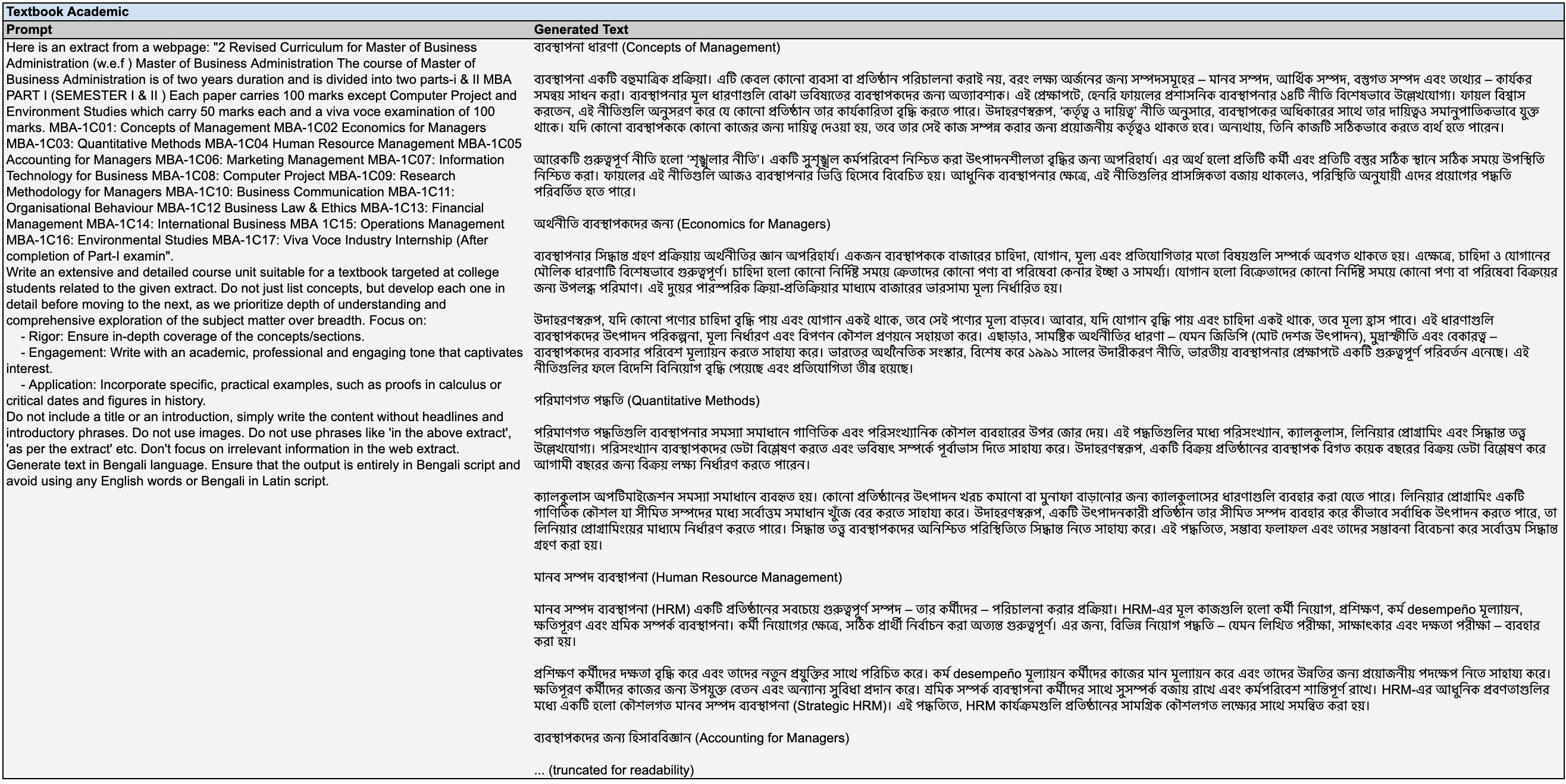} \caption{Textbook Academic style Example.}
    \label{fig:textbook_academic_example}
\end{figure*}

\begin{figure*}[htbp]
    \centering
    \scriptsize
\includegraphics[width=1.0\linewidth]{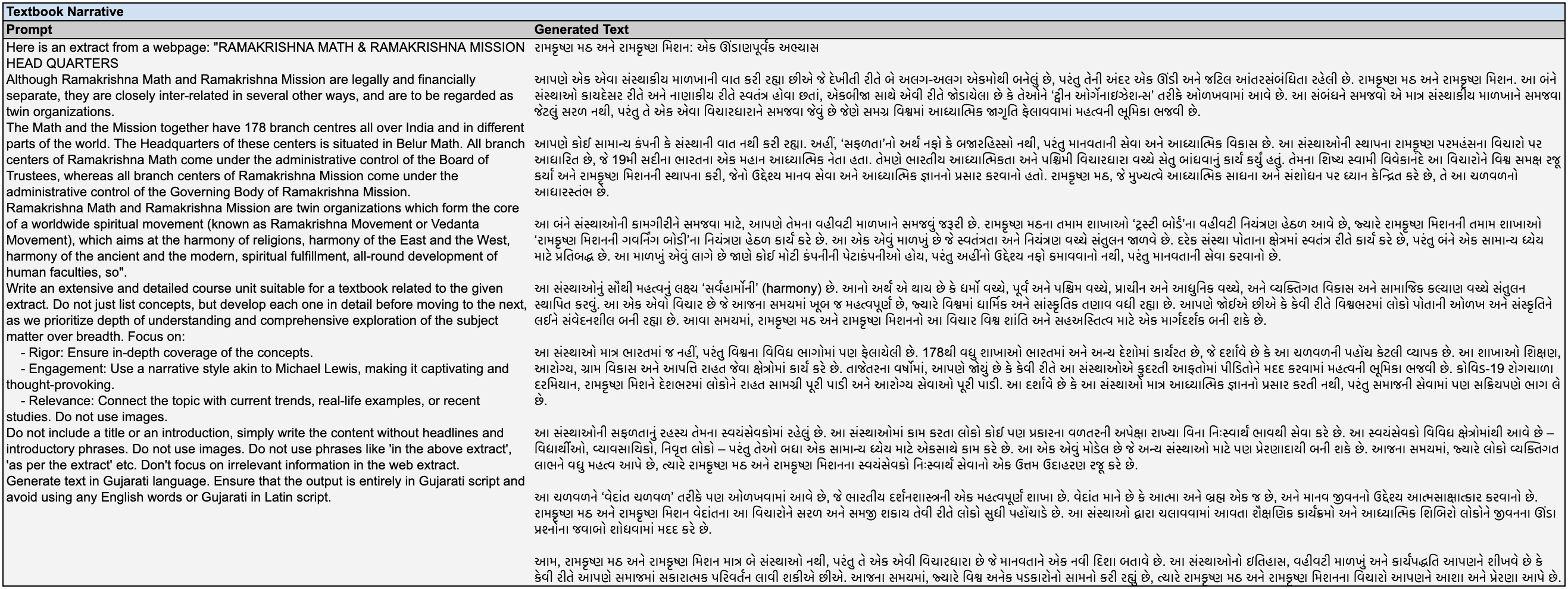} \caption{Textbook Narrative style Example.}
    \label{fig:textbook_narrative_example}
\end{figure*}

\begin{figure*}[htbp]
    \centering
    \scriptsize
\includegraphics[width=1.0\linewidth]{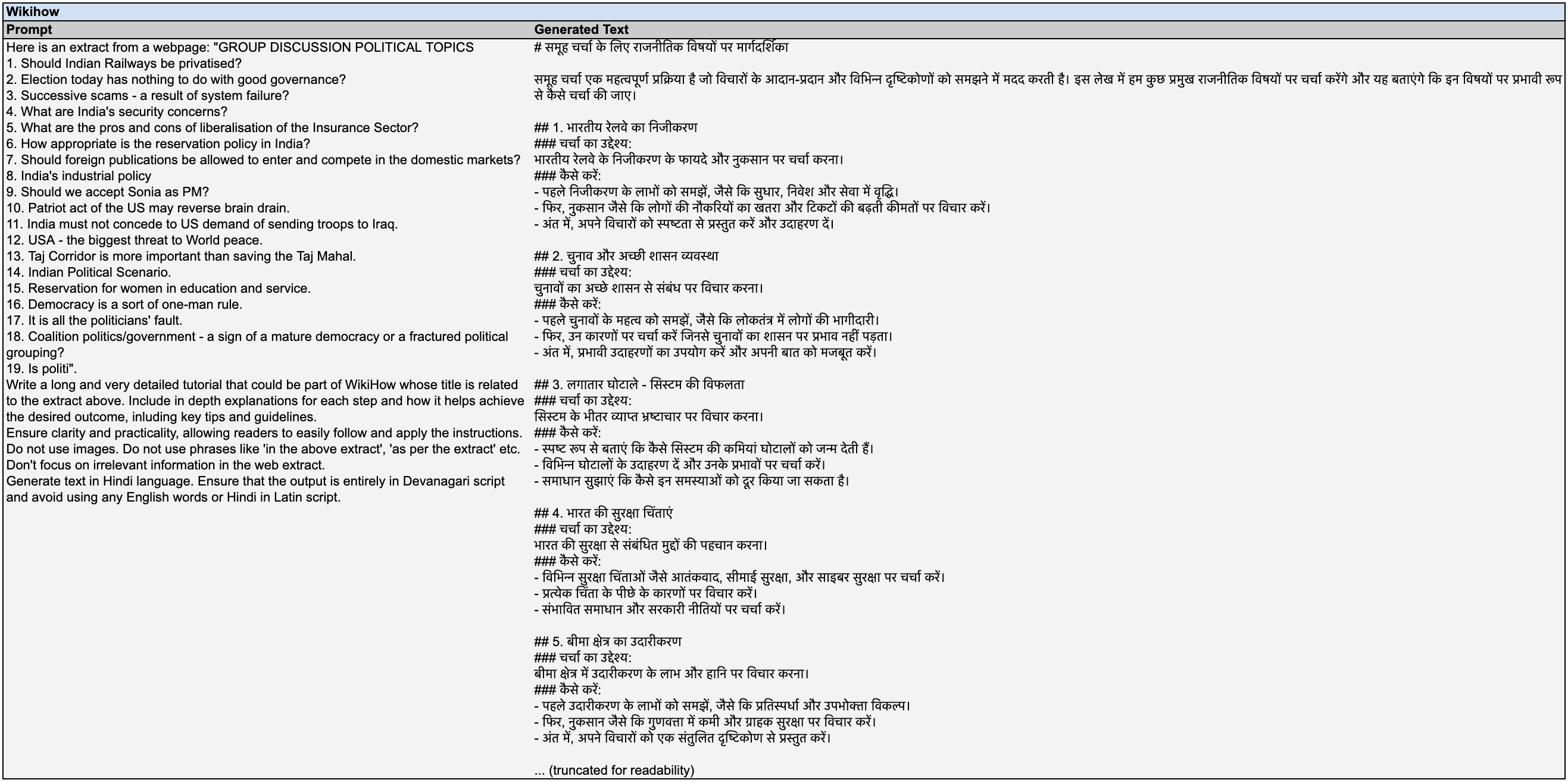} \caption{Wikihow style Example.}
    \label{fig:wikihow_example}
\end{figure*}

\begin{figure*}[htbp]
    \centering
    \scriptsize
\includegraphics[width=1.0\linewidth]{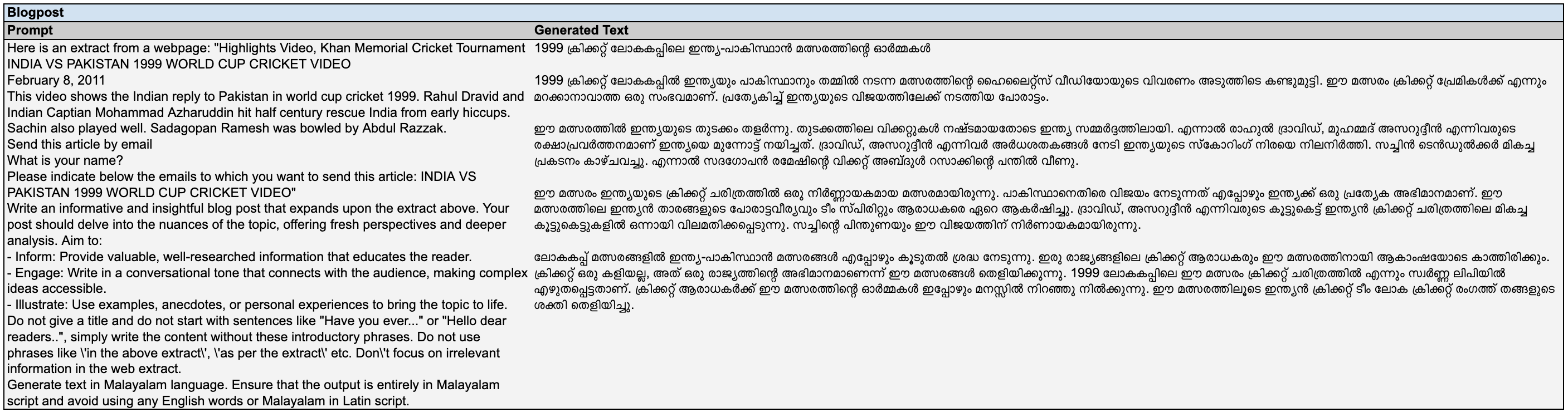} \caption{Blogpost style Example.}
    \label{fig:blogpost_example}
\end{figure*}

\begin{figure*}[htbp]
    \centering
    \scriptsize
\includegraphics[width=1.0\linewidth]{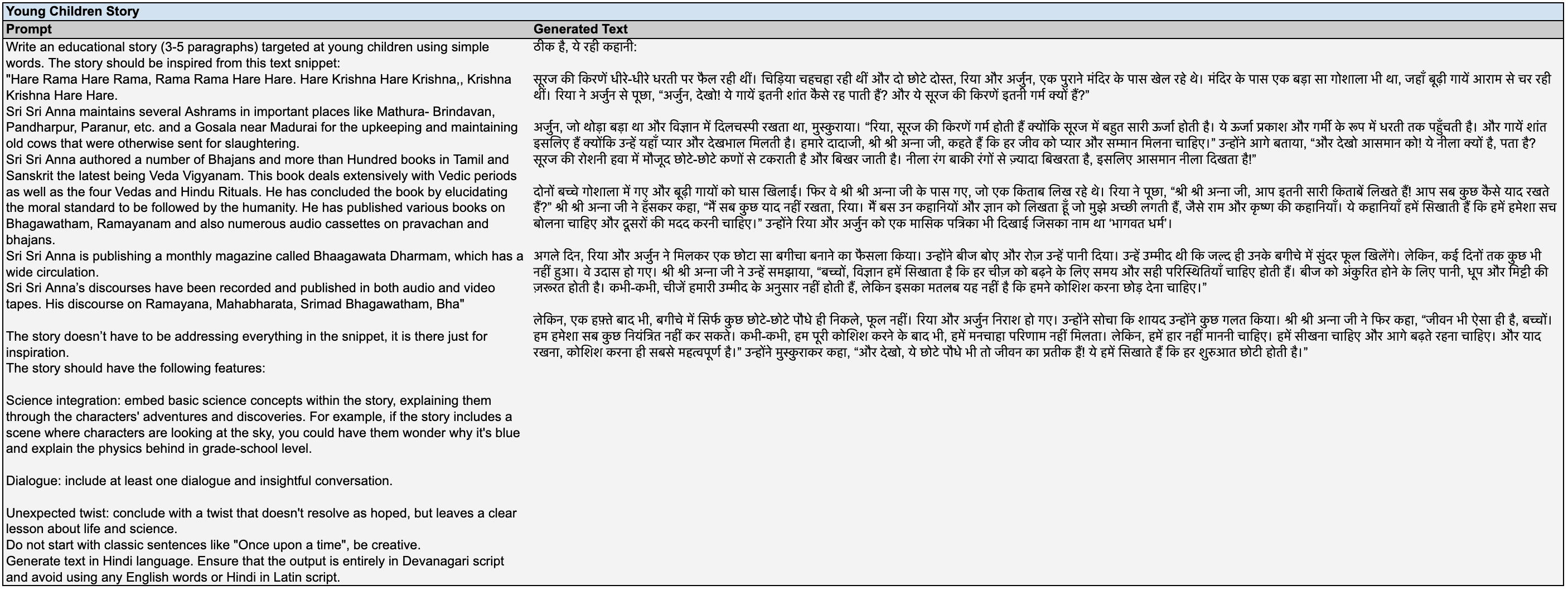} \caption{Young Children Story style Example.}
    \label{fig:young_children_story_example}
\end{figure*}

\begin{figure*}[htbp]
    \centering
    \scriptsize
\includegraphics[width=1.0\linewidth]{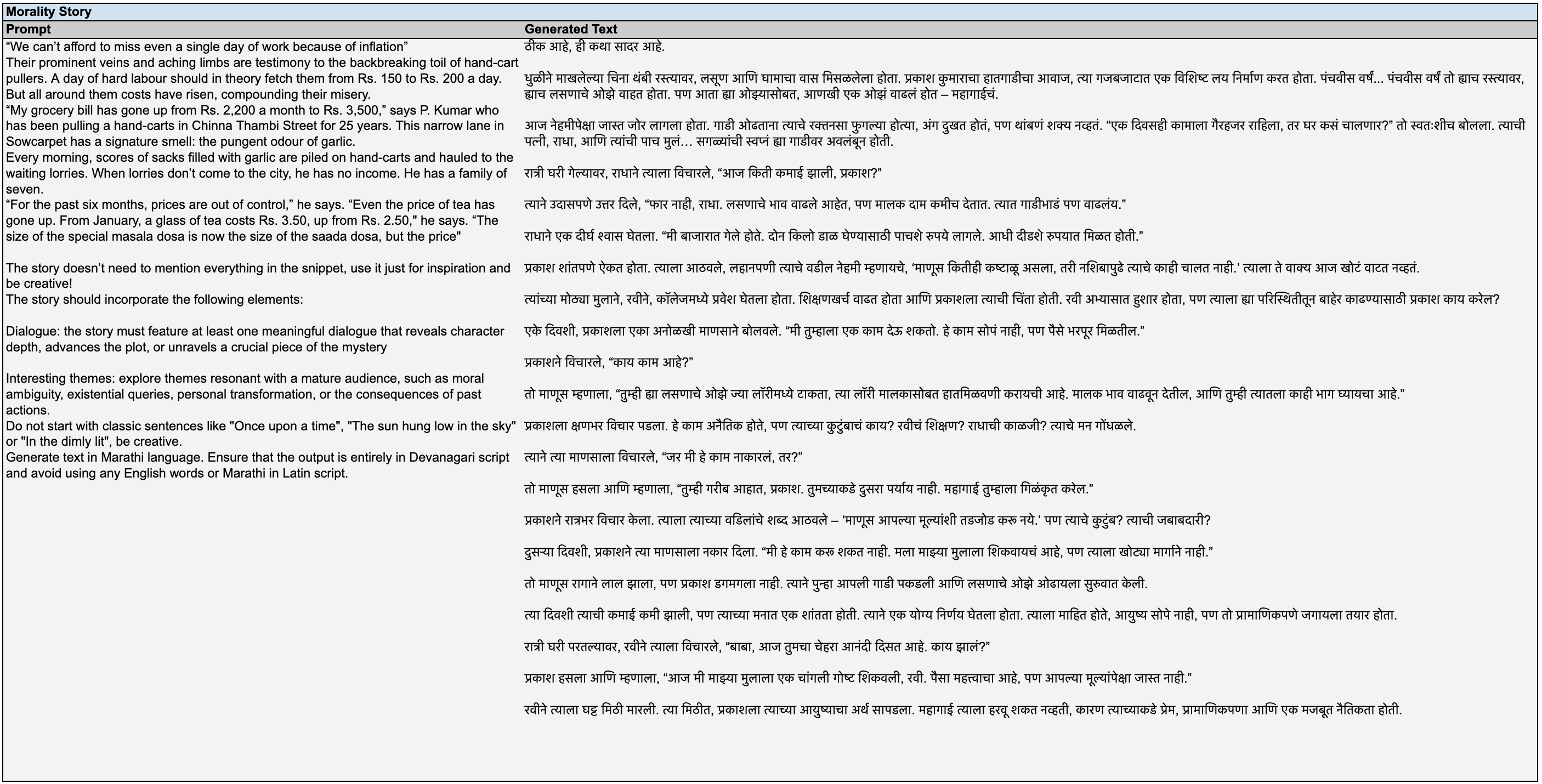} \caption{Morality Story style Example.}
    \label{fig:morality_story_example}
\end{figure*}

\begin{figure*}[htbp]
    \centering
    \scriptsize
\includegraphics[width=1.0\linewidth]{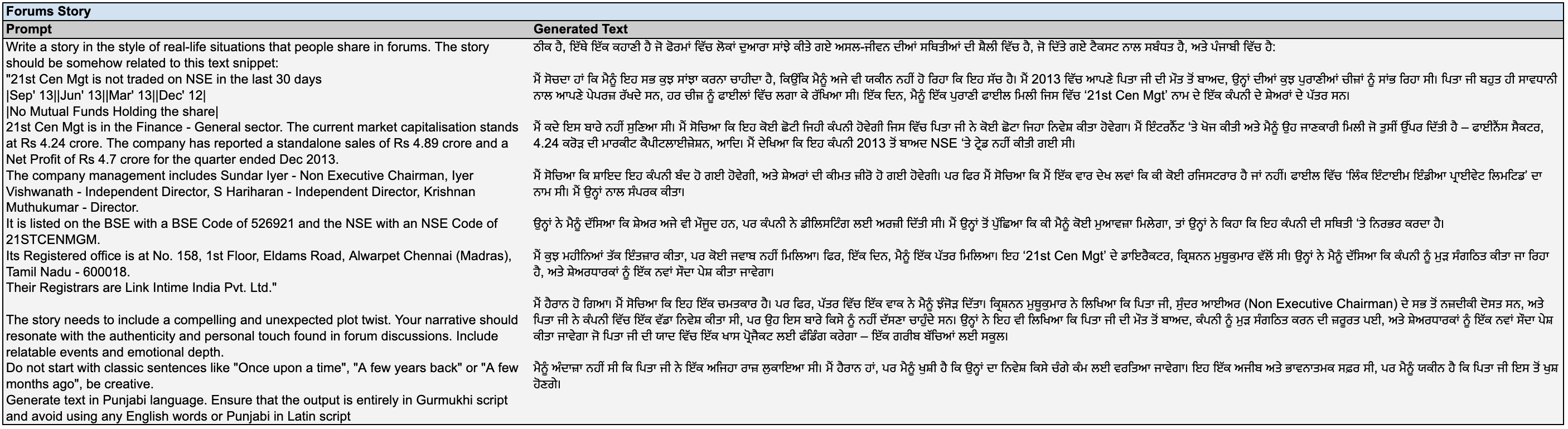} \caption{Forums Story style Example.}
    \label{fig:forums_story_example}
\end{figure*}

\begin{figure*}[htbp]
    \centering
    \scriptsize
\includegraphics[width=1.0\linewidth]{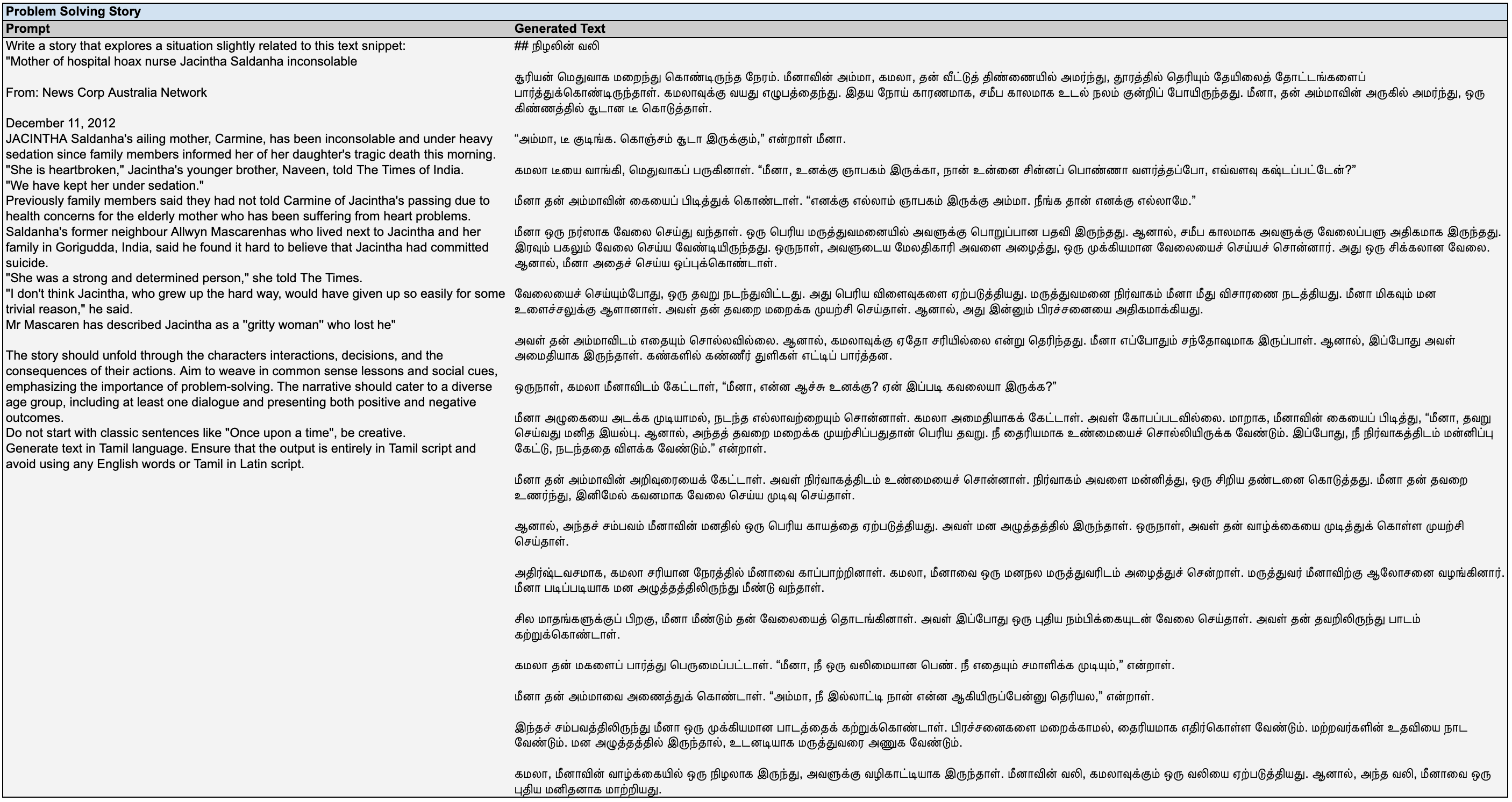} \caption{Problem Solving Story style Example.}
    \label{fig:problem_solving_story_example}
\end{figure*}

\begin{figure*}[htbp]
    \centering
    \scriptsize
\includegraphics[width=1.0\linewidth]{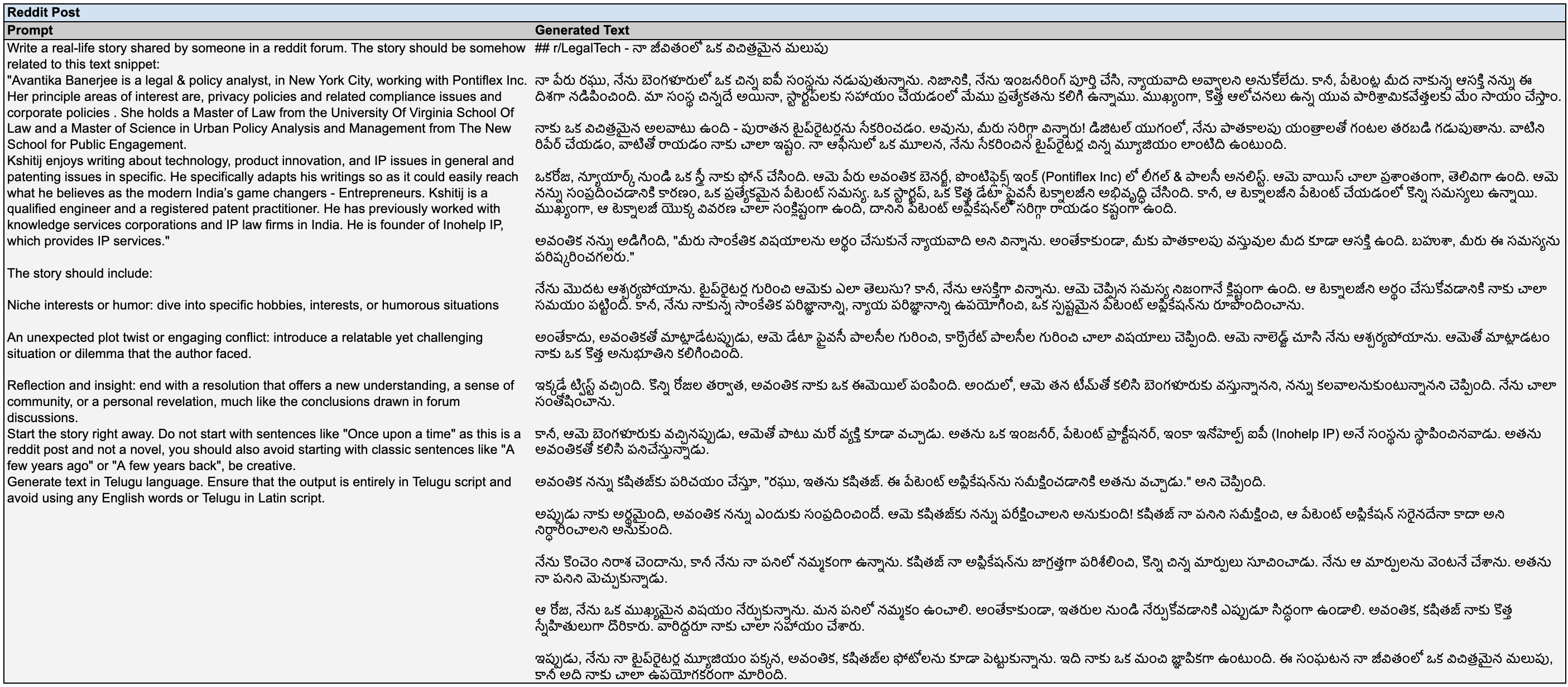} \caption{Reddit Post style Example.}
    \label{fig:reddit_post_example}
\end{figure*}

\begin{figure*}[htbp]
    \centering
    \scriptsize
\includegraphics[width=1.0\linewidth]{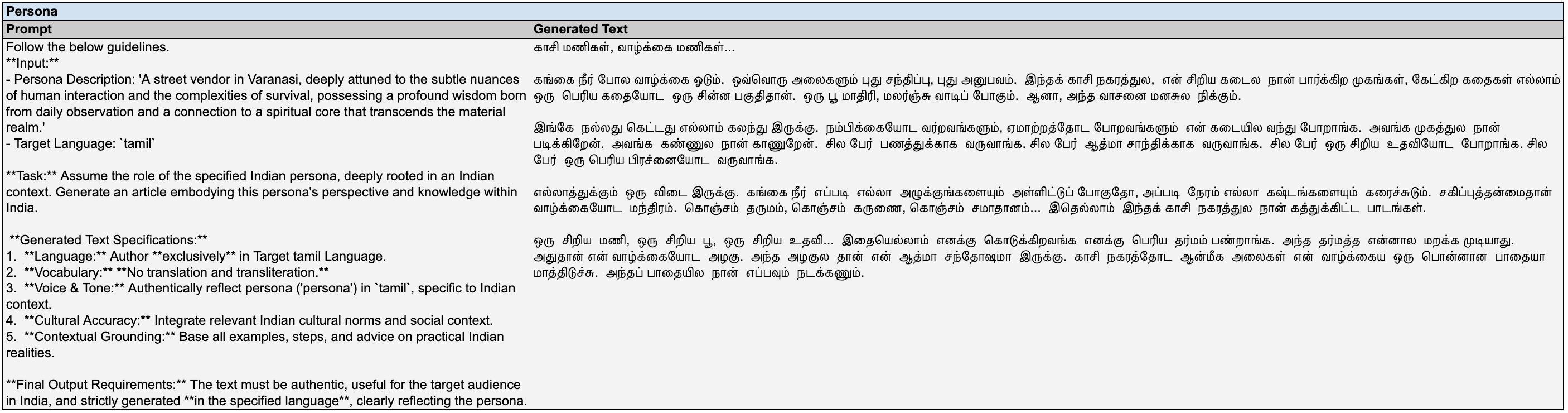} \caption{Persona style Example.}
    \label{fig:persona_example}
\end{figure*}

\begin{figure*}[htbp]
    \centering
    \scriptsize
\includegraphics[width=1.0\linewidth]{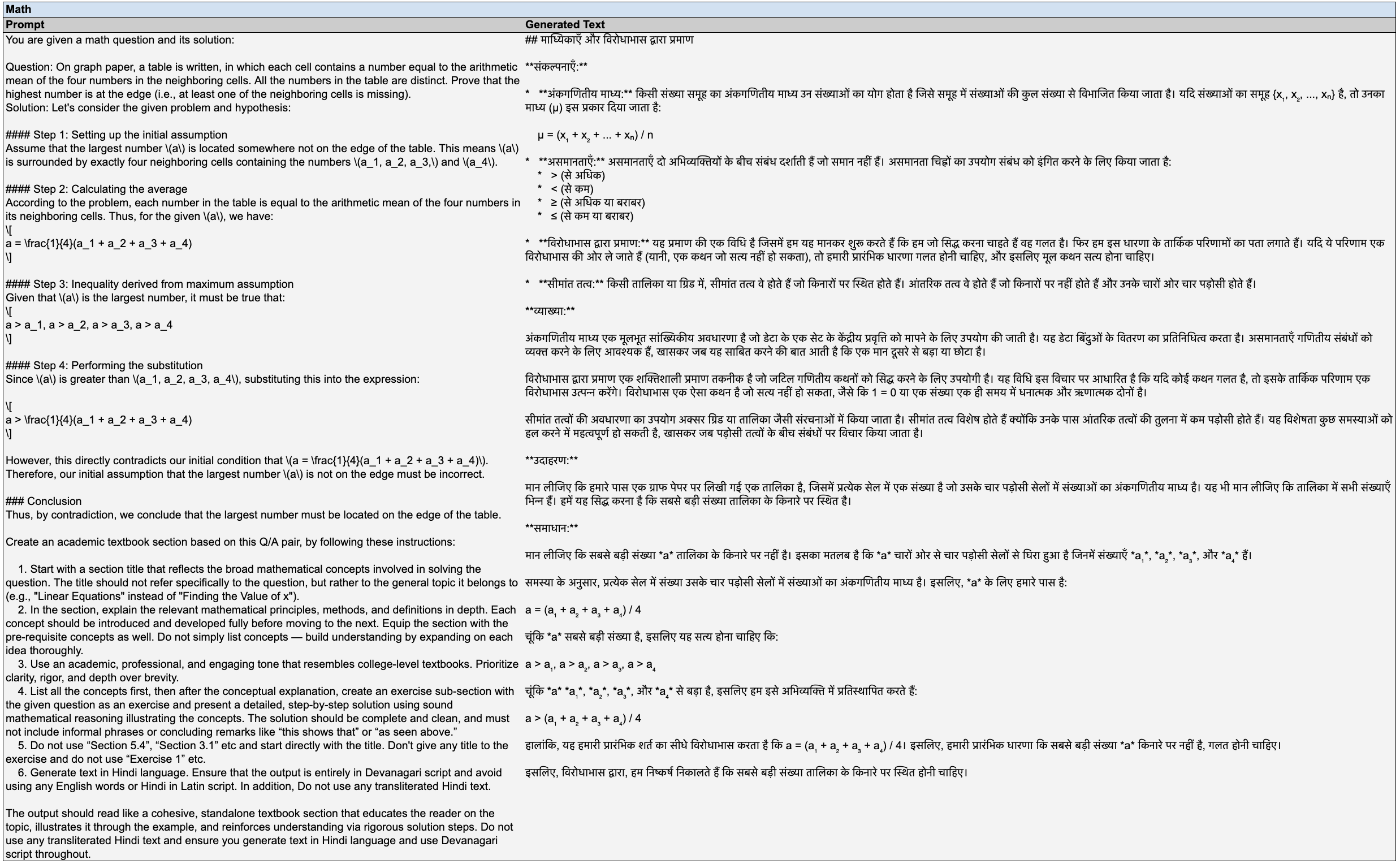} \caption{Math generation Example.}
    \label{fig:math_example}
\end{figure*}

We also show the example generations through different LLMs across various styles in the Figures \ref{fig:textbook_academic_example} - \ref{fig:math_example}. 

\end{document}